\documentclass[10pt,twocolumn,letterpaper]{article}

\usepackage{iccv}
\usepackage{times}
\usepackage{epsfig}
\usepackage{graphicx}
\usepackage{amsmath}
\usepackage{amssymb}
\usepackage{bbm}
\usepackage{booktabs}
\usepackage{multirow}
\usepackage{float}
\usepackage{amsfonts}
\usepackage{comment}
\usepackage{url}
\usepackage[caption=false]{subfig}
\usepackage{relsize}
\usepackage{graphicx,wrapfig,lipsum}
\usepackage{array}
\usepackage{mynotation}
\usepackage[dvipsnames]{xcolor}
\newcommand{\parsection}[1]{\vspace{1mm}\noindent\textbf{#1:}~}
\usepackage{enumitem}
\usepackage{blindtext}
\def\etal{\emph{et al}\onedot}
\def\eg{\emph{e.g}\onedot} 
\def\ie{\emph{i.e}\onedot}

\def\wrt{w.r.t\onedot} 
\def\w{$W$-bipath\xspace}
\def\ipj{$I'J$-bipath\xspace}
\def\ji{$JI$-bipath\xspace}
\makeatother

\newcommand{\bp}[1]{\textbf{#1}}
\newcolumntype{C}[1]{>{\centering\let\newline\\\arraybackslash\hspace{0pt}}m{#1}}

\newcommand{\px}{\mathbf{x}}

\usepackage[pagebackref=true,breaklinks=true,letterpaper=true,colorlinks,bookmarks=false]{hyperref}
\iccvfinalcopy 

\def\iccvPaperID{****} 

\ificcvfinal\fi

\begin{document}

\title{Warp Consistency for Unsupervised Learning of Dense Correspondences}
\author{Prune Truong \qquad Martin Danelljan \qquad Fisher Yu \qquad Luc Van Gool \\
Computer Vision Lab, ETH Zurich, Switzerland\\
\small{\texttt{\{prune.truong, martin.danelljan, vangool\}@vision.ee.ethz.ch  \qquad i@yf.io}} \\
}

\maketitle
\ificcvfinal\thispagestyle{empty}\fi

\begin{abstract}
    
The key challenge in learning dense correspondences lies in the lack of ground-truth matches for real image pairs.
While photometric consistency losses provide unsupervised alternatives, they struggle with large appearance changes, which are ubiquitous in geometric and semantic matching tasks.
Moreover, methods relying on synthetic training pairs often suffer from poor generalisation to real data. 

We propose Warp Consistency, an unsupervised learning objective for dense correspondence regression. Our objective is effective even in settings with large appearance and view-point changes. 
Given a pair of real images, we first construct an image triplet by applying a randomly sampled warp to one of the original images. 
We derive and analyze all flow-consistency constraints arising between the triplet. From our observations and empirical results, we design a general unsupervised objective employing two of the derived constraints. 
We validate our warp consistency loss by training three recent dense correspondence networks for the geometric and semantic matching tasks. Our approach sets a new state-of-the-art on several challenging benchmarks, including MegaDepth, RobotCar and TSS. Code and models are at \url{github.com/PruneTruong/DenseMatching}. 

\end{abstract}

\section{Introduction}

\begin{figure}[t]
\centering%
\vspace{-3mm}
\includegraphics[width=0.75\columnwidth, trim={0 5 0 0}]{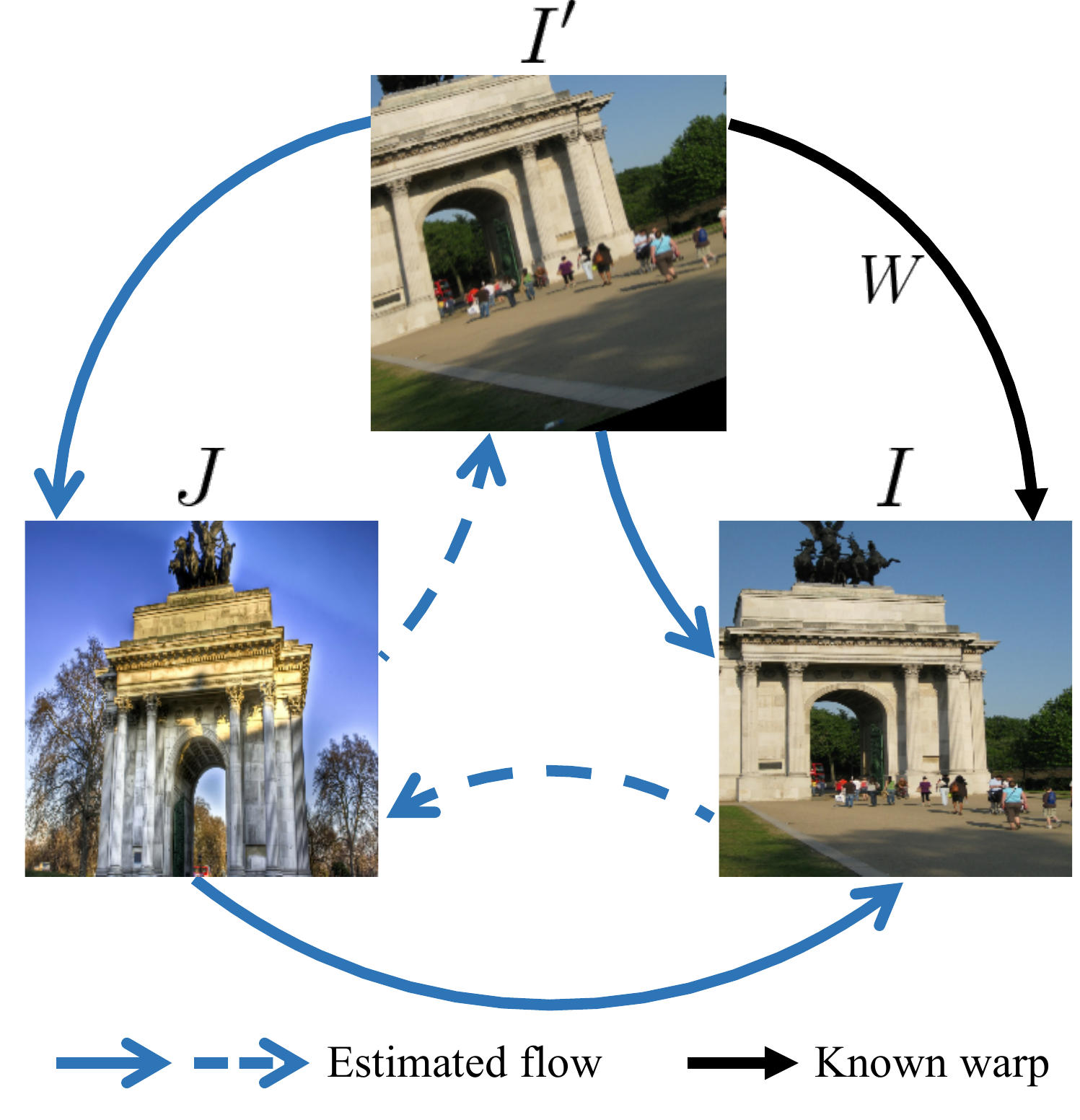}
\vspace{-0mm}\caption{We introduce the warp consistency graph of the image triplet $( I, I', J)$. The image $I'$ is constructed by warping $I$ according to a randomly sampled flow $W$ (black arrow). The blue arrows represent predicted flows. 
Our unsupervised loss is derived from the two constraints represented by the solid arrows, which predict $W$ by the composition $I' \rightarrow J \rightarrow I$ and directly by $I' \rightarrow I$.
}\vspace{-4mm}
\label{fig:intro}
\end{figure}

Finding dense correspondences between images continues to be a fundamental vision problem, with many applications in video analysis~\cite{SimonyanZ14}, image registration~\cite{GLAMpoint, shrivastava-sa11}, image manipulation~\cite{HaCohenSGL11, LiuYT11}, and style transfer~\cite{Kim2019, Liao2017}. 
While supervised deep learning methods have achieved impressive results, they are limited by the availability of ground-truth annotations. 
In fact, collecting dense ground-truth correspondence data of real scenes is extremely challenging and costly, if not impossible. Current approaches therefore resort to artificially rendered datasets~\cite{Dosovitskiy2015, Ilg2017a, Sun2018, Hui2018}, sparsely computed matches~\cite{DusmanuRPPSTS19, D2D}, or sparse manual annotations~\cite{ArbiconNet,MinLPC20, SCNet}. These strategies lack realism, accuracy, or scalability. In contrast, there is a virtually endless source of unlabelled image and video data, which calls for the design of effective unsupervised learning approaches. 

Photometric objectives, relying on the brightness constancy assumption, have prevailed in the context of unsupervised optical flow~\cite{RenYNLYZ17, BackToBasics, Meister2017}.
However, in the more general case of geometric matching, the images often stem from radically different views, captured at different occasions, and under different conditions. This leads to large appearance transformations between the frames, which significantly undermine the brightness constancy assumption. It is further invalidated in the semantic matching task~\cite{LiuYT11},
where the images depict different instances of the same object class. 
As a prominent alternative to photometric objectives, warp-supervision~\cite{GLUNet, GOCor, Rocco2017a, Melekhov2019}, also known as self-supervised learning~\cite{Rocco2018a, SeoLJHC18, MinLPC20}, trains the network on synthetically warped versions of an image. While benefiting from direct supervision, the lack of real image pairs often leads to poor generalization to real data. 

We introduce \emph{Warp Consistency}, an unsupervised learning objective for dense correspondence regression. Our loss leverages real image pairs without invoking the photometric consistency assumption. 
Unlike previous approaches, it is capable of handling large appearance and view-point changes, while also generalizing to unseen real data.
From a real image pair $(I, J)$, we construct a third image $I'$ by warping $I$ with a known flow field $W$, that is created by randomly sampling \eg homographies, from a specified distribution. 
We then consider the consistency graph arising from the resulting image triplet $(I, I', J)$, visualized in Fig.~\ref{fig:intro}. It is used to derive a family of new flow-consistency constraints. By carefully analyzing their properties, we propose an unsupervised loss based on predicting the flow $W$ by the composition $I' \rightarrow J \rightarrow I$ via image $J$ (Fig.~\ref{fig:intro}).  
Our final warp consistency objective is then obtained by combining it with the warp-supervision constraint, also derived from our consistency graph by the direct path $I' \rightarrow I$.

We perform comprehensive empirical analysis of the objectives derived from our warp consistency graph and compare them to existing unsupervised alternatives. 
In particular, our warp consistency loss outperforms approaches based on photometric consistency and warp-supervision on multiple geometric matching datasets. 
We further perform extensive experiments for two tasks by integrating our approach into three recent dense matching architectures, namely GLU-Net~\cite{GLUNet} and RANSAC-Flow~\cite{RANSAC-flow} for geometric matching, and SemanticGLU-Net~\cite{GLUNet} for semantic matching. 
Our unsupervised learning approach brings substantial gains: $+18.2\%$ PCK-5 on MegaDepth~\cite{megadepth} for GLU-Net, $+2.8\%$ PCK-5 for RANSAC-Flow on RobotCar~\cite{RobotCar, RobotCarDatasetIJRR}, as well as $+16.1\%$ and $+ 4.4\%$ PCK-0.05 on PF-Pascal~\cite{PFPascal} and TSS~\cite{Taniai2016} respectively, for SemanticGLU-Net. This leads to a new state-of-the-art on all four datasets. Example predictions are shown in Fig.~\ref{fig:image-type}.

\section{Related work}

\newcommand{\widd}{1.0cm}%
\begin{figure}[t]
\centering%
\vspace{-1mm}
\includegraphics*[width=.95\columnwidth, trim={10 0 0 0}]{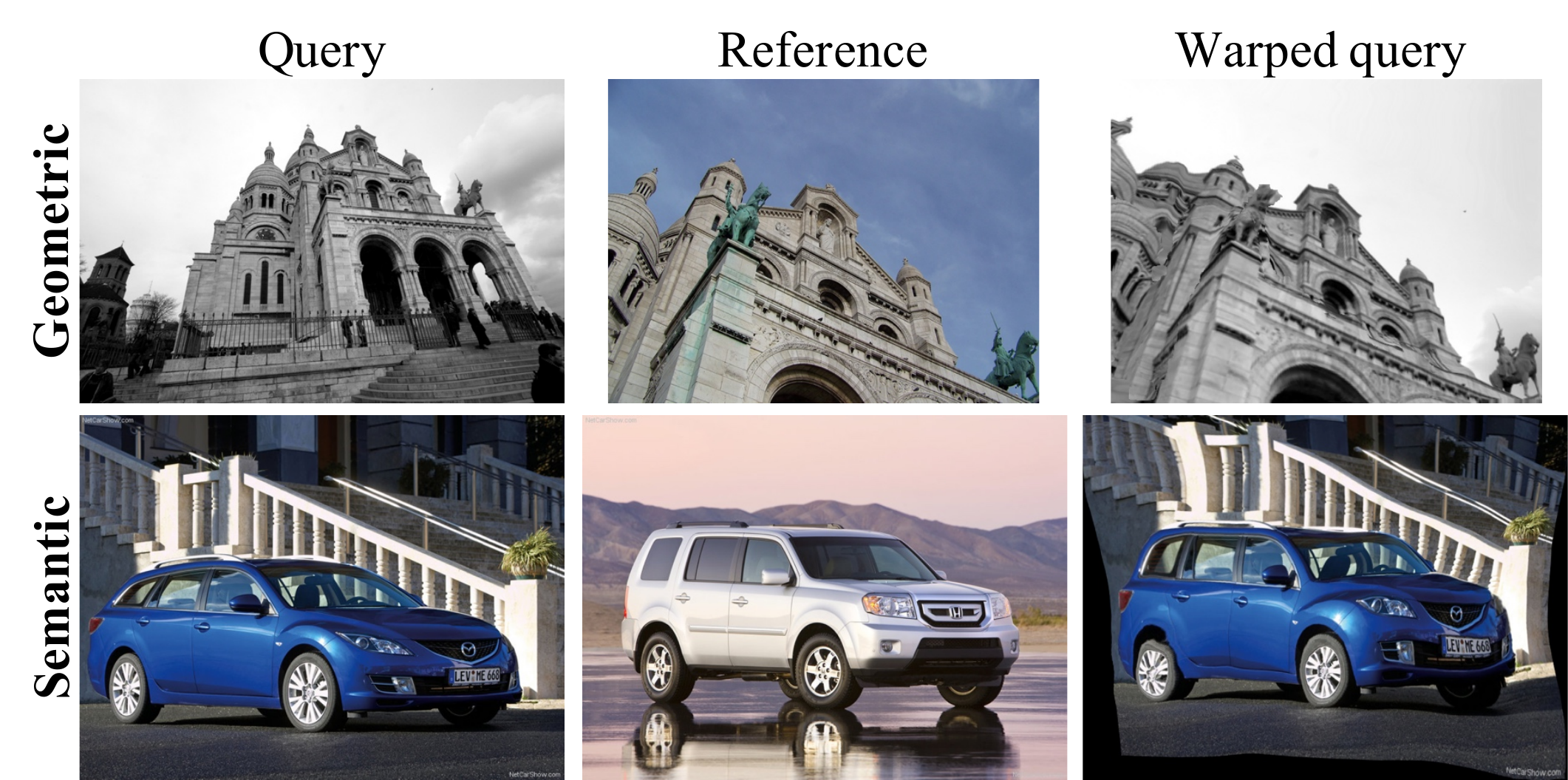}
\vspace{-0mm}\caption{Warped query image (right) according to our predicted flow. Geometric and semantic matching applications pose highly challenging appearance and geometric transformations.
}\vspace{-4mm}
\label{fig:image-type}
\end{figure}

\parsection{Unsupervised optical flow} While supervised optical flow networks need carefully designed synthetic datasets for their training~\cite{Dosovitskiy2015, Mayer2016ALD}, unsupervised approaches do not require ground-truth annotations. Inspired by classical optimization-based methods~\cite{Horn1981}, they instead learn deep models based on brightness constancy and spatial smoothness losses~\cite{RenYNLYZ17, BackToBasics}. The predominant technique mainly relies on photometric losses, \eg Charbonnier penalty~\cite{BackToBasics}, census loss~\cite{Meister2017}, or SSIM \cite{WangBSS04, UnOS}. Such losses are often combined with forward-backward consistency~\cite{Meister2017} and edge-aware smoothness regularization~\cite{OccAwareFlow}. Occlusion estimation techniques~\cite{MFOccFlow, Meister2017, OccAwareFlow} are also employed to mask out occluded or outlier regions from the objective.
Recently, several works~\cite{DDFlow, SefFlow, ARFlow} use a data distillation approach to improve the flow predictions in occluded regions.
However, all aforementioned approaches rely on the assumption of limited appearance changes between two consecutive frames. 
While this assumption holds to a large degree in optical flow data, it is challenged by the drastic appearance changes encountered in geometric or semantic matching applications, as visualised in Fig.~\ref{fig:image-type}.

\parsection{Unsupervised geometric matching}
Geometric matching focuses on the more general case where the geometric transformations and appearance changes between two frames may be substantial. 
Methods either estimate a dense flow field~\cite{Melekhov2019, GLUNet, GOCor, RANSAC-flow} or output a cost volume~\cite{Rocco2018b, D2D}, which can be further refined to increase accuracy~\cite{RoccoAS20, LiHLP20, abs-2012-09842}. The later approaches train the feature embedding, which is then used to compute dense similarity scores. Recent works further leverage the temporal consistency in videos to learn a suitable representation for feature matching~\cite{DwibediATSZ19,JabriOE20, WangJE19}. 
Our work focuses on the first class of methods, which directly learn to regress a dense flow field. 
Recently, Xen~\etal~\cite{RANSAC-flow} use classical photometric and forward-backward consistency losses to train RANSAC-Flow. They partially alleviate the sensitivity of photometric losses to large appearance changes by pre-aligning the images with Ransac. 
Several methods~\cite{Melekhov2019, GLUNet, GOCor} instead use a warp-supervision loss. By posing the network to regress a randomly sampled warp during training, a direct supervisory signal is obtained, but at the cost of poorer generalization abilities to real data.

\parsection{Semantic correspondences}
Semantic matching poses additional challenges due to intra-class appearance and shape variations. 
Manual annotations in this context are ill-defined and ambiguous, making it crucial to develop unsupervised objectives. 
Methods rely on warp-supervision strategies~\cite{Rocco2017a, Rocco2018a, ArbiconNet, SeoLJHC18, GLUNet}, use proxy losses on the cost volume~\cite{DCCNet, Rocco2018b, Rocco2018a, MinLPC20}, 
identify correct matches from forward-backward consistency of the cost volumes~\cite{Jeon}, or jointly learn semantic correspondence with attribute transfer~\cite{Kim2019} or segmentation~\cite{SFNet}. 
Most related to our work are~\cite{Zhou2016, abs-2004-09061, ZhouLYE15}. 
Zhou~\etal~\cite{Zhou2016} learn to align multiple images using 3D-guided cycle-consistency by leveraging the ground-truth matches between multiple CAD models. However, the need for 3D CAD models greatly limits its applicability in practice.
In FlowWeb~\cite{ZhouLYE15}, the authors optimize online pre-existing pair-wise correspondences using the cycle consistency of flows between images in a collection. 
Unlike these approaches, we require pairs of images as unique supervision and propose a general loss formulation, learning to regress dense correspondences directly.

\section{Method}
\label{sec:method}

\subsection{Problem formulation and notation}
\label{Sec:formulation-and-notations}

We address the problem of finding pixel-wise correspondences between two images $I \in \reals^{h \times w \times 3}$ and $J \in \reals^{h \times w \times 3}$.
Our goal is to estimate a dense displacement field $F_{I \rightarrow  J} \in \reals^{h \times w \times 2}$, often referred to as flow, relating pixels in $I$ to $J$. The flow field $F_{I \rightarrow  J}$ represents the pixel-wise 2D motion vectors in the coordinate system of image $I$. 
It is directly related to the mapping $M_{I \rightarrow  J} \in \reals^{h \times w \times 2}$, which encodes the absolute location $M_{I \rightarrow  J}(\px) \in \reals^2$ in $J$ corresponding to the pixel location $\px \in \reals^2$ in image $I$. It is thus related to the flow through $M_{I \rightarrow  J}(\px) = \px + F_{I \rightarrow  J}(\px)$. It is important to note that the flow and mapping representations are asymmetric. $M_{I \rightarrow  J}$ parametrizes a mapping from each pixel in image $I$, which is not necessarily bijective. 

With a slight abuse of notation, we interchangeably view $F_{I \rightarrow  J}$ and $M_{I \rightarrow  J}$ as either elements of $\reals^{h \times w \times 2}$ or as functions $F_{I \rightarrow  J}, M_{I \rightarrow  J} : \reals^2 \rightarrow \reals^2$. The latter is generally obtained by a bilinear interpolation of the former, and the interpretation will be clear from context when important. 
We define the \emph{warping} $\warp_F (T)$ of a function $T : \reals^2 \rightarrow \reals^d$ by the flow  $F$ as $\warp_F (T)(\px) = T \left( \px + F(\px) \right)$. This is more compactly expressed as $\warp_F (T) = T \circ M_F$, where $M_F$ is the mapping defined by $F$ and $\circ$ denotes function composition. Lastly, we let $\mathbb{I} : \reals^2 \rightarrow \reals^2$ be the identity map $\mathbb{I}(\px) = \px$. 

The goal of this work is to learn a neural network $f_\theta$, with parameters $\theta$, that predicts an estimated flow $\widehat{F}_{I \rightarrow  J} = f_{\theta} ( I, J )$ relating $I$ to $J$. We will consistently use the hat $\widehat{\cdot}$ to denote an estimated or predicted quantity. The straightforward approach to learn $f_\theta$ is to minimize the discrepancy between the estimated flow $\widehat{F}_{I \rightarrow  J}$ and the ground-truth flow $F_{I \rightarrow  J}$ over a collection of real training image pairs $(I,J)$. 
However, such supervised training requires large quantities of densely annotated data, which is extremely difficult to acquire for real scenes. This motivates the exploration of unsupervised alternatives for learning dense correspondences.

\subsection{Unsupervised data losses}
\label{sec:unsu-data-losses}

To develop our approach, we first briefly review relevant existing alternatives for unsupervised learning of flow. 
While there is no general agreement in the literature, we adopt a \emph{practical} definition of unsupervised learning in our context. We call a learning formulation `unsupervised' if it does not require any information (\ie supervision) other than pairs of images $(I,J)$ depicting the same scene or object.
Specifically, unsupervised methods do not require any annotations made by humans or other matching algorithms.

\begin{figure}[t]
{\centering%
\vspace{-2mm}
\newcommand{\wid}{0.99\columnwidth}%
\includegraphics*[width=\wid, trim=0 16 0 0 ]{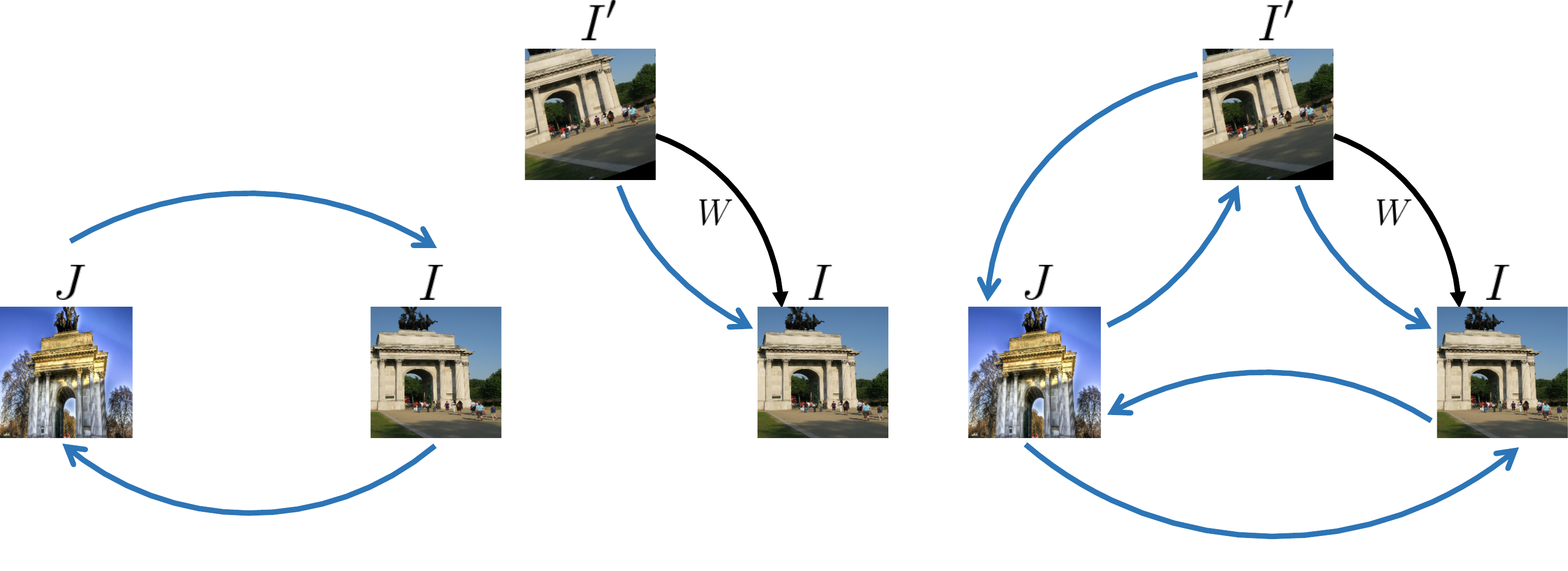}}
\footnotesize{\hspace{0.4cm}(a) Forw.-backw.~\eqref{eq:forward-backward}  \hspace{0.2cm} (b) Warp-superv.~\eqref{eq:warp-supervision} \hspace{0.3cm} (c) Warp consistency 
}
\caption{Alternative unsupervised strategies. }\vspace{-4mm}
\label{fig:unsu-alternatives}
\end{figure}

\parsection{Photometric losses}
Most unsupervised approaches train the network using a photometric loss~\cite{BackToBasics, Meister2017, OccAwareFlow, RANSAC-flow}. Under the photometric consistency assumption, it minimizes the difference between image $I$ and image $J$ warped according to the estimated flow field $\widehat{F}_{I \rightarrow J}$ as,
\begin{equation}
    \label{eq:photo-loss}
    L_{\text{photo}} = \rho \left( I \,,\, \warp_{\widehat{F}_{I \rightarrow J}} (J) \right) \,.
\end{equation}
Here, $\rho(\cdot,\cdot)$ is a function measuring the difference between two images, \eg $L_2$~\cite{BackToBasics}, SSIM~\cite{WangBSS04}, or census~\cite{Meister2017}.

\parsection{Forward-backward consistency}
By constraining the backward flow $\widehat{F}_{J \rightarrow  I}$ to yield the reverse displacement of its forward counterpart $\widehat{F}_{I \rightarrow  J}$, we achieve the forward-backward consistency loss~\cite{Meister2017},
\begin{equation}
    \label{eq:forward-backward}
   L_{\text{fb}} = \big \|  \widehat{F}_{I \rightarrow  J} + 
   \warp_{\widehat{F}_{I \rightarrow  J}} (\widehat{F}_{J \rightarrow  I}) \big \|  \,.
\end{equation}
Here, $\|\cdot\|$ denotes a suitable norm.
While well motivated,~\eqref{eq:forward-backward} is enforced by the trivial degenerate solution of always predicting zero flow $\widehat{F}_{I \rightarrow  J}=\widehat{F}_{J \rightarrow  I} = 0$. It therefore bares the risk of degrading performance by biasing the prediction towards zero, even if combined with a photometric loss~\eqref{eq:photo-loss}. 
Both aforementioned losses are most often used together with a visibility mask that filters out the influence of occluded regions from the objective.

\parsection{Warp-supervision}
Another approach relies on synthetically generated training pairs, where the ground-truth flow is obtained by construction~\cite{GLUNet, Rocco2017a, Melekhov2019}. Given only a single image $I$, a training pair $\left( I, I' \right)$ is created by applying a randomly sampled transformation $W$, \eg a homography, to $I$ as $I' = \warp_{W} \left( I \right)$. Here, $W$ is the synthetic flow field, which serves as direct supervision through a regression loss,
\begin{equation}
    \label{eq:warp-supervision}
    L_{\text{warp}} = \big \| \widehat{F}_{I' \rightarrow I} - W \big \| \,.
\end{equation} 
While this results in a strong and direct training signal, warp supervision methods struggle to generalize to real image pairs $(I,J)$. 
This can lead to over-smooth predictions and instabilities in the presence of unseen appearance changes.

\begin{figure*}[t]
\centering%
\vspace{-4mm}
\newcommand{\wid}{0.20\textwidth}%
\subfloat[\ipj \label{fig:F_I_prime_to_J}]{\includegraphics[width=\wid]{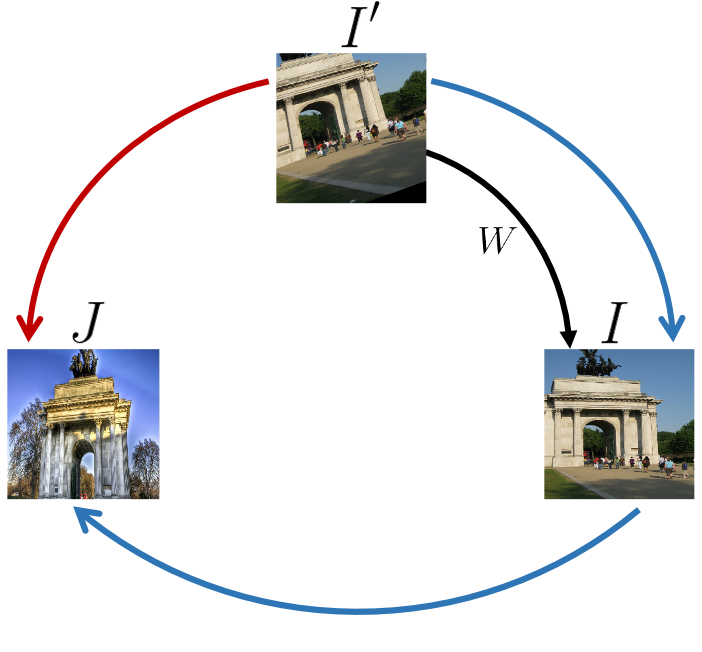}}~%
\subfloat[\ji \label{fig:F_J_to_I}]{\includegraphics[width=\wid, trim=0 0 0 50]{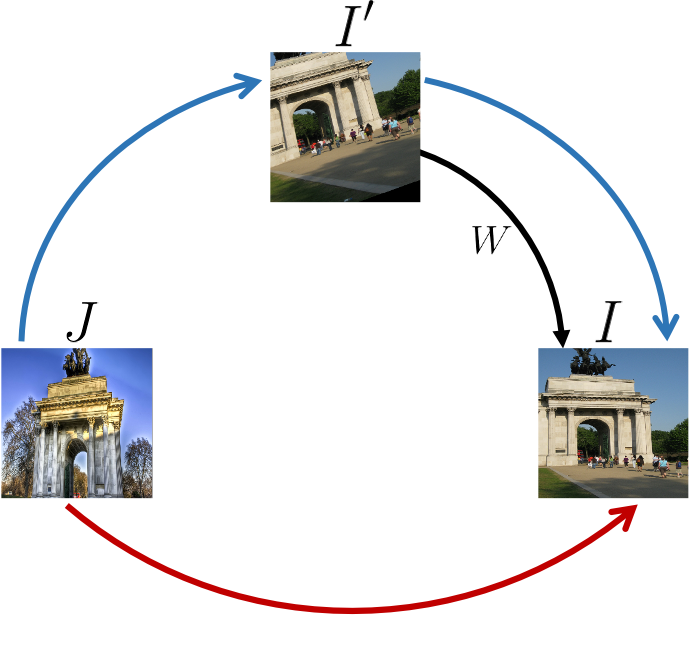}}~%
\subfloat[\w \label{fig:W}]{\includegraphics[width=\wid, trim=0 0 0 50]{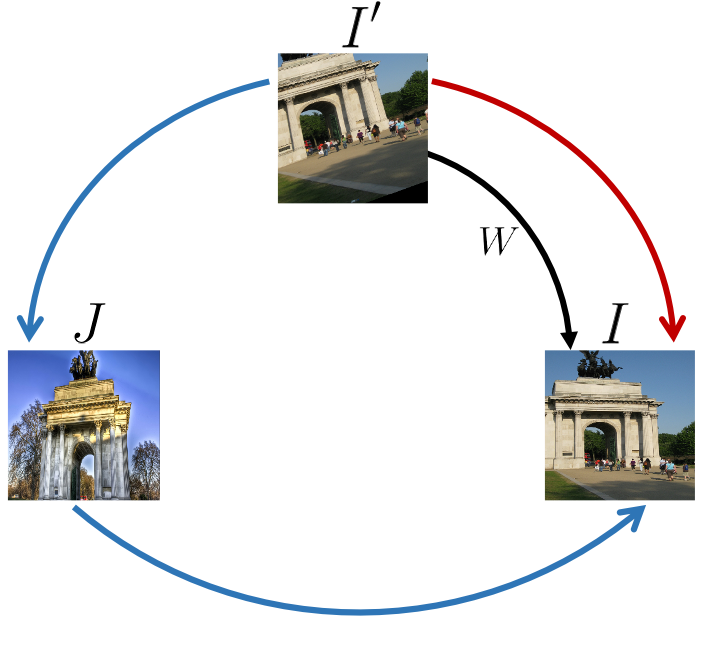}}~%
\subfloat[\centering Cycle consistency \label{fig:cycles}]{\includegraphics[width=\wid, trim=0 0 0 50]{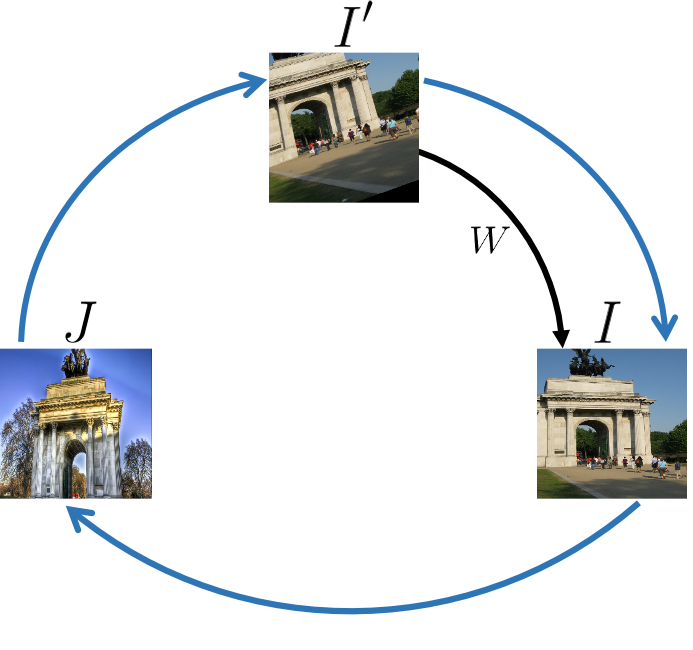}}~%
\subfloat[\centering Pair-wise consistency \label{fig:pair-wise}]{\includegraphics[width=\wid, trim=0 0 0 50]{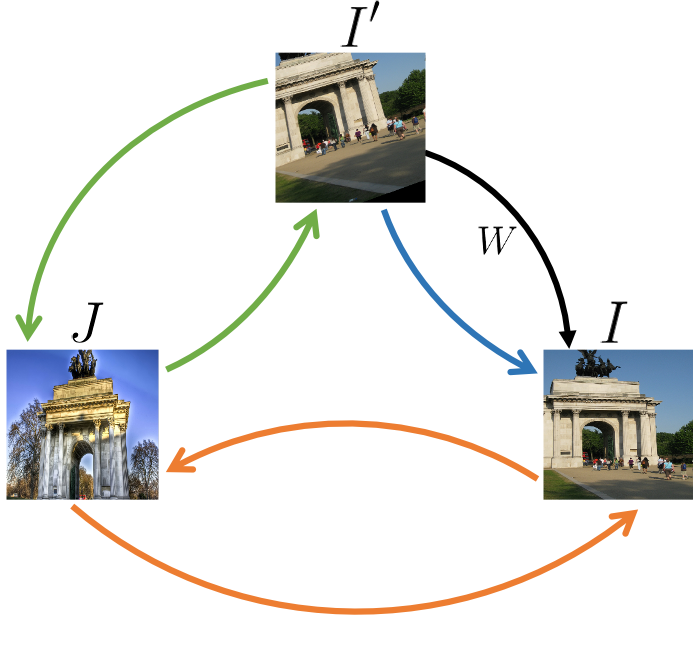}}
\vspace{1mm}\caption{Consistency relations derived from our warp consistency graph constructed between images $( I, I', J)$.
For the bipaths constraints a, b and c, the red and blue arrows indicate the paths used for the left and right hand side, respectively, of the constraints in~\eqref{eq:bipath-mapping}-\eqref{eq:bipath-flow}.
}\vspace{-4mm}
\label{fig:cycles-method}
\end{figure*}

\subsection{Warp consistency graph}
\label{sec:cyclic-consistency-graph}

We set out to find a new unsupervised objective suitable for scenarios with large appearance and view-point changes, where photometric based losses struggle. While the photometric consistency assumption is avoided in the forward-backward consistency (Fig.~\ref{fig:unsu-alternatives}a) and warp-supervision (Fig.~\ref{fig:unsu-alternatives}b) objectives, these methods suffer from severe drawbacks in terms of degenerate solutions and lack of realism, respectively. To address these issues, we consider all possible consistency relations obtained from the three images involved in both aforementioned objectives. Using this generalization, we not only retrieve forward-backward and warp-supervision as special cases, but also derive a \emph{family of new consistency relations}.

From an image pair $(I,J)$, we first construct an image triplet $(I, I', J)$ by warping $I$ with a known flow-field $W$ in order to generate the new image $I' = \warp_W (I)$. We now consider the full consistency graph, visualized in Fig.~\ref{fig:unsu-alternatives}c, encompassing all flow-consistency constraints derived from the triplet of images $(I, I', J)$. Crucially, we exploit the fact that the transformation $F_{I' \rightarrow I} = W$ is known. 
The goal is to find consistency relations that translate to suitable learning objectives. Particularly, we wish to improve the network prediction between the real image pair $(I,J)$. We therefore first explore the possible consistency constraints that can be derived from the graph shown in Fig.~\ref{fig:unsu-alternatives}c. For simplicity, we do not explicitly denote visible or valid regions of the stated consistency relations. They should be interpreted as an equality constraint for all pixel locations $\mathbf{x}$ where both sides represent a valid, non-occluded mapping or flow.

\parsection{Pair-wise constraints}
We first consider the consistency constraints recovered from pairs of images, as visualized in Fig.~\ref{fig:pair-wise}. From the pair $\left ( I, J \right)$, and analogously $\left ( J, I' \right )$, we recover the standard forward-backward consistency constraint $\mathbb{I} = M_{J\rightarrow I} \circ M_{I\rightarrow J}$, from which we derive~\eqref{eq:forward-backward}. 
Furthermore, from the pair $\left ( I', I \right )$ we can derive the warp-supervision constraint~\eqref{eq:warp-supervision} $F_{I' \rightarrow I} = W$ .\footnote{While $\mathbb{I} = M_{I\rightarrow I'} \circ M_{W}$ and $\mathbb{I} = M_W \circ M_{I\rightarrow I'}$ are also possible, they offer no advantage over standard warp-supervision: $M_{I'\rightarrow I} = M_{W}$.}

\parsection{Bipath constraints} 
The novel consistency relations stem from constraints that involve all three images in the triplet $( I, I', J)$. These appear in two distinct types, here termed \emph{bipath} and \emph{cycle} constraints, respectively. We first consider the former, which have the form $M_{1 \rightarrow 2} = M_{3 \rightarrow 2} \circ M_{1 \rightarrow 3}$. That is, we obtain the same mapping by either proceeding directly from image 1 to 2 or by taking the detour through image 3. We thus compute the same mapping by two different \emph{paths}: $1 \rightarrow 2$ and $1 \rightarrow 3 \rightarrow 2$, from which we derive the name of the constraint. The images 1, 2, and 3 represent any enumeration of the triplet $(I, I', J)$ that respects the direction $I' \rightarrow I$, specified by the known warp $W$. There thus exist three different bipath constraints, detailed in Sec.~\ref{sec:semicycle}.

\parsection{Cycle constraints} The last category of constraints is formulated by starting from any of the three images in Fig.~\ref{fig:cycles} and composing the mappings in a full cycle. Since we return to the starting image, the resulting composition is equal to the identity map. This is expressed in a general form as $\mathbb{I} = M_{3 \rightarrow 1} \circ M_{2 \rightarrow 3} \circ M_{1 \rightarrow 2}$, where we have proceeded in the cycle $1 \rightarrow 2 \rightarrow 3 \rightarrow 1$.
Again constraining the direction $I' \rightarrow I$, we obtain three different constraints, as visualized in Fig.~\ref{fig:cycles}. Compared to the bipath constraints, the cycle variants require two consecutive warping operations, stemming from the additional mapping composition. Each warp reduces the valid region and introduces interpolation noise and artifacts in practice. Constraints involving fewer warping operations are thus desirable, which is an advantage of the class of bipath constraints. 
In the next parts, we therefore focus on the later class to find a suitable unsupervised objective for dense correspondence estimation.

\subsection{Bipath constraints}
\label{sec:semicycle}

As mentioned in the previous section, there exist three different bipath
constraints that preserve the direction of the known warp $W$. 
These are stated in terms of mappings as, 
\begin{subequations}
\label{eq:bipath-mapping}
\begin{align}
M_{I'\rightarrow J} &= M_{I \rightarrow J} \circ M_{W} \label{eq:semi-cycle-map-a} \\ 
M_{J \rightarrow I} &= M_{W} \circ M_{J\rightarrow I'}  \label{eq:semi-cycle-map-b} \\
M_{W} &= M_{J\rightarrow I'} \circ M_{I\rightarrow J}  \label{eq:semi-cycle-map-c} \,.
\end{align}
\end{subequations} 
From~\eqref{eq:bipath-mapping}, we can derive the equivalent flow constraints as,
\begin{subequations}
\label{eq:bipath-flow}
\begin{align}
F_{I'\rightarrow J} &= W + \warp_{W} (F_{I \rightarrow J} )  \label{eq:semi-cycle-flow-a} \\ 
F_{J \rightarrow I} &= F_{J\rightarrow I'} + \warp_{F_{J\rightarrow I'}} (W)  \label{eq:semi-cycle-flow-b}  \\ 
W &= F_{I'\rightarrow J} + \warp _{F_{I'\rightarrow J}} (F_{J \rightarrow I} )  \label{eq:semi-cycle-flow-c} \,.
\end{align}
\end{subequations}
Each constraint is visualized in Fig.~\ref{fig:cycles-method}a, b and c respectively. At first glance, any one of the constraints in~\eqref{eq:bipath-flow} could be used as an unsupervised loss by minimizing the error between the left and right hand side. However, by separately analyzing each constraint in~\eqref{eq:bipath-mapping}-\eqref{eq:bipath-flow}, we will find them to have radically different properties which impact their suitability as an unsupervised learning objective.

\parsection{\ipj}  The constraint~\eqref{eq:semi-cycle-map-a},~\eqref{eq:semi-cycle-flow-a} is derived from the two possible paths from $I'$ to $J$ (Fig.~\ref{fig:F_I_prime_to_J}). While not obvious from~\eqref{eq:semi-cycle-flow-a}, it can be directly verified from~\eqref{eq:semi-cycle-map-a} that this constraint has a degenerate trivial solution. In fact,~\eqref{eq:semi-cycle-map-a} is satisfied for any $W$ by simply mapping all inputs $\mathbf{x}$ to a constant pixel location $\mathbf{c} \in \reals^2$ as $\widehat{M}_{I'\rightarrow J}(\mathbf{x}) = \widehat{M}_{I \rightarrow J}(\mathbf{x}) = \mathbf{c}$. In order to satisfy this constraint, the network can thus learn to predict the same flow $\widehat{F} = \mathbf{c} - \mathbb{I}$ for any input image pair. 

\parsection{\ji} From the paths $J \rightarrow I$ in Fig.~\ref{fig:F_J_to_I}, we achieve the constraint~\eqref{eq:semi-cycle-map-b},~\eqref{eq:semi-cycle-flow-b}. The resulting unsupervised loss is formulated as 
\begin{equation}
\label{eq:ji-bipath}
L_{J \rightarrow I} =  \big \| \widehat{F}_{J\rightarrow I'} + \warp_{\widehat{F}_{J\rightarrow I'}} (W)  - \widehat{F}_{J \rightarrow I} \big \|.
\end{equation}
Unfortunately, this objective suffers from another theoretical disadvantage. Due to the cancellation effect between the estimated flow terms $\widehat{F}_{J\rightarrow I'}$ and $\widehat{F}_{J \rightarrow I}$, the objective~\eqref{eq:ji-bipath} is insensitive to a constant bias in the prediction. Specifically, if a small constant bias $\mathbf{b} \in \reals^2$ is added to all flow predictions in~\eqref{eq:ji-bipath}, it can be shown that the increase in the loss $\eqref{eq:ji-bipath}$ is approximately bounded by $\big\| \warp_{\widehat{F}_{J\rightarrow I'}}(DW \mathbf{b}) \big\|$. Here, the bias error $\mathbf{b}$ is scaled with the Jacobian $DW$ of the warp $W$. Since a smooth and invertible warp $W$ implies a generally small Jacobian $DW$, the change in the loss will be negligible. The resulting insensitivity of~\eqref{eq:ji-bipath} to a prediction bias is further confirmed empirically by our experiments. 
We provide derivations in the suppl.~\ref{subsec-sup:ji}. 
To further understand and compare the bipath constraints~\eqref{eq:bipath-flow}, it is also useful to consider the limiting case of reducing the magnitude of the warps $\|W\| \rightarrow 0$. By setting $W=0$ it can be observed that~\eqref{eq:ji-bipath} becomes zero, \ie no learning signal remains. 

\parsection{\w} The third bipath constraint~\eqref{eq:semi-cycle-map-c},~\eqref{eq:semi-cycle-flow-c} is derived from the paths $I' \rightarrow I$, which is determined by $W$ (Fig.~\ref{fig:W}). It leads to the \w consistency loss,
\begin{equation}
    \label{eq:w-bipath}
    L_W =  \big \| \widehat{F}_{I'\rightarrow J} + \warp _{\widehat{F}_{I'\rightarrow J}} (\widehat{F}_{J \rightarrow I})  - W  \big \| \,.
\end{equation}
We first analyze the limiting case $\|W\| \rightarrow 0$ by setting $W=0$, which leads to standard forward-backward consistency~\eqref{eq:forward-backward} since $I'=I$. The \w is thus a direct generalization of the latter constraint. Importantly, by randomly sampling non-zero warps $W$, degenerate solutions are avoided, effectively solving the one fatal issue of forward-backward consistency objectives. 
In addition to avoiding degenerate solutions, \w does not experience cancellation of prediction bias, as in~\eqref{eq:ji-bipath}. 
Furthermore, compared to warp-supervision \eqref{eq:warp-supervision}, it enables to directly learn the flow prediction $\widehat{F}_{J \rightarrow I}$ between the real pair $(I,J)$. In the next section, we therefore develop our final unsupervised objective based on the \w consistency. 

\subsection{Warp consistency loss}
\label{sec:our-loss}

In this section, we develop our warp consistency loss, an unsupervised learning objective for dense correspondence estimation, using the consistency constraints derived in Sec.~\ref{sec:cyclic-consistency-graph} and \ref{sec:semicycle}. Specifically, following the observations in Sec.~\ref{sec:semicycle}, we base our loss on the \w constraint.

\parsection{\w consistency term} 
To formulate an objective based on the \w consistency constraint~\eqref{eq:semi-cycle-flow-c}, we further integrate a visibility mask $V \in [0,1]^{w \times h}$. The mask $V$ takes a value $V(\mathbf{x}) = 1$ for any pixel $\mathbf{x}$ where both sides of~\eqref{eq:semi-cycle-map-c},~\eqref{eq:semi-cycle-flow-c} represent a valid, non-occluded mapping, and $V(\mathbf{x}) = 0$ otherwise. The loss~\eqref{eq:w-bipath} is then extended as, 
\begin{equation}
\label{eq:vis-w-bipath}
\centering
    L_{\text{W-vis}} =  \left \| \widehat{V} \cdot \left (\widehat{F}_{I'\rightarrow J} + \warp _{\widehat{F}_{I'\rightarrow J}} (\widehat{F}_{J \rightarrow I})  - W \right ) \right \|  \,.
\end{equation}
Since we do not know the true $V$, we replace it with an estimate $\widehat{V}$. While there are different techniques for estimating visibility masks~\cite{MFOccFlow, Meister2017, OccAwareFlow}, we base our strategy on~\cite{Meister2017}. Specifically, we compute our visibility mask as,
\begin{align}\label{eq:vis-mask}
    \widehat{V} = \mathbbm{1}\bigg[& \big| \widehat{F}_{I'\rightarrow J} + \warp _{\widehat{F}_{I'\rightarrow J}} (\widehat{F}_{J \rightarrow I} ) - W \big|^2_2 <  \alpha_2 \,+  \\ 
    & \alpha_1 \left ( \big| \widehat{F}_{I'\rightarrow J} \big|^2_2 + \big| \warp _{\widehat{F}_{I'\rightarrow J}} (\widehat{F}_{J \rightarrow I} )\big|^2_2  + \left| W \right|^2_2   \right ) \bigg]. \nonumber
\end{align}
Here, $\mathbbm{1}[\cdot]$ takes the value 1 or 0 if the input statement is true or false, respectively. The scalars $\alpha_1$ and $\alpha_2$ are hyperparameters controlling the sensitivity of the mask estimation.
For the warp operation $\warp_{\widehat{F}_{I' \rightarrow J}}(\widehat{F}_{J \rightarrow I})$, we generally found it beneficial not to back-propagate gradients through the flow $\widehat{F}_{I' \rightarrow J}$ used for warping. We believe that this better encourages the network to directly adjust the flow $\widehat{F}_{J \rightarrow I}$, rather than `move' the flow vectors using the warp $\warp_{\widehat{F}_{I' \rightarrow J}}$.

\parsection{Warp-supervision term} 
In addition to our \w objective~\eqref{eq:vis-w-bipath}, we use the warp-supervision~\eqref{eq:warp-supervision}, found as a pairwise constraint in our consistency graph (Fig.~\ref{fig:pair-wise}). 
Benefiting from the strong and direct supervision provided by the synthetic flow $W$, the warp-supervision term increases convergence speed and helps in driving the network towards higher accuracy.
Further, by the direct regression loss against the flow $W$, which is smooth by construction, it also acts as a smoothness constraint.  
On the other hand, through the \w loss~\eqref{eq:vis-w-bipath}, the network learns the realistic motion patterns and appearance changes present between real images $(I,J)$. As a result, both loss terms are mutually beneficial. From a practical perspective, the warp-supervision loss can be integrated at a low computational and memory cost, since the backbone feature extraction for the three images $I,I',J$ can be shared between the two loss terms.

\parsection{Adaptive loss balancing} 
Our final unsupervised objective combines the losses~\eqref{eq:vis-w-bipath} and~\eqref{eq:warp-supervision} as $L = L_{\text{W-vis}} + \lambda L_{\text{warp}}$. This raises the question of how to set the trade-off $\lambda$. Instead of resorting to manual tuning, we eliminate this hyper-parameter by automatically balancing the weights over each training batch as $\lambda = L_{\text{W-vis}} / L_{\text{warp}}$. Since $\lambda$ is a weighting factor, we do not backpropagate gradients through it.

\subsection{Sampling warps $\textbf{W}$}
\label{sec:our-transfo}

The key element of our warp consistency objective is the sampled warp $W$. During training, we randomly sample it from a distribution $W \sim p_W$, which we need to design. As discussed in Sec.~\ref{sec:semicycle},  the \w loss~\eqref{eq:vis-w-bipath} approaches the forward-backward consistency loss~\eqref{eq:forward-backward} when the magnitude of the warps decreases $\|W\| \rightarrow 0$. Exclusively sampling too small warps $W \approx 0$ therefore risks biasing the prediction towards zero. On the other hand, too large warps would render the estimation of $\widehat{F}_{I'\rightarrow J}$ challenging and introduce unnecessary invalid image regions. As a rough guide, the distribution $p_W$ should yield warps of similar magnitude as the real transformations $\|F_{J\rightarrow I}\|$, thus giving similar impact to all three terms in~\eqref{eq:vis-w-bipath}. Fortunately, as analyzed in the supplementary Sec.~\ref{sec-sup:transfo-analysis}, our approach is not sensitive to these settings as long as they are within reasonable bounds. 

We construct $W$ by sampling homography, Thin-plate Spline (TPS) and affine-TPS transformations randomly, following a procedure similar to previous approaches using warp-supervision~\cite{Rocco2017a}.
\bp{(i)} Homographies are constructed by randomly translating the four image corner locations. The magnitudes of the translations are chosen independently through Gaussian or uniform sampling, with standard-deviation or range equal to $\sigma_H$.
\bp{(ii)} For TPS, we randomly jitter a $3 \times 3$ grid of control points by independently translating each point. We use the same standard deviation or range $\sigma_H$ as for our homographies. 
\bp{(iii)} To generate larger scale and rotation changes, we also compose affine and TPS. We first sample affine transformations by selecting scale, rotation, translation and shearing parameters according to a Gaussian or uniform sampling. The TPS transform is then sampled as explained above and the final synthetic flow $W$ is a composition of both flows. 

To make the warps $W$ harder, we optionally also compose the flow obtained from (i), (ii) and (iii) with randomly sampled elastic transforms.  Specifically, we generate an elastic deformation motion field, as described in~\cite{Simard2003} and apply it in multiple regions selected randomly. 
Elastic deformations drive the network to be more accurate to small details.
Detailed settings are provided in the supplementary Sec.~\ref{sec-sup:glunet},~\ref{sec-sup:RANSAC-flow} and~\ref{sec-sup:semanticglunet}.

\section{Experiments}
\label{sec:exp-val}

We evaluate our unsupervised learning approach for three dense matching networks and two tasks, namely GLU-Net~\cite{GLUNet} and RANSAC-Flow~\cite{RANSAC-flow} for geometric matching, and SemanticGLU-Net~\cite{GLUNet} for semantic matching. We extensively analyze our method and compare it to earlier unsupervised objectives, defining a new state-of-the-art on multiple datasets. 
Further results, analysis, visualizations and implementation details are provided in the supplementary.

\begin{figure}[t]
\centering%
\vspace{-4mm}
\newcommand{\wid}{0.98\columnwidth}%
\includegraphics*[width=\wid, trim={0 0 12 0}]{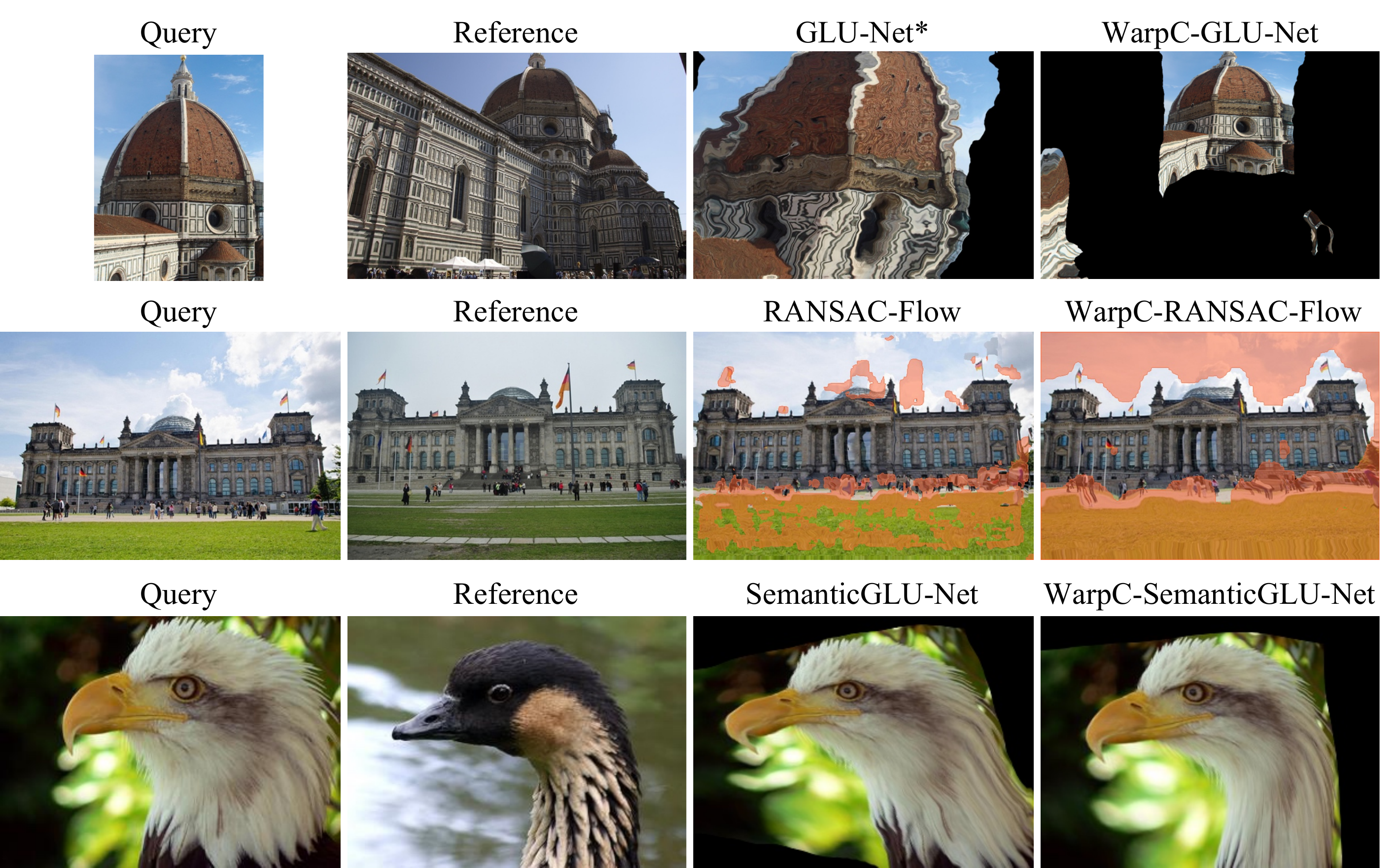} 
\vspace{0mm}\caption{Warped query according to baseline network and our approach. In the middle row, we visualize the predicted mask by RANSAC-Flow based networks in red (unmatchable regions).
}\vspace{-4mm}
\label{fig:quali}
\end{figure}

\subsection{Method analysis}
\label{sec:method-analysis}

We first perform a comprehensive analysis of our approach. We adopt GLU-Net~\cite{GLUNet} as our base architecture. It is a 4-level pyramidal network operating at two image resolutions to estimate dense flow fields. 

\parsection{Experimental set-up for GLU-Net} 
We slightly simplify the GLU-Net~\cite{GLUNet} architecture by replacing the dense decoder connections with standard residual blocks, which drastically reduces the number of network parameters with negligible impact on performance.
As in~\cite{GLUNet}, the feature extraction network is set to a VGG-16~\cite{Chatfield14} with ImageNet pre-trained weights. We train the rest of the architecture from scratch in two stages. We first train GLU-Net using our unsupervised objective, described in Sec.~\ref{sec:our-loss}, but without the visibility mask $\widehat{V}$. 
As a second stage, we add the visibility mask and employ stronger warps $W$, with elastic transforms. For both stages, we use the training split of the MegaDepth dataset~\cite{megadepth}, which comprises diverse internet images of 196 different world monuments.

\parsection{Datasets and metrics} We evaluate on standard datasets with sparse ground-truth, namely \textbf{RobotCar}~\cite{RobotCarDatasetIJRR, RobotCar} and \textbf{MegaDepth}~\cite{megadepth}. For the latter, we use the test split of ~\cite{RANSAC-flow}, which consists of 19 scenes not seen during training. 
Images in Robotcar depict outdoor road scenes and are particularly challenging due to their many textureless regions. MegaDepth images show extreme view-point and appearance variations.
In line with~\cite{RANSAC-flow}, we use the Percentage of Correct Keypoints at a given pixel threshold $T$ (PCK-$T$) as the evaluation metric (in \%). 
We also employ the 59 sequences of the homography dataset \textbf{HPatches}~\cite{Lenc}. We evaluate with the Average End-Point-Error (AEPE) and PCK. 

\parsection{Warp consistency graph losses} In Tab.~\ref{tab:graph-analysis} we empirically compare the constraints extracted from our warp consistency graph (Sec.~\ref{sec:cyclic-consistency-graph}). All networks are trained with only the first stage, on the same synthetic transformations $W$. Since we observed it to give a general improvement, we stop gradients through the flow used for warping (but not the flow that is warped).
The \ipj (II) and \ji (III) losses lead to a degenerate solution and a large predicted bias respectively, which explains the very poor performance of the networks. 
The cycle loss (V) obtains much better results but does not reach the performance of the \w constraint (IV). We only show the cycle starting from $I'$ here (V), since it performs best among all cycle losses (see suppl.~\ref{subsec-sup:analysis}). 
While the warp-supervision loss (I) results in a better accuracy on all datasets (PCK-1 and PCK-5 for HPatches), it is significantly less robust to large view-point changes than the \w objective (IV), as evidenced by results in PCK-10 and AEPE. These two losses have complementary behaviors and combining them (VIII) leads to a significant gain in both accuracy and robustness. Combining the warp-supervison loss (I) with \ipj (II) in (VI) or with \ji (III) in (VII) instead results in drastically lower performance than (VIII). The cycle loss (V) with the warp-supervision (I) in (IX) is also slightly worse.

\parsection{Ablation study} In Tab.~\ref{tab:ablation-study} we analyze the key components of our approach. We first show the importance of not back-propagating gradients in the warp operation.
Adding the warp-supervision objective with constant weights of $\lambda = 1$ increases both the network's accuracy and robustness for all datasets. Further using adaptive loss balancing (Sec.~\ref{sec:our-loss}) provides a significant improvement in accuracy (PCK-1) for MegaDepth with only minor loss on other thresholds.  
Including our visibility mask $\widehat{V}$ in the second training stage drastically improves all metrics for all datasets. Finally, further sampling harder transformations results in better accuracy, particularly for PCK-1 on MegaDepth. We therefore use this as our standard setting in the following experiments, where we denote it as \textbf{WarpC}.

\parsection{Comparison to alternative losses} Finally, in Tab.~\ref{tab:comparison-loss} we compare and combine our proposed objective with alternative losses. The census loss~\cite{Meister2017} (I), popular in optical flow, does not have  sufficient invariance to appearance changes and thus leads to poor results on geometric matching datasets. The SSIM loss~\cite{WangBSS04} (II) is more robust to the large appearance variations present in MegaDepth. Further combining SSIM with the forward-backward consistency loss (III) leads to a small improvement. Compared to SSIM (III) on MegaDepth, our WarpC approach (VI) achieves superior PCK-5 (+7.8\%) and PCK-10 (+10.2\%) at the cost of a slight reduction in sub-pixel accuracy. Furthermore, our approach demonstrates superior generalization capabilities by outperforming all other alternatives on the RobotCar and HPatches datasets.
For completeness, we also evaluate the combination (VII) of our loss with the photometric SSIM loss. This leads to improved PCK-1 on MegaDepth but degrades other metrics compared to WarpC (VI). Nevertheless, adding WarpC significantly improves upon SSIM (II) for all thresholds and datasets. 
Moreover, combining the warp-supervision (IV) with the forward-backward loss in (V) leads to an improvement compared to (IV). It is however significantly worse than combining the warp-supervision with our \w loss in (VI), which can be seen as a generalization of the forward-backward loss. 
Finally, we compare with using the sparse ground-truth supervision provided by SfM reconstruction of the MegaDepth training images. 
Interestingly, training the dense prediction network from scratch with solely sparse annotations (VIII) leads to inferior performance compared to our unsupervised objective (VI). 
Lastly, we fine-tune (IX) our proposed network (VI) with sparse annotations. While this leads to a moderate gain on MegaDepth, it comes at the cost of worse generalization properties on RobotCar and HPatches.

\begin{table}[t]
\centering
\resizebox{\columnwidth}{!}{%
\begin{tabular}{l@{~}|@{~}l@{~}c@{~~}c@{~~}c|c@{~~}c@{~~}c|c@{~~}c} \toprule
& & \multicolumn{3}{c}{\textbf{MegaDepth}} & \multicolumn{3}{c}{\textbf{RobotCar}} & \multicolumn{2}{c}{\textbf{HPatches}} \\
& & PCK-1  & PCK-5 & PCK-10  & PCK-1 & PCK-5 & PCK-10  & AEPE & PCK-5 \\ \midrule
I & Warp-supervision~\eqref{eq:warp-supervision} & \textbf{35.98} & 57.21 & 63.88 & \textbf{2.43} & 33.63 & 54.50 & 28.50 & \textbf{76.76}  \\ 
II & \ipj~\eqref{eq:semi-cycle-flow-a} & 0.00 & 0.05 & 0.21 & 0.00 & 0.00 & 0.13 & 370.80 & 0.01 \\
III & \ji~\eqref{eq:semi-cycle-flow-b},\eqref{eq:ji-bipath}  & 0.00 & 0.06 & 0.21 & 0.00 & 0.05 & 0.21 & 162.50 & 0.04 \\
IV & \w~\eqref{eq:semi-cycle-flow-c},\eqref{eq:w-bipath} &  29.55 & \textbf{67.70} & \textbf{74.42} & 2.25 & \textbf{33.88} & \textbf{55.38} & \textbf{26.13} & 70.51 \\ 
V & $I'$-cycle & 25.04 & 64.44 & 71.75 & 2.19 & 32.79 & 54.55 & 27.51 & 66.16 \\ 
\midrule
VI & \ipj + warp-sup. & 0.00 & 0.11 & 0.45 & 0.01 & 0.35 & 1.52 & 255.40 & 0.02 \\
VII & \ji + warp-sup. &  33.72 & 61.10 & 67.44 & 2.26 & 34.06 & 55.07 & 28.91 & 71.52 \\ 
VIII & \w + warp-sup.  & \textbf{43.47} & \textbf{69.90} & \textbf{75.23} & 2.49 & 35.28 & 56.45 & \textbf{22.83} & \textbf{78.60} \\ 
IX & $I'$-cycle + warp-sup. & 42.11 & 68.84 & 74.28 & \textbf{2.52} & \textbf{35.75} & \textbf{56.96} & 24.16 & 78.58  \\ 
\bottomrule
\end{tabular}%
}\vspace{1mm}
\caption{Analysis of warp consistency graph losses (Sec.~\ref{sec:cyclic-consistency-graph}-\ref{sec:semicycle}).}
\label{tab:graph-analysis}\vspace{-2mm}
\end{table}

\begin{table}[t]
\centering
\resizebox{\columnwidth}{!}{%
\begin{tabular}{l@{~}c@{~~}c@{~~}c|c@{~~}c@{~~}c|c@{~~}c} \toprule
& \multicolumn{3}{c}{\textbf{MegaDepth}} & \multicolumn{3}{c}{\textbf{RobotCar}} & \multicolumn{2}{c}{\textbf{HPatches}} \\
& PCK-1  & PCK-5 & PCK-10  & PCK-1 & PCK-5 & PCK-10  & AEPE & PCK-5 \\ \midrule
\w~\eqref{eq:w-bipath}, grad in warp  & 20.06 & 58.57 & 67.83 & 2.04 & 31.70 & 53.57 & 29.37 & 60.40 \\ 
\w~\eqref{eq:w-bipath} & 29.55 & 67.70 & 74.42 & 2.25 & 33.88 & 55.38 & 26.13 & 70.51 \\ 
+ warp-supervision~\eqref{eq:warp-supervision} & 39.66 &  70.38 & 76.06 & 2.45 & 34.92  & 56.37 & 22.52 & 78.65 \\ 
+ adaptive loss balancing & 43.47 & 69.90 & 75.23 & 2.49 & 35.28 & 56.45 & 22.83 & 78.60 \\ 
+ visibility mask $\widehat{V}$~\eqref{eq:vis-w-bipath} & 48.86 & 77.58 & 82.27 & \textbf{2.51} &  35.78 & 57.19 & 21.63 & 82.55  \\ 
+ harder warps $W$ & \textbf{50.61} & \textbf{78.61} & \textbf{82.94} & \textbf{2.51} & \textbf{35.92} & \textbf{57.44} & \textbf{21.00} & \textbf{83.24} \\
\bottomrule
\end{tabular}%
}\vspace{1mm}
\caption{Ablation study by incrementally adding each component. }\label{tab:ablation-study}\vspace{-2mm}
\end{table}

\begin{table}[t]
\centering
\resizebox{\columnwidth}{!}{%
\begin{tabular}{l@{~}|@{~}l@{~}c@{~~}c@{~~}c|c@{~~}c@{~~}c|c@{~~}c} \toprule
& & \multicolumn{3}{c}{\textbf{MegaDepth}} & \multicolumn{3}{c}{\textbf{RobotCar}} & \multicolumn{2}{c}{\textbf{HPatches}} \\
& & PCK-1  & PCK-5 & PCK-10  & PCK-1 & PCK-5 & PCK-10  & AEPE & PCK-5 \\ \midrule
I & Census~\eqref{eq:photo-loss} & 33.49 & 58.44 & 61.42 & 1.85 & 28.25 & 48.37 & 59.85 & 48.15 \\
II & SSIM~\eqref{eq:photo-loss} & 51.93 & 69.58 & 71.58 & 2.18 & 31.48 & 51.65 & 38.62 & 62.61 \\
III & SSIM~\eqref{eq:photo-loss} + f-b~\eqref{eq:forward-backward} & 52.59 & 70.78 & 72.78 & 2.12 & 31.86 & 52.13 & 35.79 & 64.48  \\

IV & Warp-superv.~\eqref{eq:warp-supervision} & 38.50 & 59.60 & 66.21 & 2.36 &  33.28 & 54.47 & 25.04 & 78.60 \\ %
V & Warp-superv. + f-b~\eqref{eq:forward-backward} & 45.62 & 71.36 & 75.92 & 2.50 & \textbf{36.04} & 57.13 & 23.10 & 79.64 \\

VI & \textbf{WarpC} (\eqref{eq:vis-w-bipath} + \eqref{eq:warp-supervision}) &  50.61 & \textbf{78.61} & \textbf{82.94} & \textbf{2.51} & 35.92 & \textbf{57.44} & \textbf{21.00} & \textbf{83.24} \\
VII & \textbf{WarpC} + SSIM
& \textbf{54.92} & 75.65 & 78.04 &   2.43 & 35.01 & 56.44 & 26.01 & 74.64 \\

\midrule
VIII & Supervised & 38.83 & 72.42 & 77.34 & 2.15 & 32.52 & 53.88 & 37.91 &  56.15\\
IX & \textbf{WarpC} + Sup.\ ft. & \textbf{56.68} & \textbf{81.33} & \textbf{84.76} & \textbf{2.41} & \textbf{34.67} & \textbf{55.89} & \textbf{22.78} & \textbf{78.19} \\ 
\bottomrule
\end{tabular}%
}\vspace{1mm}
\caption{Analysis and comparison of learning objectives.}\label{tab:comparison-loss}\vspace{-4mm}
\end{table}

\subsection{Geometric matching}

Here, we train the recent GLU-Net~\cite{GLUNet} and RANSAC-Flow~\cite{RANSAC-flow} architectures with our unsupervised learning approach and compare them against state-of-the-art dense geometric matching methods.

\parsection{Experimental set-up for GLU-Net} We follow the training procedure explained in Sec.~\ref{sec:method-analysis} and refer to the resulting model as WarpC-GLU-Net.
The original GLU-Net~\cite{GLUNet} is trained using solely the warp-supervision~\eqref{eq:warp-supervision} on a different training set. For fair comparison, we also report results of our altered GLU-Net architecture when trained on MegaDepth with our warp distribution. This corresponds to setting (IV) in Tab.~\ref{tab:comparison-loss}, which we here call GLU-Net*.

\parsection{Experimental set-up for RANSAC-Flow} We additionally use our unsupervised strategy to train RANSAC-Flow~\cite{RANSAC-flow}. 
In the original work~\cite{RANSAC-flow}, the network is trained on MegaDepth~\cite{megadepth} image pairs that are coarsely pre-aligned using feature matching and Ransac. Training is separated into three stages. First, the network is trained using the SSIM loss~\eqref{eq:photo-loss}, which is further combined with the forward-backward consistency loss~\eqref{eq:forward-backward} in the second stage. In the last stage, a matchability mask is also trained, by weighting the previous losses with the predicted mask and including a mask regularization term. 
For our WarpC-RANSAC-Flow, we also follow a three-step training using the same training pairs. As for the WarpC-GLU-Net training, we add our visibility mask $\widehat{V}$ in the second training stage. In the third stage, we train the matchability mask by simply replacing $\widehat{V}$ in \eqref{eq:vis-w-bipath} with the predicted mask, and adding the same mask regularizer as in RANSAC-Flow. 

\begin{table}[t]
\centering
\vspace{-4mm}
\resizebox{\columnwidth}{!}{%
\begin{tabular}{lcccc|cccc} \toprule
& \multicolumn{4}{c}{\textbf{MegaDepth}~\cite{megadepth}} & \multicolumn{4}{c}{\textbf{RobotCar}~\cite{RobotCarDatasetIJRR, RobotCar}} \\
& PCK-1  & PCK-3  & PCK-5 & PCK-10 & PCK-1  & PCK-3 & PCK-5 & PCK-10  \\ \midrule
SIFT-Flow~\cite{LiuYT11} & 8.70 & 12.19 & 13.30 & - & 1.12 & 8.13 & 16.45 & - \\
NCNet~\cite{Rocco2018b} & 1.98 & 14.47 & 32.80 & - & 0.81 & 7.13 & 16.93 & - \\
DGC-Net~\cite{Melekhov2019} & 3.55 & 20.33 & 32.28 & - & 1.19 & 9.35 & 20.17 & - \\
GLU-Net~\cite{GLUNet, GOCor} & 21.58 & 52.18 & 61.78 & 69.81 & 2.30 & 17.15 & 33.87 & 55.67 \\
GLU-Net-GOCor~\cite{GOCor} &  37.28 &  61.18  & 68.08 & 74.39 & 2.31 & 17.62 & 35.18 & 57.26 \\ 
\midrule

GLU-Net* & 38.50 & 59.60 & 60.33 & 66.21 &  2.36 & 17.18 & 33.28 & 54.47  \\
\textbf{WarpC-GLU-Net} & 50.61 & 73.80 & 78.61 & 82.94 & \textbf{2.51} & \textbf{18.59} & \textbf{35.92} & \textbf{57.44} \\
\midrule

RANSAC-Flow~\cite{RANSAC-flow} &  52.60 & 83.46 & 86.80 & 88.80 & 2.09 & 15.94 & 31.61 & 53.06 \\
\textbf{WarpC-RANSAC-Flow} & 
\textbf{53.77} & \textbf{84.23} & \textbf{88.18} & \textbf{90.53} &  2.29 & 17.23 & 34.42  & 56.12 \\
\bottomrule
\end{tabular}%
}\vspace{1mm}\caption{State-of-the-art comparison for geometric matching.}\vspace{-6mm}
\label{tab:megadepth}
\end{table}

\parsection{Results} In Tab.~\ref{tab:megadepth}, we report results on MegaDepth and RobotCar.  Note that we only compare to methods that do not finetune on the test set. 
Our approach WarpC-GLU-Net outperforms the original GLU-Net and baseline GLU-Net* by a large margin at all PCK thresholds. Our proposed unsupervised objective enables the network to handle the large and complex 3D motions present in real image pairs, as evidenced in Fig.~\ref{fig:quali}, top. 
Our unsupervised approach WarpC-RANSAC-Flow also achieves a substantial improvement compared to RANSAC-Flow. Importantly, WarpC-RANSAC-Flow shows much better generalization capabilities on RobotCar. The poorer generalization of photometric-based objectives, such as SSIM~\cite{WangBSS04} here, further supports our findings in Sec.~\ref{sec:method-analysis}. Interestingly, training the matchability branch of RANSAC-Flow with our objective results in drastically more accurate mask predictions. This is visualized in Fig.~\ref{fig:quali}, middle, where our approach WarpC-RANSAC-Flow effectively identifies unreliable matching regions such as the sky (in red), whereas RANSAC-Flow, trained with the SSIM loss, is incapable of discarding the sky and field as unreliable.

\subsection{Semantic matching}

Finally, we evaluate our approach for the task of semantic matching by training SemanticGLU-Net~\cite{GLUNet}, a version of GLU-Net specifically designed for semantic images, which includes multi-resolution features and NC-Net~\cite{Rocco2018b}. 

\parsection{Experimental set-up} Following~\cite{Rocco2018a, ArbiconNet}, we only fine-tune a pre-trained network on semantic correspondence data. Specifically, we start from the SemanticGLU-Net weights provided by the authors, which are trained with warp-supervision without using any correspondences from flow annotations.
We finetune this network on the PF-PASCAL training set~\cite{PFPascal}, which consists of 20 object categories, using our unsupervised loss (Sec.~\ref{sec:our-loss}).

\parsection{Datasets and metrics} We first evaluate on the test set of \textbf{PF-Pascal}~\cite{PFPascal}. In line with~\cite{SCNet}, we report the PCK with a pixel threshold equal to $\alpha \cdot \max(h_{q}, w_{q})$, where $h_{q}$ and $w_{q}$ are the dimensions of the query image and $\alpha = (0.05, 0.1)$. 
To demonstrate generalization capabilities, we also validate our trained model on \textbf{TSS}~\cite{Taniai2016}, which provides dense flow field annotations for the foreground object in each pair. We report the PCK for $\alpha=0.05$.
We also provide results on PF-Willow~\cite{PFWillow} and SPair-71K~\cite{spair} in suppl.~\ref{subsec-sup:semantic-results}.

\begin{table}[t]
\centering
\vspace{-4mm}
\resizebox{0.48\textwidth}{!}{%
\begin{tabular}{ll|c@{~~}c@{~~}c@{~~}c|c@{~~}c}
\toprule
 &  & \multicolumn{4}{c}{\textbf{TSS}~\cite{Taniai2016}} & \multicolumn{2}{c}{\textbf{PF-Pascal}~\cite{PFPascal}} \\
Methods  & Features &  FG3DCar & JODS & Pascal & Avg. & $\alpha\!=\!0.05 $ & $\alpha\!=\!0.1$\\ \midrule
CNNGeo~\cite{Rocco2018a} & ResNet-101 & 90.1 & 76.4 & 56.3 & 74.4 & 41.0 & 69.5 \\ 
WeakAlign~\cite{Rocco2018a} & ResNet-101 & 90.3  & 76.4 & 56.5  & 74.4 & 49.0 & 75.8\\
RTNs~\cite{Kim2018} & ResNet-101 & 90.1 & 78.2 & 63.3 & 77.2  & 55.2 & 75.9 \\
PARN ~\cite{Jeon} & ResNet-101 & 89.5  & 75.9 & 71.2  & 78.8  & - & - \\
NC-Net~\cite{Rocco2018b} &  ResNet-101 & 94.5 & 81.4 & 57.1 & 77.7  & - & 78.9\\
DCCNet~\cite{DCCNet}  & ResNet-101 & 93.5  & 82.6 & 57.6  & 77.9  & 55.6 & \textbf{82.3}\\
DHPF~\cite{MinLPC20} & ResNet-101 &  - & - & - & - & 56.1 & 82.1  \\
SAM-Net~\cite{Kim2019} & VGG-19  & 96.1 & 82.2 & 67.2 & 81.8  & 60.1 & 80.2\\
GLU-Net~\cite{GLUNet} & VGG-16   &   93.2    &   73.3  &   71.1    &    79.2  & 42.2 & 69.1 \\ 
GLU-Net-GOCor~\cite{GOCor} & VGG-16  & 94.6 &  77.9 &  77.7 & 83.4 & 36.6 & 56.8 \\ 
SemanticGLU-Net~\cite{GLUNet} & VGG-16 & 94.4 & 75.5  & 78.3 & 82.8 & 46.0 & 70.6 \\ 
\textbf{WarpC-SemanticGLU-Net} & VGG-16 & \textbf{97.1} & \textbf{84.7} & \textbf{79.7} & \textbf{87.2} & \textbf{62.1} & 81.7 \\ 
\bottomrule
\end{tabular}%
}\vspace{1mm}
\caption{State-of-the-art comparison for semantic matching.}\vspace{-5mm}
\label{tab:TSS}
\end{table}

\parsection{Results} Results are reported in Tab.~\ref{tab:TSS}. Our approach WarpC-SemanticGLU-Net sets a new state-of-the-art on TSS by obtaining a remarkable improvement compared to previous works. On the PF-Pascal dataset, our method ranks first for the small threshold $\alpha=0.05$ with a substantial $2\%$ increase compared to second best method. It obtains marginally lower PCK ($0.6\%$) than DCCNet~\cite{DCCNet} for $\alpha=0.1$, but the later approach employs a much deeper feature backbone, beneficial on semantic images. 
Nevertheless, our unsupervised fine-tuning provides 16\% and 11.1\% gain, for each threshold respectively, over the baseline, demonstrating that our objective effectively copes with the radical appearance changes encountered in the semantic matching task. 
A visual example is  shown in Fig.~\ref{fig:quali}~bottom. 

\section{Conclusion}

We propose an unsupervised learning objective for dense correspondences, particularly suitable for scenarios with large changes in appearance and geometry. From a real image pair, we construct an image triplet and design a regression loss based on the flow-constraints existing between the triplet. When integrated into three recent dense correspondence networks, our approach outperforms state-of-the-art for multiple geometric and semantic matching datasets.

\parsection{Acknowledgements}
This work was supported by the ETH Z\"urich Fund (OK), a Huawei Gift, the ETH Future Computing Laboratory (EFCL) financed by a gift from Huawei Technologies, Amazon AWS, and an Nvidia GPU grant.

{\small
\bibliographystyle{ieee_fullname}
\bibliography{biblio}

\begin{thebibliography}{10}\itemsep=-1pt

\bibitem{Lenc}
Vassileios Balntas, Karel Lenc, Andrea Vedaldi, and Krystian Mikolajczyk.
\newblock Hpatches: {A} benchmark and evaluation of handcrafted and learned
  local descriptors.
\newblock In {\em 2017 {IEEE} Conference on Computer Vision and Pattern
  Recognition, {CVPR} 2017, Honolulu, HI, USA, July 21-26, 2017}, pages
  3852--3861, 2017.

\bibitem{BourdevM09}
Lubomir~D. Bourdev and Jitendra Malik.
\newblock Poselets: Body part detectors trained using 3d human pose
  annotations.
\newblock In {\em ICCV}, pages 1365--1372. IEEE Computer Society, 2009.

\bibitem{info11020125}
Alexander Buslaev, Vladimir~I. Iglovikov, Eugene Khvedchenya, Alex Parinov,
  Mikhail Druzhinin, and Alexandr~A. Kalinin.
\newblock Albumentations: Fast and flexible image augmentations.
\newblock {\em Information}, 11(2), 2020.

\bibitem{Chatfield14}
K. Chatfield, K. Simonyan, A. Vedaldi, and A. Zisserman.
\newblock Return of the devil in the details: Delving deep into convolutional
  nets.
\newblock In {\em BMVC}, 2014.

\bibitem{ArbiconNet}
Jianchun Chen, Lingjing Wang, Xiang Li, and Yi Fang.
\newblock Arbicon-net: Arbitrary continuous geometric transformation networks
  for image registration.
\newblock In {\em Advances in Neural Information Processing Systems 32: Annual
  Conference on Neural Information Processing Systems 2019, NeurIPS 2019,
  December 8-14, 2019, Vancouver, BC, Canada}, pages 3410--3420, 2019.

\bibitem{Cordts2016}
Marius Cordts, Mohamed Omran, Sebastian Ramos, Timo Rehfeld, Markus Enzweiler,
  Rodrigo Benenson, Uwe Franke, Stefan Roth, and Bernt Schiele.
\newblock The cityscapes dataset for semantic urban scene understanding.
\newblock In {\em Proc. of the IEEE Conference on Computer Vision and Pattern
  Recognition (CVPR)}, 2016.

\bibitem{superpoint}
Daniel DeTone, Tomasz Malisiewicz, and Andrew Rabinovich.
\newblock Superpoint: Self-supervised interest point detection and description.
\newblock In {\em 2018 {IEEE} Conference on Computer Vision and Pattern
  Recognition Workshops, {CVPR} Workshops 2018, Salt Lake City, UT, USA, June
  18-22, 2018}, pages 224--236, 2018.

\bibitem{Dosovitskiy2015}
Alexey Dosovitskiy, Philipp Fischer, Eddy Ilg, Philip H{\"{a}}usser, Caner
  Hazirbas, Vladimir Golkov, Patrick van~der Smagt, Daniel Cremers, and Thomas
  Brox.
\newblock Flownet: Learning optical flow with convolutional networks.
\newblock In {\em 2015 {IEEE} International Conference on Computer Vision,
  {ICCV} 2015, Santiago, Chile, December 7-13, 2015}, pages 2758--2766, 2015.

\bibitem{DusmanuRPPSTS19}
Mihai Dusmanu, Ignacio Rocco, Tom{\'{a}}s Pajdla, Marc Pollefeys, Josef Sivic,
  Akihiko Torii, and Torsten Sattler.
\newblock D2-net: {A} trainable {CNN} for joint description and detection of
  local features.
\newblock In {\em {IEEE} Conference on Computer Vision and Pattern Recognition,
  {CVPR} 2019, Long Beach, CA, USA, June 16-20, 2019}, pages 8092--8101, 2019.

\bibitem{DwibediATSZ19}
Debidatta Dwibedi, Yusuf Aytar, Jonathan Tompson, Pierre Sermanet, and Andrew
  Zisserman.
\newblock Temporal cycle-consistency learning.
\newblock In {\em {IEEE} Conference on Computer Vision and Pattern Recognition,
  {CVPR} 2019, Long Beach, CA, USA, June 16-20, 2019}, pages 1801--1810, 2019.

\bibitem{EveringhamGWWZ10}
Mark Everingham, Luc~Van Gool, Christopher K.~I. Williams, John~M. Winn, and
  Andrew Zisserman.
\newblock The pascal visual object classes {(VOC)} challenge.
\newblock {\em Int. J. Comput. Vis.}, 88(2):303--338, 2010.

\bibitem{HaCohenSGL11}
Yoav HaCohen, Eli Shechtman, Dan~B. Goldman, and Dani Lischinski.
\newblock Non-rigid dense correspondence with applications for image
  enhancement.
\newblock {\em {ACM} Trans. Graph.}, 30(4):70, 2011.

\bibitem{PFWillow}
Bumsub Ham, Minsu Cho, Cordelia Schmid, and Jean Ponce.
\newblock Proposal flow.
\newblock In {\em Proceedings of the IEEE Conference on Computer Vision and
  Pattern Recognition}, 2016.

\bibitem{PFPascal}
Bumsub Ham, Minsu Cho, Cordelia Schmid, and Jean Ponce.
\newblock Proposal flow: Semantic correspondences from object proposals.
\newblock {\em {IEEE} Trans. Pattern Anal. Mach. Intell.}, 40(7):1711--1725,
  2018.

\bibitem{SCNet}
Kai Han, Rafael~S. Rezende, Bumsub Ham, Kwan{-}Yee~K. Wong, Minsu Cho, Cordelia
  Schmid, and Jean Ponce.
\newblock Scnet: Learning semantic correspondence.
\newblock In {\em {IEEE} International Conference on Computer Vision, {ICCV}
  2017, Venice, Italy, October 22-29, 2017}, pages 1849--1858, 2017.

\bibitem{MOCO}
Kaiming He, Haoqi Fan, Yuxin Wu, Saining Xie, and Ross~B. Girshick.
\newblock Momentum contrast for unsupervised visual representation learning.
\newblock In {\em 2020 {IEEE/CVF} Conference on Computer Vision and Pattern
  Recognition, {CVPR} 2020, Seattle, WA, USA, June 13-19, 2020}, pages
  9726--9735, 2020.

\bibitem{HeZRS15}
Kaiming He, Xiangyu Zhang, Shaoqing Ren, and Jian Sun.
\newblock Deep residual learning for image recognition.
\newblock {\em CoRR}, abs/1512.03385, 2015.

\bibitem{heinly2015_reconstructing_the_world}
Jared Heinly, Johannes~Lutz Sch\"{o}nberger, Enrique Dunn, and Jan-Michael
  Frahm.
\newblock {Reconstructing the World* in Six Days *(As Captured by the Yahoo 100
  Million Image Dataset)}.
\newblock In {\em Computer Vision and Pattern Recognition (CVPR)}, 2015.

\bibitem{Horn1981}
Berthold K.~P. Horn and Brian~G. Schunck.
\newblock "determining optical flow": {A} retrospective.
\newblock {\em Artif. Intell.}, 59(1-2):81--87, 1993.

\bibitem{Huang2017}
Gao Huang, Zhuang Liu, Laurens van~der Maaten, and Kilian~Q. Weinberger.
\newblock Densely connected convolutional networks.
\newblock In {\em 2017 {IEEE} Conference on Computer Vision and Pattern
  Recognition, {CVPR} 2017, Honolulu, HI, USA, July 21-26, 2017}, pages
  2261--2269. {IEEE} Computer Society, 2017.

\bibitem{DCCNet}
Shuaiyi Huang, Qiuyue Wang, Songyang Zhang, Shipeng Yan, and Xuming He.
\newblock Dynamic context correspondence network for semantic alignment.
\newblock In {\em 2019 {IEEE/CVF} International Conference on Computer Vision,
  {ICCV} 2019, Seoul, Korea (South), October 27 - November 2, 2019}, pages
  2010--2019. {IEEE}, 2019.

\bibitem{Hui2018}
Tak{-}Wai Hui, Xiaoou Tang, and Chen~Change Loy.
\newblock Liteflownet: {A} lightweight convolutional neural network for optical
  flow estimation.
\newblock In {\em 2018 {IEEE} Conference on Computer Vision and Pattern
  Recognition, {CVPR} 2018, Salt Lake City, UT, USA, June 18-22, 2018}, pages
  8981--8989, 2018.

\bibitem{Ignatov2017}
Andrey Ignatov, Nikolay Kobyshev, Radu Timofte, Kenneth Vanhoey, and Luc~Van
  Gool.
\newblock Dslr-quality photos on mobile devices with deep convolutional
  networks.
\newblock In {\em {IEEE} International Conference on Computer Vision, {ICCV}
  2017, Venice, Italy, October 22-29, 2017}, pages 3297--3305, 2017.

\bibitem{Ilg2017a}
Eddy Ilg, Nikolaus Mayer, Tonmoy Saikia, Margret Keuper, Alexey Dosovitskiy,
  and Thomas Brox.
\newblock Flownet 2.0: Evolution of optical flow estimation with deep networks.
\newblock In {\em 2017 {IEEE} Conference on Computer Vision and Pattern
  Recognition, {CVPR} 2017, Honolulu, HI, USA, July 21-26, 2017}, pages
  1647--1655. {IEEE} Computer Society, 2017.

\bibitem{JabriOE20}
Allan Jabri, Andrew Owens, and Alexei~A. Efros.
\newblock Space-time correspondence as a contrastive random walk.
\newblock In {\em Advances in Neural Information Processing Systems 33: Annual
  Conference on Neural Information Processing Systems 2020, NeurIPS 2020,
  December 6-12, 2020, virtual}, 2020.

\bibitem{MFOccFlow}
Joel Janai, Fatma G{\"{u}}ney, Anurag Ranjan, Michael~J. Black, and Andreas
  Geiger.
\newblock Unsupervised learning of multi-frame optical flow with occlusions.
\newblock In {\em Computer Vision - {ECCV} 2018 - 15th European Conference,
  Munich, Germany, September 8-14, 2018, Proceedings, Part {XVI}}, pages
  713--731, 2018.

\bibitem{Jeon}
Sangryul Jeon, Seungryong Kim, Dongbo Min, and Kwanghoon Sohn.
\newblock {PARN:} pyramidal affine regression networks for dense semantic
  correspondence.
\newblock In {\em Computer Vision - {ECCV} 2018 - 15th European Conference,
  Munich, Germany, September 8-14, 2018, Proceedings, Part {VI}}, pages
  355--371, 2018.

\bibitem{Kim2018}
Seungryong Kim, Stephen Lin, Sangryul Jeon, Dongbo Min, and Kwanghoon Sohn.
\newblock Recurrent transformer networks for semantic correspondence.
\newblock In {\em Advances in Neural Information Processing Systems 31: Annual
  Conference on Neural Information Processing Systems 2018, NeurIPS 2018, 3-8
  December 2018, Montr{\'{e}}al, Canada.}, pages 6129--6139, 2018.

\bibitem{Kim2019}
Seungryong Kim, Dongbo Min, Somi Jeong, Sunok Kim, Sangryul Jeon, and Kwanghoon
  Sohn.
\newblock Semantic attribute matching networks.
\newblock In {\em {IEEE} Conference on Computer Vision and Pattern Recognition,
  {CVPR} 2019, Long Beach, CA, USA, June 16-20, 2019}, pages 12339--12348,
  2019.

\bibitem{adam}
Diederik~P. Kingma and Jimmy Ba.
\newblock Adam: {A} method for stochastic optimization.
\newblock In {\em 3rd International Conference on Learning Representations,
  {ICLR} 2015, San Diego, CA, USA, May 7-9, 2015, Conference Track
  Proceedings}, 2015.

\bibitem{Hinton2012}
Alex Krizhevsky, Ilya Sutskever, and Geoffrey~E. Hinton.
\newblock Imagenet classification with deep convolutional neural networks.
\newblock In {\em Advances in Neural Information Processing Systems 25: 26th
  Annual Conference on Neural Information Processing Systems 2012. Proceedings
  of a meeting held December 3-6, 2012, Lake Tahoe, Nevada, United States.},
  pages 1106--1114, 2012.

\bibitem{RobotCar}
M{\aa}ns Larsson, Erik Stenborg, Lars Hammarstrand, Marc Pollefeys, Torsten
  Sattler, and Fredrik Kahl.
\newblock A cross-season correspondence dataset for robust semantic
  segmentation.
\newblock In {\em {IEEE} Conference on Computer Vision and Pattern Recognition,
  {CVPR} 2019, Long Beach, CA, USA, June 16-20, 2019}, pages 9532--9542, 2019.

\bibitem{SFNet}
Junghyup Lee, Dohyung Kim, Jean Ponce, and Bumsub Ham.
\newblock Sfnet: Learning object-aware semantic correspondence.
\newblock In {\em {IEEE} Conference on Computer Vision and Pattern Recognition,
  {CVPR} 2019, Long Beach, CA, USA, June 16-20, 2019}, pages 2278--2287, 2019.

\bibitem{LiHLP20}
Xinghui Li, Kai Han, Shuda Li, and Victor Prisacariu.
\newblock Dual-resolution correspondence networks.
\newblock In {\em Advances in Neural Information Processing Systems 33: Annual
  Conference on Neural Information Processing Systems 2020, NeurIPS 2020,
  December 6-12, 2020, virtual}, 2020.

\bibitem{megadepth}
Zhengqi Li and Noah Snavely.
\newblock Megadepth: Learning single-view depth prediction from internet
  photos.
\newblock In {\em 2018 {IEEE} Conference on Computer Vision and Pattern
  Recognition, {CVPR} 2018, Salt Lake City, UT, USA, June 18-22, 2018}, pages
  2041--2050, 2018.

\bibitem{Liao2017}
Jing Liao, Yuan Yao, Lu Yuan, Gang Hua, and Sing~Bing Kang.
\newblock Visual attribute transfer through deep image analogy.
\newblock {\em ACM Trans. Graph.}, 36(4), July 2017.

\bibitem{LiuYT11}
Ce Liu, Jenny Yuen, and Antonio Torralba.
\newblock {SIFT} flow: Dense correspondence across scenes and its applications.
\newblock {\em {IEEE} Trans. Pattern Anal. Mach. Intell.}, 33(5):978--994,
  2011.

\bibitem{ARFlow}
Liang Liu, Jiangning Zhang, Ruifei He, Yong Liu, Yabiao Wang, Ying Tai, Donghao
  Luo, Chengjie Wang, Jilin Li, and Feiyue Huang.
\newblock Learning by analogy: Reliable supervision from transformations for
  unsupervised optical flow estimation.
\newblock In {\em 2020 {IEEE/CVF} Conference on Computer Vision and Pattern
  Recognition, {CVPR} 2020, Seattle, WA, USA, June 13-19, 2020}, pages
  6488--6497, 2020.

\bibitem{DDFlow}
Pengpeng Liu, Irwin King, Michael~R. Lyu, and Jia Xu.
\newblock Ddflow: Learning optical flow with unlabeled data distillation.
\newblock In {\em The Thirty-Third {AAAI} Conference on Artificial
  Intelligence, {AAAI} 2019, The Thirty-First Innovative Applications of
  Artificial Intelligence Conference, {IAAI} 2019, The Ninth {AAAI} Symposium
  on Educational Advances in Artificial Intelligence, {EAAI} 2019, Honolulu,
  Hawaii, USA, January 27 - February 1, 2019}, pages 8770--8777, 2019.

\bibitem{SefFlow}
Pengpeng Liu, Michael~R. Lyu, Irwin King, and Jia Xu.
\newblock Selflow: Self-supervised learning of optical flow.
\newblock In {\em {IEEE} Conference on Computer Vision and Pattern Recognition,
  {CVPR} 2019, Long Beach, CA, USA, June 16-20, 2019}, pages 4571--4580, 2019.

\bibitem{SIFT}
David~G. Lowe.
\newblock Distinctive image features from scale-invariant keypoints.
\newblock {\em Int. J. Comput. Vision}, 60(2):91–110, Nov. 2004.

\bibitem{RobotCarDatasetIJRR}
Will Maddern, Geoff Pascoe, Chris Linegar, and Paul Newman.
\newblock {1 Year, 1000km: The Oxford RobotCar Dataset}.
\newblock {\em The International Journal of Robotics Research (IJRR)},
  36(1):3--15, 2017.

\bibitem{Mayer2016ALD}
N. Mayer, Eddy Ilg, Philip H{\"a}usser, P. Fischer, D. Cremers, A. Dosovitskiy,
  and T. Brox.
\newblock A large dataset to train convolutional networks for disparity,
  optical flow, and scene flow estimation.
\newblock {\em 2016 IEEE Conference on Computer Vision and Pattern Recognition
  (CVPR)}, pages 4040--4048, 2016.

\bibitem{Meister2017}
Simon Meister, Junhwa Hur, and Stefan Roth.
\newblock {UnFlow}: Unsupervised learning of optical flow with a bidirectional
  census loss.
\newblock In {\em AAAI}, New Orleans, Louisiana, Feb. 2018.

\bibitem{Melekhov2019}
Iaroslav Melekhov, Aleksei Tiulpin, Torsten Sattler, Marc Pollefeys, Esa Rahtu,
  and Juho Kannala.
\newblock {DGC-Net}: Dense geometric correspondence network.
\newblock In {\em Proceedings of the IEEE Winter Conference on Applications of
  Computer Vision (WACV)}, 2019.

\bibitem{spair}
Juhong Min, Jongmin Lee, Jean Ponce, and Minsu Cho.
\newblock Spair-71k: {A} large-scale benchmark for semantic correspondence.
\newblock {\em CoRR}, abs/1908.10543, 2019.

\bibitem{MinLPC20}
Juhong Min, Jongmin Lee, Jean Ponce, and Minsu Cho.
\newblock Learning to compose hypercolumns for visual correspondence.
\newblock In {\em Computer Vision - {ECCV} 2020 - 16th European Conference,
  Glasgow, UK, August 23-28, 2020, Proceedings, Part {XV}}, pages 346--363,
  2020.

\bibitem{RenYNLYZ17}
Zhe Ren, Junchi Yan, Bingbing Ni, Bin Liu, Xiaokang Yang, and Hongyuan Zha.
\newblock Unsupervised deep learning for optical flow estimation.
\newblock In {\em Proceedings of the Thirty-First {AAAI} Conference on
  Artificial Intelligence, February 4-9, 2017, San Francisco, California,
  {USA}}, pages 1495--1501, 2017.

\bibitem{Rocco2017a}
Ignacio Rocco, Relja Arandjelovic, and Josef Sivic.
\newblock Convolutional neural network architecture for geometric matching.
\newblock In {\em 2017 {IEEE} Conference on Computer Vision and Pattern
  Recognition, {CVPR} 2017, Honolulu, HI, USA, July 21-26, 2017}, pages 39--48,
  2017.

\bibitem{Rocco2018a}
Ignacio Rocco, Relja Arandjelovic, and Josef Sivic.
\newblock End-to-end weakly-supervised semantic alignment.
\newblock In {\em 2018 {IEEE} Conference on Computer Vision and Pattern
  Recognition, {CVPR} 2018, Salt Lake City, UT, USA, June 18-22, 2018}, pages
  6917--6925, 2018.

\bibitem{RoccoAS20}
Ignacio Rocco, Relja Arandjelovic, and Josef Sivic.
\newblock Efficient neighbourhood consensus networks via submanifold sparse
  convolutions.
\newblock In {\em Computer Vision - {ECCV} 2020 - 16th European Conference,
  Glasgow, UK, August 23-28, 2020, Proceedings, Part {IX}}, pages 605--621,
  2020.

\bibitem{Rocco2018b}
Ignacio Rocco, Mircea Cimpoi, Relja Arandjelovic, Akihiko Torii, Tom{\'{a}}s
  Pajdla, and Josef Sivic.
\newblock Neighbourhood consensus networks.
\newblock In {\em Advances in Neural Information Processing Systems 31: Annual
  Conference on Neural Information Processing Systems 2018, NeurIPS 2018, 3-8
  December 2018, Montr{\'{e}}al, Canada.}, pages 1658--1669, 2018.

\bibitem{COLMAP}
Johannes~L. Sch{\"{o}}nberger and Jan{-}Michael Frahm.
\newblock Structure-from-motion revisited.
\newblock In {\em 2016 {IEEE} Conference on Computer Vision and Pattern
  Recognition, {CVPR} 2016, Las Vegas, NV, USA, June 27-30, 2016}, pages
  4104--4113, 2016.

\bibitem{SeoLJHC18}
Paul~Hongsuck Seo, Jongmin Lee, Deunsol Jung, Bohyung Han, and Minsu Cho.
\newblock Attentive semantic alignment with offset-aware correlation kernels.
\newblock In {\em Computer Vision - {ECCV} 2018 - 15th European Conference,
  Munich, Germany, September 8-14, 2018, Proceedings, Part {IV}}, pages
  367--383, 2018.

\bibitem{RANSAC-flow}
Xi Shen, Fran{\c{c}}ois Darmon, Alexei~A Efros, and Mathieu Aubry.
\newblock Ransac-flow: generic two-stage image alignment.
\newblock In {\em 16th European Conference on Computer Vision}, 2020.

\bibitem{shrivastava-sa11}
Abhinav Shrivastava, Tomasz Malisiewicz, Abhinav Gupta, and Alexei~A. Efros.
\newblock Data-driven visual similarity for cross-domain image matching.
\newblock {\em ACM Transaction of Graphics (TOG) (Proceedings of ACM SIGGRAPH
  ASIA)}, 30(6), 2011.

\bibitem{Simard2003}
Patrice~Y. Simard, Dave Steinkraus, and John~C. Platt.
\newblock Best practices for convolutional neural networks applied to visual
  document analysis.
\newblock In {\em Proceedings of the Seventh International Conference on
  Document Analysis and Recognition - Volume 2}, ICDAR '03, page 958, USA,
  2003. IEEE Computer Society.

\bibitem{SimonyanZ14}
Karen Simonyan and Andrew Zisserman.
\newblock Two-stream convolutional networks for action recognition in videos.
\newblock In Z. Ghahramani, M. Welling, C. Cortes, N. Lawrence, and K.~Q.
  Weinberger, editors, {\em Advances in Neural Information Processing Systems},
  volume~27, pages 568--576. Curran Associates, Inc., 2014.

\bibitem{Sun2018}
Deqing Sun, Xiaodong Yang, Ming{-}Yu Liu, and Jan Kautz.
\newblock Pwc-net: Cnns for optical flow using pyramid, warping, and cost
  volume.
\newblock In {\em 2018 {IEEE} Conference on Computer Vision and Pattern
  Recognition, {CVPR} 2018, Salt Lake City, UT, USA, June 18-22, 2018}, pages
  8934--8943, 2018.

\bibitem{Taniai2016}
Tatsunori Taniai, Sudipta~N. Sinha, and Yoichi Sato.
\newblock Joint recovery of dense correspondence and cosegmentation in two
  images.
\newblock In {\em 2016 {IEEE} Conference on Computer Vision and Pattern
  Recognition, {CVPR} 2016, Las Vegas, NV, USA, June 27-30, 2016}, pages
  4246--4255, 2016.

\bibitem{YFCC}
Bart Thomee, David~A. Shamma, Gerald Friedland, Benjamin Elizalde, Karl Ni,
  Douglas Poland, Damian Borth, and Li{-}Jia Li.
\newblock {YFCC100M:} the new data in multimedia research.
\newblock {\em Commun. {ACM}}, 59(2):64--73, 2016.

\bibitem{abs-2012-09842}
Georgi Tinchev, Shuda Li, Kai Han, David Mitchell, and Rigas Kouskouridas.
\newblock Xresolution correspondence networks.
\newblock {\em CoRR}, abs/2012.09842, 2020.

\bibitem{GLU-Net}
Prune Truong.
\newblock {GLU-Net: Github project page.}
\newblock \url{https://github.com/PruneTruong/GLU-Net}, 2020.

\bibitem{GLAMpoint}
Prune Truong, Stefanos Apostolopoulos, Agata Mosinska, Samuel Stucky, Carlos
  Ciller, and Sandro~De Zanet.
\newblock Glampoints: Greedily learned accurate match points.
\newblock {\em 2019 IEEE/CVF International Conference on Computer Vision
  (ICCV)}, pages 10731--10740, 2019.

\bibitem{GOCor}
Prune Truong, Martin Danelljan, Luc~Van Gool, and Radu Timofte.
\newblock {GOCor}: Bringing globally optimized correspondence volumes into your
  neural network.
\newblock In {\em Annual Conference on Neural Information Processing Systems,
  NeurIPS}, 2020.

\bibitem{GLUNet}
Prune Truong, Martin Danelljan, and Radu Timofte.
\newblock {GLU-Net}: Global-local universal network for dense flow and
  correspondences.
\newblock In {\em 2020 {IEEE} Conference on Computer Vision and Pattern
  Recognition, {CVPR} 2020}, 2020.

\bibitem{WangJE19}
Xiaolong Wang, Allan Jabri, and Alexei~A. Efros.
\newblock Learning correspondence from the cycle-consistency of time.
\newblock In {\em {IEEE} Conference on Computer Vision and Pattern Recognition,
  {CVPR} 2019, Long Beach, CA, USA, June 16-20, 2019}, pages 2566--2576, 2019.

\bibitem{UnOS}
Yang Wang, Peng Wang, Zhenheng Yang, Chenxu Luo, Yi Yang, and Wei Xu.
\newblock Unos: Unified unsupervised optical-flow and stereo-depth estimation
  by watching videos.
\newblock In {\em {IEEE} Conference on Computer Vision and Pattern Recognition,
  {CVPR} 2019, Long Beach, CA, USA, June 16-20, 2019}, pages 8071--8081, 2019.

\bibitem{OccAwareFlow}
Yang Wang, Yi Yang, Zhenheng Yang, Liang Zhao, and Wei Xu.
\newblock Occlusion aware unsupervised learning of optical flow.
\newblock {\em CoRR}, abs/1711.05890, 2017.

\bibitem{WangBSS04}
Zhou Wang, Alan~C. Bovik, Hamid~R. Sheikh, and Eero~P. Simoncelli.
\newblock Image quality assessment: from error visibility to structural
  similarity.
\newblock {\em {IEEE} Trans. Image Process.}, 13(4):600--612, 2004.

\bibitem{D2D}
Olivia Wiles, S{\'{e}}bastien Ehrhardt, and Andrew Zisserman.
\newblock {D2D:} learning to find good correspondences for image matching and
  manipulation.
\newblock {\em CoRR}, abs/2007.08480, 2020.

\bibitem{6836101}
Y. Xiang, R. Mottaghi, and S. Savarese.
\newblock Beyond pascal: A benchmark for 3d object detection in the wild.
\newblock In {\em 2014 IEEE Winter Conference on Applications of Computer
  Vision (WACV)}, volume~00, pages 75--82, March 2014.

\bibitem{abs-2004-09061}
Yang You, Chengkun Li, Yujing Lou, Zhoujun Cheng, Lizhuang Ma, Cewu Lu, and
  Weiming Wang.
\newblock Semantic correspondence via 2d-3d-2d cycle.
\newblock {\em CoRR}, abs/2004.09061, 2020.

\bibitem{BackToBasics}
Jason~J. Yu, Adam~W. Harley, and Konstantinos~G. Derpanis.
\newblock Back to basics: Unsupervised learning of optical flow via brightness
  constancy and motion smoothness.
\newblock In {\em Computer Vision - {ECCV} 2016 Workshops - Amsterdam, The
  Netherlands, October 8-10 and 15-16, 2016, Proceedings, Part {III}}, pages
  3--10, 2016.

\bibitem{OANet}
Jiahui Zhang, Dawei Sun, Zixin Luo, Anbang Yao, Lei Zhou, Tianwei Shen, Yurong
  Chen, Hongen Liao, and Long Quan.
\newblock Learning two-view correspondences and geometry using order-aware
  network.
\newblock In {\em 2019 {IEEE/CVF} International Conference on Computer Vision,
  {ICCV} 2019, Seoul, Korea (South), October 27 - November 2, 2019}, pages
  5844--5853, 2019.

\bibitem{Zhou2019}
Bolei Zhou, Hang Zhao, Xavier Puig, Tete Xiao, Sanja Fidler, Adela Barriuso,
  and Antonio Torralba.
\newblock Semantic understanding of scenes through the {ADE20K} dataset.
\newblock {\em Int. J. Comput. Vis.}, 127(3):302--321, 2019.

\bibitem{Zhou2016}
Tinghui Zhou, Philipp Kr{\"{a}}henb{\"{u}}hl, Mathieu Aubry, Qi{-}Xing Huang,
  and Alexei~A. Efros.
\newblock Learning dense correspondence via 3d-guided cycle consistency.
\newblock In {\em 2016 {IEEE} Conference on Computer Vision and Pattern
  Recognition, {CVPR} 2016, Las Vegas, NV, USA, June 27-30, 2016}, pages
  117--126, 2016.

\bibitem{ZhouLYE15}
Tinghui Zhou, Yong~Jae Lee, Stella~X. Yu, and Alexei~A. Efros.
\newblock Flowweb: Joint image set alignment by weaving consistent, pixel-wise
  correspondences.
\newblock In {\em {IEEE} Conference on Computer Vision and Pattern Recognition,
  {CVPR} 2015, Boston, MA, USA, June 7-12, 2015}, pages 1191--1200, 2015.

\end{thebibliography}
}

\clearpage
\newpage
\appendix
\begin{center}
	\textbf{\Large Appendix}
\end{center}
In this supplementary material, we give additional details about our approach, experiment settings and results.
We first give additional details about the flow-constraints derived from our introduced warp consistency graph in Sec.~\ref{sec-sup:warpgraph}. We also provide additional empirical comparisons between the corresponding regression losses.
We follow by explaining the triplet image creation and the sampling process of our synthetic warps $W$ in Sec.~\ref{sec-sup:transfo}.
In Sec.~\ref{sec-sup:glunet}, we then focus on the training procedure to obtain WarpC-GLU-Net in more depth.  We subsequently continue by explaining the training details of WarpC-RANSAC-Flow and WarpC-SemanticGLU-Net in respectively Sec.~\ref{sec-sup:RANSAC-flow} and~\ref{sec-sup:semanticglunet}. For completeness, in Sec.~\ref{sec-sup:ablation-details}, we also provide details about the training of all networks compared in the method analysis, corresponding to Sec.~\ref{sec:method-analysis} of the main paper. 
In all aforementioned sections, we provide additional information about the architecture, its original training strategy, our proposed training approach comprising the sampled transformations $W$, as well as implementation details. We then follow by analysing the effect of the strength of the sampled warps $W$ in Sec.~\ref{sec-sup:transfo-analysis}. 
In Sec.~\ref{sup-sec:time_memory}, we follow by discussing the time and memory efficiency of our proposed approach during training and testing.
Subsequently, we perform additional ablative and method analysis experiments in Sec.~\ref{sec-sup:ablation}. 
In Sec.~\ref{sec-sup:details-evaluation}, we extensively explain the evaluation datasets and set-up.  Finally we present more detailed quantitative and qualitative results in Sec.~\ref{sec-sup:results}.
In particular, we show quantitative results on the pose estimation dataset YFCC100M~\cite{YFCC} as well as the geometric matching dataset HPatches~\cite{Lenc}. We also provide results on the semantic datasets PF-Willow~\cite{PFWillow} and SPair-71k~\cite{spair}. 
Finally, we show the possible extension of our unsupervised approach to optical flow data.

\section{Warp consistency graph regression losses}
\label{sec-sup:warpgraph}

In this section, we provide additional details about the possible flow-constraints derived from our warp consistency graph (Sec.~\ref{sec:cyclic-consistency-graph} of the main paper). We also show qualitative and quantitative comparisons between the trained networks using each possible regression loss.

\subsection{Details about \ji constraint}
\label{subsec-sup:ji}

We here provide the detailed derivation of the bias insensitivity of the \ji loss, which is given by (eq.~\eqref{eq:ji-bipath} in the main paper) as,
\begin{equation}
\label{eq:ji-bipath-sup}
L_{J \rightarrow I} =  \big \| \widehat{F}_{J\rightarrow I'} + \warp_{\widehat{F}_{J\rightarrow I'}} (W)  - \widehat{F}_{J \rightarrow I} \big \|.
\end{equation}
We derive an upper bound for the change in the loss $\Delta L_{J \rightarrow I}$ when a constant bias $\mathbf{b} \in \reals^2$ is added to all flow predictions $\widehat{F}$. We have,
\begin{align}
    \label{eq:ji-bias}
    \Delta L_{J \rightarrow I} =\,& \big \| \widehat{F}_{J\rightarrow I'} + \mathbf{b} + \warp_{\widehat{F}_{J\rightarrow I'} + \mathbf{b}} (W) - (\widehat{F}_{J \rightarrow I} + \mathbf{b}) \big \| \nonumber \\
    & -\big \| \widehat{F}_{J\rightarrow I'} + \warp_{\widehat{F}_{J\rightarrow I'}} (W)  - \widehat{F}_{J \rightarrow I} \big \| \nonumber\\
    =\,& \big \| \widehat{F}_{J\rightarrow I'} + \warp_{\widehat{F}_{J\rightarrow I'}} (W)  - \widehat{F}_{J \rightarrow I} \nonumber\\ 
    &\qquad + \warp_{\widehat{F}_{J\rightarrow I'} + \mathbf{b}} (W) - \warp_{\widehat{F}_{J\rightarrow I'}} (W) \big \| \nonumber \\
    & - \big \| \widehat{F}_{J\rightarrow I'} + \warp_{\widehat{F}_{J\rightarrow I'}} (W)  - \widehat{F}_{J \rightarrow I} \big \| \nonumber\\
    \leq\,& \big \| \widehat{F}_{J\rightarrow I'} + \warp_{\widehat{F}_{J\rightarrow I'}} (W)  - \widehat{F}_{J \rightarrow I}\big \| \nonumber\\  &+\big \|\warp_{\widehat{F}_{J\rightarrow I'} + \mathbf{b}} (W) - \warp_{\widehat{F}_{J\rightarrow I'}} (W) \big \| \nonumber \\
    &- \big \| \widehat{F}_{J\rightarrow I'} + \warp_{\widehat{F}_{J\rightarrow I'}} (W)  - \widehat{F}_{J \rightarrow I} \big \| \nonumber\\
    =\,& \big \|\warp_{\widehat{F}_{J\rightarrow I'} + \mathbf{b}} (W) - \warp_{\widehat{F}_{J\rightarrow I'}} (W) \big \| \,.
\end{align}
Here we have used the triangle inequality. From the bound above, we can already see that $\Delta L_{J \rightarrow I}$ will be small if $W$ is changing slowly. We can see this more clearly by assuming the bias $\mathbf{b}$ to be small, and doing a first order Taylor expansion,
\begin{align}
    \label{eq:ji-taylor}
    \warp_{\widehat{F}_{J\rightarrow I'} + \mathbf{b}}& (W)(\mathbf{x}) = W \big(\mathbf{x} + \widehat{F}_{J\rightarrow I'}(\mathbf{x}) + \mathbf{b}\big) \nonumber\\
    \approx\,& W \big(\mathbf{x} + \widehat{F}_{J\rightarrow I'}(\mathbf{x})\big) + D W \big(\mathbf{x} + \widehat{F}_{J\rightarrow I'}(\mathbf{x})\big)\mathbf{b} \nonumber \\
    =\,& \warp_{\widehat{F}_{J\rightarrow I'}} (W)(\mathbf{x}) + \warp_{\widehat{F}_{J\rightarrow I'}} (DW\mathbf{b})(\mathbf{x}) \,.
\end{align}
Here, $ D W (\mathbf{x}) \in \reals^{2\times 2}$ is the Jacobian of $W$ at location $\mathbf{x} \in \reals^2$. Thus, $DW\mathbf{b}$ denotes the function obtained from the matrix-vector product between the Jacobian $D W$ and bias $\mathbf{b}$ at every location. Inserting~\eqref{eq:ji-taylor} into~\eqref{eq:ji-bias} gives an approximate bound valid for small $\mathbf{b}$,
\begin{equation}
    \label{eq:ji-bound}
    \Delta L_{J \rightarrow I} \lessapprox \big\|\warp_{\widehat{F}_{J\rightarrow I'}} (DW\mathbf{b})\big\| \,.
\end{equation}
A smooth and invertible warp $W$ implies a generally small Jacobian $DW$. Since the bias $\mathbf{b}$ is scaled with $DW$, the resulting change in the loss will also be small. As a spacial case, it is immediately seen from~\eqref{eq:ji-bias} that the change in the loss is always zero if $W$ is a pure translation. The bias insensitivity of the \ji constraint largely explains its poor performance. As visualized in Fig.~\ref{Fig.:warp-graph-analysis}, the predictions of a network trained with solely the \ji loss~\eqref{eq:ji-bipath} suffer from a large translation bias.

\subsection{Cycle constraints} 

Here, we provide additional details about the cycle constraints, extracted from our warp consistency graph. 
As explained in Sec.~\ref{sec:cyclic-consistency-graph} of the main paper, because of the fixed direction of the known flow $W$ which corresponds to $I' \rightarrow I$, three cycle constraints are possible, starting from either images $I$, $I'$ or $J$ and composing mappings so that the resulting composition is equal to the identity map. They are respectively formulated as follows, 
\begin{subequations}
\label{eq:cycle-mapping}
\begin{align}
    \mathbb{I} &= M_{W} \circ M_{J \rightarrow  I'} \circ  M_{I \rightarrow J}\\
    \mathbb{I} &= M_{J \rightarrow  I'} \circ M_{I \rightarrow  J} \circ  M_{W} \\ 
    \mathbb{I} &= M_{I \rightarrow J} \circ M_{W} \circ  M_{J \rightarrow I'}
\end{align}
\end{subequations}
The corresponding regression losses are obtained by converting the mapping constraints~\eqref{eq:cycle-mapping} to flow constraints and considering only the flow $W$ as known. We provide the expression for each of the three cycle losses in the following.

\parsection{Cycle from $I$} By starting from image $I$ and performing a full cycle, the resulting regression loss is expressed as,
\begin{align}
\label{eq: loss-cycle-i}
    L_{\text{cycle-I}} =&  \big \| \widehat{F}_{I \rightarrow J} + \warp_{\widehat{F}_{I \rightarrow J}} ( \widehat{F}_{J \rightarrow I'}) \,\, +  \\
    & \warp_{ \widehat{F}_{I \rightarrow J} + \warp_{\widehat{F}_{I \rightarrow J}} ( \widehat{F}_{J \rightarrow I'}  ) } ( W )   \big\| \nonumber
\end{align}

\parsection{Cycle from $I'$} Starting from image $I'$ instead leads to the following regression loss,
\begin{align}
\label{eq: loss-cycle-iprime}
    L_{\text{cycle-I'}}  =&  \big\|  W + \warp_{W} ( \widehat{F}_{I \rightarrow J} ) \, \, +   \\
    & \warp_{W + \warp_{W} ( \widehat{F}_{I \rightarrow J})} ( \widehat{F}_{J \rightarrow I'} )  \big\|  \nonumber
\end{align}

\parsection{Cycle from $J$} Finally, using image $J$ as starting point for the cycle constraint results in this regression loss,
\begin{align}
\label{eq: loss-cycle-j}
    L_{\text{cycle-J}}  =&  \big \| \widehat{F}_{J \rightarrow I'} + \warp_{\widehat{F}_{J \rightarrow I'}} (W) \,\,+  \\
    & \warp_{  \widehat{F}_{J \rightarrow I'} + \warp_{\widehat{F}_{J \rightarrow I'}} (W) } ( \widehat{F}_{I \rightarrow J}) \big\| \nonumber
\end{align}

\parsection{Note about warp consistenty loss} Concerning the adaptive loss balancing
of our final warp consistency unsupervised objective $L = L_{\text{W-vis}} + \lambda L_{\text{warp}}$ (Sec.~\ref{sec:our-loss} of the main paper), since $\lambda$ is a weighting factor, we do not backpropagate gradients through it. 

\subsection{Quantitative and qualitative analysis}
\label{subsec-sup:analysis}

\parsection{Extension of quantitative analysis} We first extend Tab.~\ref{tab:graph-analysis} of the main paper, by analysing the remaining warp consistency graph losses. Results on MegaDepth, RobotCar and HPatches are presented in Tab.~\ref{Tab.-sup:graph-analysis}. As in Tab.~\ref{tab:graph-analysis} of the main paper, all networks are trained following the first training stage of WarpC-GLU-Net (See Sec.~\ref{sec:method-analysis} of main paper or Sec.~\ref{sec-sup:glunet}).

We first provide evaluation results of networks trained using the cycle losses, starting from images $I$ and $J$.
The cycle loss starting from $I$ obtains very poor results. The cycle starting from $J$ instead achieves better performance, but still lower than the cycle loss from $I'$. 
The \w constraint obtains the best results overall. 

We then compare combinations of the derived losses with the warp-supervision objective (eq. \eqref{eq:warp-supervision} of the main paper). Between the cycle losses, the combination of the warp-supervision with the cycle loss from $I'$ achieves the best results compared to the combinations with the cycle losses from $I$ and $J$.
The combination of the warp-supervision and forward-backward losses (eq. \eqref{eq:warp-supervision} of the main paper), which are both retrieved as pair-wise constraints from the warp consistency graph (Sec.~\ref{sec:cyclic-consistency-graph} and Fig.~\ref{fig:cycles-method}e of the main paper), leads to lower generalisation abilities on the HPatches dataset than our warp consistency loss. It also achieves substantially lower PCK-1 on MegaDepth. Moreover, because the forward-backward consistency loss leads to a degenerate trivial solution when used alone, manual tuning of a weighting hyper-parameter is required to balance the warp-supervision and the forward-backward loss terms. If it is too high, the forward-backward term gains too much importance and drives the network to zero. If it is too small instead, its contribution becomes insignificant. 
On the contrary, our proposed unsupervised learning objective (Sec.~\ref{sec:our-loss} of the main paper) does not require expensive manual tuning of such hyperparameters.

\begin{table}[t]
\centering
\resizebox{\columnwidth}{!}{%
\begin{tabular}{lccc|ccc|cc} \toprule
& \multicolumn{3}{c}{\textbf{MegaDepth}} & \multicolumn{3}{c}{\textbf{RobotCar}} & \multicolumn{2}{c}{\textbf{HPatches}} \\
& PCK-1  & PCK-5 & PCK-10  & PCK-1 & PCK-5 & PCK-10  & AEPE & PCK-5 \\ \midrule
Warp-supervision~\eqref{eq:warp-supervision} & 35.98 & 57.21 & 63.88 & 2.43 & 33.63 & 54.50 & 28.50 & 76.76  \\
\ipj~\eqref{eq:semi-cycle-map-a} & 0.00 & 0.05 & 0.21 & 0.00 & 0.00 & 0.13 & 370.80 & 0.01 \\
\ji~\eqref{eq:semi-cycle-map-b},\eqref{eq:ji-bipath}  & 0.00 & 0.06 & 0.21 & 0.00 & 0.05 & 0.21 & 162.50 & 0.04 \\
\w~\eqref{eq:semi-cycle-map-c},\eqref{eq:w-bipath} &  \textbf{29.55} & \textbf{67.70} & \textbf{74.42} & \textbf{2.25} & \textbf{33.88} & \textbf{55.38} & \textbf{26.13} & \textbf{70.51} \\ 
$I'$-cycle~\eqref{eq: loss-cycle-iprime} & 25.04 & 64.44 & 71.75 & 2.19 & 32.79 & 54.55 & 27.51 & 66.16 \\ 
$I$-cycle~\eqref{eq: loss-cycle-i} &  0.00 & 0.14 & 0.56 & 0.03 & 0.74 & 5.29 & 232.24 & 0.04 \\ $J$-cycle~\eqref{eq: loss-cycle-j} & 17.91 & 54.95 & 62.81 & 2.05 & 30.96 & 52.06 & 42.67 & 49.06 \\
\midrule
\ipj + warp-sup. & 0.00 & 0.11 & 0.45 & 0.01 & 0.35 & 1.52 & 255.40 & 0.02 \\
\ji + warp-sup. &  33.72 & 61.10 & 67.44 & 2.26 & 34.06 & 55.07 & 28.91 & 71.52 \\ 
\w + warp-sup.  & \textbf{43.47} & \textbf{69.90} & \textbf{75.23} & 2.49 & 35.28 & 56.45 & \textbf{22.83} & \textbf{78.60} \\ 
$I'$-cycle + warp-sup. & 42.11 & 68.84 & 74.28 & \textbf{2.52} & \textbf{35.75} & \textbf{56.96} & 24.16 & 78.58  \\ 
$I$-cycle + warp-sup. & 0.00 & 0.16 & 0.69 & 0.05 & 1.26 & 4.67 & 225.94 & 0.04 \\
$J$-cycle + warp-sup. & 41.56 & 68.33 & 73.85 & 2.37 & 35.20 & 56.36 & 24.69 & 75.35 \\
Warp-super. + f-b & 41.54 & 69.78 & 74.83 & 2.47 & 35.25 & 56.39 & 26.15 & 74.83 \\
\bottomrule
\end{tabular}%
}\vspace{1mm}
\caption{Analysis of warp consistency graph losses (Sec.~\ref{sec:cyclic-consistency-graph}-\ref{sec:semicycle} of the main paper). }
\label{Tab.-sup:graph-analysis}\vspace{-2mm}
\end{table}

\begin{figure*}[t]
\centering%
\vspace{-6mm}
\includegraphics[width=0.85\textwidth]{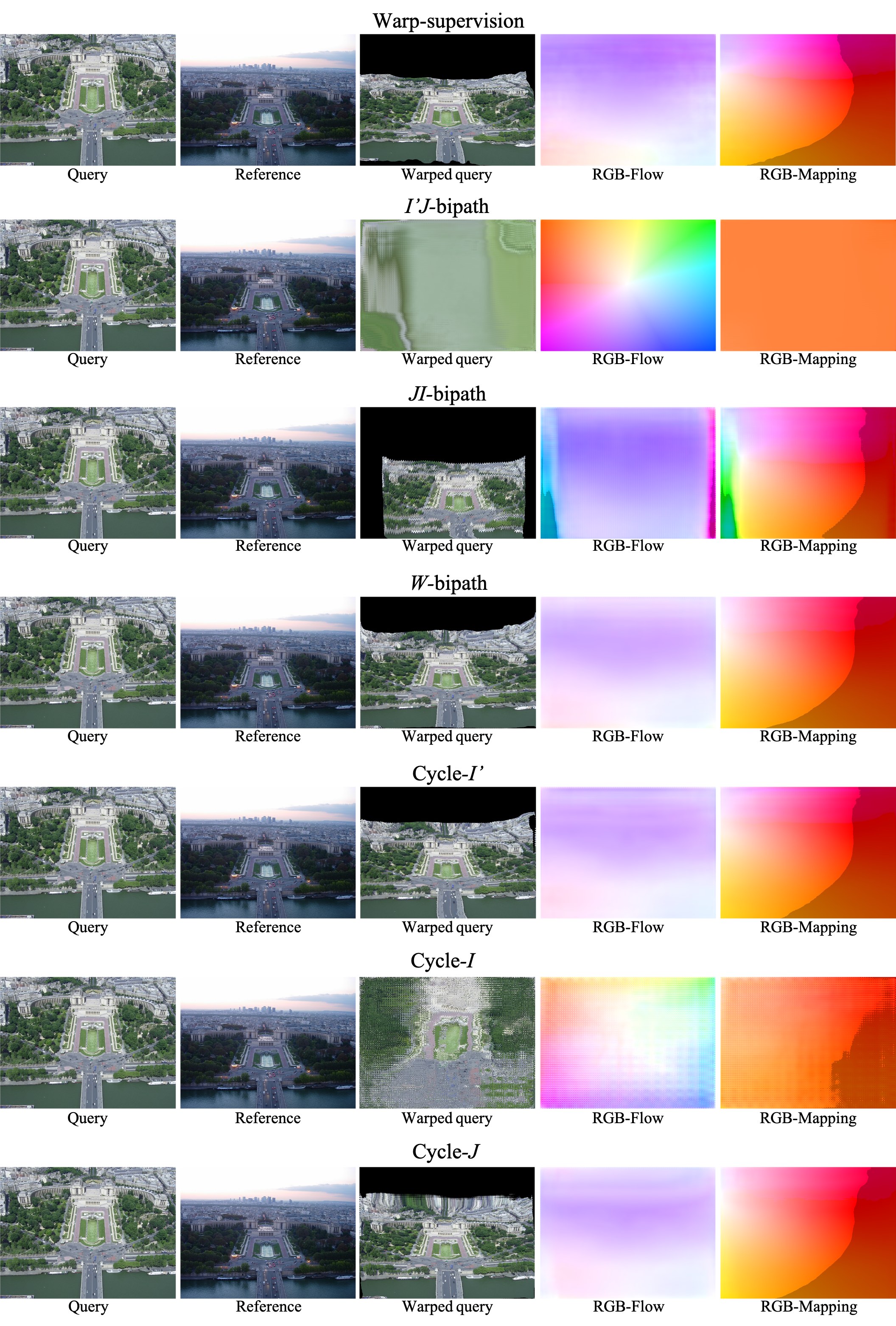}
\vspace{-1mm}
\caption{Visual comparison on an image pairs of the MegaDepth dataset, of the performance of the different losses derived from the warp-consistency graph. We additionally show for each loss, the estimated mapping and flow using flow RGB representation. Note that all networks are trained following the first training stage of WarpC-GLU-Net (See Sec.~\ref{sec:method-analysis} of main paper or Sec.~\ref{sec-sup:glunet}).}
\vspace{-2mm}
\label{Fig.:warp-graph-analysis}
\end{figure*}

\parsection{Qualitative comparison} In Fig.~\ref{Fig.:warp-graph-analysis}, we visually compare the estimated flows by GLU-Net networks trained using each of the flow-consistency losses retrieved from the warp consistency graph. Training using the warp-supervision loss alone results in an unstable estimated flow, and corresponding warped query.
It can directly be seen that the \ipj loss results in the network learning a degenerate trivial solution, in the form of a constant predicted mapping independently of the input images. 
Training with the \ji objective instead makes the network insensitive to an additional predicted bias. Indeed, in Fig.~\ref{Fig.:warp-graph-analysis}, third row, it is easily seen that the warped query is shifted towards the right and bottom, compared to the reference image. This is due to a constant predicted bias by the network. 
The \w objective leads to a drastically better warped query. Also note that the estimated flow leads to a more accurate warped query than when trained with the $I'$-cycle loss. Training with the cycle loss from $I$ leads to very poor results instead. Finally, the cycle loss derived by starting from image $J$ results in a reasonable warped query, but it has more out-of-regions artifacts compared to the prediction of the network trained with the \w loss.

\section{Triplet creation and sampling of warps $W$}
\label{sec-sup:transfo}

\subsection{Triplet creation}

Our introduced unsupervised learning approach requires to construct an image triplet $(I, I', J)$ from an original image pair $(I, J)$, where all three images must have training dimensions $s \times s$. We construct the triplet $(I, I', J)$ as follows. 
The original training image pairs $(I, J)$ are first resized to a fixed size $s_r \times s_r$, larger than the desired training image size $s \times s$. 
We then sample a dense flow $W$ of the same dimension $s_r \times s_r$, and create $I'$ by warping image $I$ with $W$, as $I' = \warp_{W} (I)$. 
Each of the images of the resulting image triplet $(I, I', J)$ are then centrally cropped to the fixed training image size $s \times s$. The central cropping is necessary to remove most of the black areas in $I'$ introduced from the warping operation with large sampled flows $W$ as well as possible warping artifacts arising at the image borders. 
We then additionally apply appearance transformations to the created image $I'$, such as brightness and contrast changes. 
This procedure is similar to~\cite{Rocco2017a}, which employs solely the warp-supervision objective on $(I', I)$.

\subsection{Sampling of warps $W$} 

As mentioned in the main paper Sec.~\ref{sec:our-transfo}, we key question raised by our proposed loss formulation is how to sample the synthetic flows $W$. The analysis of the properties of the proposed \w loss brought some insight into what magnitude $\|W\|$ of warps to sample during training. If the generated warps are too small, there is still a risk of biasing the prediction towards zero. Instead, using warps of roughly similar order of magnitude $\|W\|$ as the underlying transformations $\|F_{J \rightarrow I}\|$ would give equal impact to all three terms in eq. \eqref{eq:vis-w-bipath} of the main paper. 
During training, we randomly sample it from a distribution $W \sim p_W$, which we need to design.

\parsection{Base transformation sampling} We construct $W$ by sampling homography, Thin-plate Spline (TPS), or affine-TPS transformations with equal probability. The transformations parameters are then converted to dense flows of dimension $s_r \times s_r$. 

Specifically, for homographies and TPS, the four image corners and a $3 \times 3$ grid of control points respectively, are randomly translated in both horizontal and vertical directions, according to a desired sampling scheme. The translated and original points are then used to compute the corresponding homography and TPS parameters. Finally, the transformations parameters are converted to dense flows. 
For both transformation types, the magnitudes of the translations are sampled according to a uniform or Gaussian distribution with a range or standard-deviation $\sigma_H$ respectively.
Note that for the uniform distribution, the sampling range is actually $\left [- \sigma_H, \sigma_H \right]$, when it is centered at zero, or similarly $\left [1 - \sigma_H, 1 + \sigma_H \right]$ if centered at 1 for example. 
Importantly, the image points coordinates are previously normalized to be in the interval $\left [ -1 , 1\right]$. Therefore $\sigma_H$ should be within $\left [ 0, 1\right]$. 

For the affine transformations, all parameters, \ie scale, translations, shearing and rotation angles, are sampled from a uniform or Gaussian distribution with range or standard-deviation equal to $\tau$, $t$, $\alpha$ and $\alpha$ respectively. 
For the affine scale parameter, the corresponding Gaussian sampling is centered at one whereas for all other parameters, it is centered at zero. Similarly, for a uniform sampling instead, the affine scale parameters is sampled within $\left [1-\tau, 1 + \tau \right]$ with center at 1, while for all other parameters, the sampling interval is centered at zero. 

\parsection{Elastic transformations} To make the synthetic flow $W$ harder for the network to estimate, we also optionally compose the base flow resulting from sampling homography, TPS and Affine-TPS transformations, with a dense elastic deformation grid. 
We generate the corresponding elastic residual flow $\epsilon = \sum_i \varepsilon_i$, by adding small local perturbations $\varepsilon_i \in \reals^{s_r \times s_r \times 2}$.
More specifically, we create the residual flow by first generating an elastic deformation motion field $E$ on a dense grid of dimension $s_r \times s_r$, as described in~\cite{Simard2003}. Since we only want to include elastic perturbations in multiple small regions, we generate binary masks $S_i \in \mathbb{R}^{s_r \times s_r}$, each delimiting the area on which to apply one local perturbation $\varepsilon_i$.  The final elastic residual flow thus take the form of $\epsilon = \sum_i \varepsilon_i$, where $\varepsilon_i = E \cdot S_i$. 
The final synthetic warp $W$ is achieved by composing the base flow  with the elastic residual flow $\epsilon$. 

In practise, for the elastic deformation field $E$, we use the implementation of~\cite{info11020125}. The masks $S_i$ should be between 0 and 1 and offer a smooth transition between the two, so that the perturbations appear smoothly. To create each mask $S_i$, we thus generate a 2D Gaussian centered at a random location and with a random standard deviation (up to a certain value) on a dense grid of size $s_r \times s_r$. It is then scaled to 2.0 and clipped to 1.0, to obtain smooth regions equal to 1.0 where the perturbations will be applied, and transition regions on all sides from 1.0 to 0.0.

\subsection{Hyper-parameters} 

In summary, to construct our image triplet $(I, I', J)$, the hyper-parameters are the following: 

\bp{(i)} $s_r$, the resizing image size, on which is applied $W$ to obtain $I'$ before cropping.

\bp{(ii)} $s$, the training image size, which correspond to the size of the training images after cropping. 

\bp{(iii)} $\sigma_H$, the range or standard deviation used for sampling the homography and TPS transformations.  

\bp{(iv)} $\tau$, the range or standard deviation used for sampling the scaling parameter of the affine transformations. 

\bp{(v)} $t$, the range or standard deviation used for sampling the translation parameter of the affine transformations. 

\bp{(vi)} $\alpha$, the range or standard deviation used for sampling the rotation angle of the affine transformations. It is also used as shearing angle.  

\bp{(vii)} $\sigma_{tps}$, the range or standard deviation used for sampling the TPS transformations, used for the Affine-TPS compositions. 

For simplicity, in all experiments including elastic deformations, we use the same elastic transformations hyper-parameters. Moreover, for all experiments and networks, we apply the same appearance transformations to image $I'$.
Specifically, we use color transformations, by adjusting contrast, saturation, brightness, and hue. 
With probability 0.2, we additionally use a Gaussian blur with a kernel between 3 and 7, and a standard deviation sampled within $\left[ 0.2, 2.0 \right]$.

\section{Training details for WarpC-GLU-Net}
\label{sec-sup:glunet}

We first provide details about the original GLU-Net architecture and the modifications we made for this work. We also briefly review the training strategy of the original work. We then extensively explain our training approach and the corresponding implementation details.

\subsection{Details about GLU-Net}

\parsection{Architecture} We use GLU-Net as our base architecture. It is a 4 level pyramidal network, using a VGG-16 feature backbone~\cite{Chatfield14}, initialized with pre-trained weights on ImageNet. It is composed of two sub-networks, L-Net and H-Net which act at two different resolutions. 
The L-Net takes as input rescaled images to $h_L \times w_L= 256 \times 256$ and process them with a global feature correlation layer followed by a local feature correlation layer. The resulting flow is then upsampled to the lowest resolution of the H-Net to serve as initial flow, by warping the query features according to the estimated flow. The H-Net takes as input images the original images at unconstrained resolution $h\times w$, and refines the estimated flow with two local feature correlation layers. 
We adopt the GLU-Net architecture and simply replace the DenseNet connections~\cite{Huang2017} of the flow decoders by residual connections. We also include residual blocks in the mapping decoder. This drastically reduces the number of weights while having limited impact on performance. 

\parsection{Training strategy in original work} In the original GLU-Net~\cite{GLU-Net}, the network is trained with the warp-supervision loss (referred to as a type of self-supervised training strategy in original publication), which corresponds to equation \eqref{eq:warp-supervision} of the main paper. As for the synthetic sampled transformations $W$, Truong \etal~\cite{GLU-Net} use the same 40k synthetic transformations (affine, thin-plate and homographies) than in DGC-Net~\cite{Melekhov2019}, but apply them to images collected from the DPED~\cite{Ignatov2017}, CityScapes~\cite{Cordts2016} and ADE-20K~\cite{Zhou2019} datasets.

\subsection{WarpC-GLU-Net: our training strategy}

We here explain the different steps of our training strategy in more depth. 

\parsection{Training stages}
In the first training stage, we train GLU-Net using our warp consistency loss (Sec.~\ref{sec:our-loss} of the main paper)  without the visibility mask. This is because the estimated flow field needs to reach a reasonable performance in order to compute the visibility mask (eq.~\ref{eq:vis-mask} of the main paper). 
In the second training stage, we further introduce the visibility mask in the \w loss term (eq.~\ref{eq:vis-w-bipath} of the main paper). In order to enhance difficulty in the second stage, we increase the transformations strengths and include additional elastic transformations for the sampled warps $W$. Note that the feature backbone is initialized to the ImageNet weights and not further trained.  

\parsection{Training dataset} For training, we use the MegaDepth dataset, consisting of 196 different scenes reconstructed from 1,070,468 internet photos using COLMAP~\cite{COLMAP}. 
Specifically, we use 150 scenes of the dataset and sample up to 500 random images per scene.  
It results in around 58k training pairs. Note that we use the same set of training pairs at each training epoch. For the validation dataset, we sample up to 100 image pairs from 25 different scenes, leading to  approximately 1800 image pairs. 
Importantly, while we can get the corresponding sparse ground-truth correspondences from the SfM reconstructions, we do not use them during training in this work and only retrieve the image pairs.

\parsection{Warps $W$ sampling} We resize the image pairs $(I, J)$ to $s_r \times s_r = 750 \times 750$, sample a dense flow $W$ of the same dimension and create $I'$. Each of the images of the resulting image triplet $(I, I', J)$ is then centrally cropped to $s \times s = 520 \times 520$. In the following, we give the parameters used for the sampling of the flow $W$ in both training stages.

In the first stage, the flows $W$ are created by sampling homographies, TPS and Affine-TPS transformations with equal probability.
For homographies and TPS, we use a uniform sampling scheme with a range equal to $\left [-\sigma_H, \sigma_H \right ]$, where $\sigma_H = 0.33$, which corresponds to a displacement of up to 250 pixels for the image size $s_r = 750$.
For the affine transformations, we also sample all parameters, \ie scale, translation, shear and rotation angles, from uniform distributions with ranges respectively equal to $\tau = 0.45$, $t = 0.25$, and $\alpha = \pi/12$ for both angles.
We compose the affine transformations with TPS transformations, for which we sample the translation magnitudes uniformly with a range $\sigma_{tps} = 0.08$, thus corresponding to a displacement of up to 60 pixels. We chose a smaller range for the TPS compositions because we have found empirically that large ranges led to very drastic resulting dense Affine-TPS flows, which were not necessarily beneficial in the first training stage.

In the second stage, we also sample homographies, TPS and Affine-TPS transformations, but increase their strength. Specifically, for homography and TPS transformations, we use a range $\sigma_H = 0.4$ (displacements up to 300 pixels). The affine parameters are sampled as in the first training stage, but we increase the range of the uniform sampling for the TPS transformations to $\sigma_{tps}=0.26$ (displacements up to 200px). To make the flows $W$ even harder to estimate, we additionally include elastic transformations, sampled as explained in Sec.~\ref{sec-sup:transfo}. 

\parsection{Baseline comparison} For fair comparison, we retrain GLU-Net using the original training strategy, which corresponds to the warp-supervision training loss, on the same MegaDepth training images. We also use the same altered GLU-Net architecture as for WarpC-GLU-Net. Moreover, we make use of  the same synthetic transformations $W$ as in our first and second training stages. We call this version GLU-Net*.

\subsection{Implementation details} Since GLU-Net is a pyramidal architecture with $K$ levels, we employ a multi-scale training loss, where the loss at different pyramid levels account for different weights. 
\begin{equation}
\label{eq:multiscale-loss}
    \mathcal{L}(\theta)=\sum_{l=1}^{K} \gamma_{l} L_l +\eta \left\| \theta \right\|\,,
\end{equation}
where $\gamma_{l}$ are the weights applied to each pyramid level and $L_l$ is the corresponding loss computed at each level, which refers to the warp-supervision loss (eq.~\ref{eq:warp-supervision} of the main paper) for baseline GLU-Net* and our proposed warp consistency loss (Sec.~\ref{sec:our-loss} of the main paper) for WarpC-GLU-Net. 
The second term of the loss~\eqref{eq:multiscale-loss} regularizes the weights of the network. 
The hyper-parameters used in the estimation of our visibility mask $\widehat{V}$ (eq.~\ref{eq:vis-mask} of the main paper) are set to $\alpha_1 = 0.025$ and $\alpha_2 = 0.5$.
During training, we down-sample and scale the sampled $W$ from original resolution $h\times w$ to $h_L \times w_L$ in order to obtain the flow field $W$ for L-Net. For the loss computation, we down-sample the known flow field $W$ from the base resolution to the different pyramid resolutions without further scaling, so as to obtain the supervision signals at the different levels. 

For training, we use similar training parameters as in~\cite{GLUNet}. Specifically, as a preprocessing step, the training images are mean-centered and normalized using mean and standard deviation of the ImageNet dataset~\cite{Hinton2012}. For all local correlation layers, we employ a search radius $r=4$. 

For our network WarpC-GLU-Net and the baseline GLU-Net*, the weights in the training loss~\eqref{eq:multiscale-loss} are set to be $\gamma_{1}=0.32, \gamma_{2}=0.08, \gamma_{3}=0.02, \gamma_{4}=0.01$. During the first training stage, both networks are trained with a batch size of 6 for 400k iterations. The learning rate is initially equal to $10^{-4}$, and halved after 250k and 325k iterations. 
For the second training stage, we train for 225k iteration with an initial learning rate of $5.10^{-5}$, which is halved after 100k, 150k and 200k iterations. The networks are trained using Adam optimizer~\cite{adam} with weight decay of $0.0004$.

\section{Training details for WarpC-RANSAC-Flow}
\label{sec-sup:RANSAC-flow}

In this section, we first review the RANSAC-Flow architecture as well as their original training strategy. We then explain in more depth the different steps of our training, leading to WarpC-RANSAC-Flow.

\subsection{Details about RANSAC-Flow}

\parsection{Architecture} RANSAC-Flow inference is divided in two steps. First, the image pairs are pre-aligned by computing the homography relating them, using multi-scale feature matching based on off-the-shelf MOCO features~\cite{MOCO} and Ransac. 
As a second step, the pre-aligned image pairs are input to the trained RANSAC-Flow model, which predicts the flow and matchability mask relating them. The final flow field relating the original images is computed as a composition of the flow corresponding to the homography computed in the pre-alignment step, and the predicted flow field. 
RANSAC-Flow is a shallow architecture taking image pairs as input, and which regresses the dense flow field and matchability mask relating one image to the other. It relies on a single local feature correlation layer computed at one eight of the input images resolution. The local feature correlation layer is computed with a small search radius of $r=3$. 
The flow decoder and matchability branch are both fully convolutional with three convolution blocks, while the feature backbone is a modified version of ResNet-18~\cite{HeZRS15}.

\parsection{Training dataset} As training dataset, RANSAC-Flow uses images of the MegaDepth dataset~\cite{megadepth}, from which they selected a subset of image pairs. They pre-aligned the image pairs using their pre-processing multi-scale strategy with off-the-shelf MOCO feature~\cite{MOCO} matching and homography estimation with Ransac. The resulting training dataset comprises 20k pre-aligned image pairs, for which the remaining geometric transformation between the frames is relatively small.

\parsection{Training strategy in original work} In the original work~\cite{RANSAC-flow}, the training is separated in three stages. First, the network is trained using the SSIM loss~\cite{WangBSS04}, which is further combined with the forward-backward cyclic consistency loss (eq. \eqref{eq:forward-backward} of the main paper) in the second stage. During the two first stages, only the feature backbone and the flow decoder are trained, while the matchability branch remains unchanged and unused. 
In the last stage, the matchability branch is also trained by weighting the previous losses with the predicted mask and including a regularization matchability loss. A disadvantage of this approach is that all losses need to be scaled with a hyper-parameter, requiring expensive manual-tuning.

\subsection{WarpC-RANSAC-Flow: our training strategy}

\parsection{Training stages}
In the first training stage, we apply our proposed loss (Sec.~\ref{sec:our-loss} of the main paper) without the visibility mask, as in the first stage of WarpC-GLU-Net.
The visibility mask (eq.~\ref{eq:vis-w-bipath} of the main paper) is introduced in the second stage of training. As in original RANSAC-Flow, the two first stages only train the feature backbone and the flow decoder while keeping the matchability branch fixed (and unused). 
In the third stage, we jointly train the feature backbone, flow decoder and the matchability branch. As training loss, we use the original matchability regularization loss and further replace our visibility mask $\widehat{V}$ in the \w loss (eq.~\ref{eq:vis-w-bipath} of the main paper) with the predicted mask, output of the matchability branch.

\parsection{Warps $W$ sampling} We resize original images $(I, J)$ to $s_r \times s_r = 300 \times 300$. Following original RANSAC-Flow, the final training images have dimension $s \times s = 224 \times 224$. 
Because RANSAC-Flow uses a single local correlation layer with a search radius of 3 computed at one eight of the original image resolution, the network can theoretically only estimate geometric transformations up to $3.8 = 24$ pixels in all directions. This is a very limited compared to GLU-Net or other matching networks. It makes RANSAC-Flow architecture very sensitive to the magnitude of the geometric transformations and limited in the range of displacements that it can actually estimate. It also implies that the RANSAC-Flow pre-alignement stage (with off-the-shelf feature matching and Ransac) is crucial for the success of the matching process in general. 
We thus need to sample transformations $W$ within the range of the network capabilities. 
As a result, we construct the warps $W$ by sampling only homographies and TPS transformations from a Gaussian distribution. This is because the Affine-TPS transformations lead to larger geometric transformations and are more difficult to parametrize for a network very sensitive to the strength of geometric transformations. The Gaussian sampling gives more importance to transformations of small magnitudes, as opposed to the uniform sampling used for WarpC-GLU-Net. 

The homography and TPS transforms are sampled from a Gaussian distribution with standard deviation $\sigma_H=0.08$, which corresponds to a displacement of 24 pixels in an image size $s_r \times s_r = 300 \times 300$. We further integrate additional elastic transformations, which were shown beneficial to boost the network accuracy. 
We use the above sampling scheme for all three training stages.

\subsection{Implementation details} RANSAC-Flow only estimates the flow at one eight of the original image resolution. Loss computations is performed at image resolution, \ie $s \times s = 224 \times 224$, after upsampling the estimated flow field. 
Following the original work, we also compute training losses at the image resolution. 
The hyper-parameters used in the estimation of our visibility mask $\widehat{V}$ (eq.~\ref{eq:vis-mask} of the main paper) are set to $\alpha_1 = 0.01$ and $\alpha_2 = 0.5$.

For training, we use similar training parameters as in~\cite{RANSAC-flow}. As pre-processing, we scale the input network images to $\left [ 0, 1 \right ]$. 
During the first training stage, WarpC-RANSAC-Flow is trained with a batch size of 10 for 300k iterations. The learning rate is initially equal to $8.10^{-4}$, and halved after 200k iterations. 
For the second training stage, we train for 140k iteration with a constant learning rate of $4.10^{-4}$. Finally, the third training stages also uses an initial learning rate of $4.10^{-4}$ halved after 200k iterations, and comprises a total of 300k iterations. To weight the matchability regularization loss with respect to our warp consistency loss in the third stage, we use a constant factor of $0.6$ applied to the matchability loss.

\section{Training details for \\ WarpC-SemanticGLU-Net}
\label{sec-sup:semanticglunet}

Here, we first review the SemanticGLU-Net architecture as well as their original training strategy. We then provide additional details about our training strategy, resulting in WarpC-SemanticGLU-Net. 

\subsection{Details about SemanticGLU-Net}

\parsection{Architecture} SemanticGLU-Net is derived from GLU-Net~\cite{GLUNet}, with two architectural modifications, making it more suitable for semantic data. 
Specifically, the global feature correlation layer is followed by a consensus network~\cite{Rocco2018b}. The features from the different levels in the L-Net are also concatenated, similarly to~\cite{Jeon}. 

\parsection{Training strategy in original work} SemanticGLU-Net was originally trained using the same procedure as GLU-Net~\cite{GLUNet}. It is explained in Sec.~\ref{sec-sup:glunet}.

\subsection{WarpC-SemanticGLU-Net: \\ our training strategy}

\parsection{Training procedure} We only finetune on semantic data, from the original pretrained SemanticGLU-Net model, initialized with the weights provided by the authors. The VGG-16 feature backbone is initialized to the ImageNet weights and not further finetuned. We use our warp consistency loss (Sec.~\ref{sec:our-loss} of the main paper), where the visibility mask $\widehat{V}$ is directly included. Note that since SemanticGLU-Net is trained using solely the warp-supervision objective, the overall training of WarpC-SemanticGLU-Net does not use any flow annotations.

\parsection{Training dataset} We use the PF-Pascal~\cite{PFPascal} images as training dataset. Following the dataset split in~\cite{SCNet}, we partition the total 1351 image pairs into a training set of 735 pairs, validation set of 308 pairs and test set of 308 pairs, respectively. The 735 training images are augmented by mirroring, random cropping and exchanging the images in the pair. It leads to a total of 2940 training image pairs.

\parsection{Warps $W$ sampling} We resize the image pairs $(I, J)$ to $s_r \times s_r = 500 \times 500$, sample a dense flow $W$ of the same dimension and create $I'$. Each of the images of the resulting image triplet $(I, I', J)$ is then centrally cropped to $s \times s = 400 \times 400$. 
The flows $W$ are created by sampling homographies, TPS and Affine-TPS transformations with equal probability. 
For homographies and TPS, we use a uniform sampling scheme with a range equal to $\left [-\sigma_H, \sigma_H \right ]$, where $\sigma_H = 0.2$, which corresponds to a displacement of $100$px, in image size $s_r = 500$.
For the affine transformations, we also sample all parameters, \ie scale, translation, shear and rotation angles, from uniform distributions with ranges respectively equal to $\tau = 0.4$, $t = 0.25$, and $\alpha = \pi/12$ for both angles.
We compose the affine transformations with TPS transformations, for which we sample the translation magnitudes uniformly with a range $\sigma_{tps} = 0.2$, thus corresponding to a displacement of 100px. 

\parsection{Implementation details} For our network WarpC-SemanticGLU-Net, the weights in the training loss~\eqref{eq:multiscale-loss} are set to $\gamma_{1}=0.32, \gamma_{2}=0.08, \gamma_{3}=0.02, \gamma_{4}=0.01$. We train with a batch size of 5, for a total of 7k iterations. The learning rate is initially equal to $8.10^{-5}$, and halved after 4k, 5k and 6k iterations. The network is trained using Adam optimizer~\cite{adam} with weight decay of $0.0004$.

\section{Training details for method analysis}
\label{sec-sup:ablation-details}

For the method analysis corresponding to Sec.~\ref{sec:method-analysis} of the main paper, we use as base network GLU-Net~\cite{GLUNet}. Architecture description and implementation details are explained in Sec.~\ref{sec-sup:glunet}. In this section, for completeness we provide additional details about the training procedure used for each of the compared networks, when necessary.

\parsection{Warp consistency graph analysis} All networks are trained following the first WarpC-GLU-Net training stage, \ie without including the visibility mask in the bipath or cycle losses. We employ the same warps $W$ for all networks, which correspond to the sampling distribution used in the first training stage, as detailed in Sec.~\ref{sec-sup:glunet}.

\parsection{Ablation study} Networks in ablation study are trained according to the stages described in Sec.~\ref{sec-sup:glunet}.

\parsection{Comparison to alternative losses} We provide implementation details for networks trained with alternative losses. For all unsupervised learning objectives, we train the network in two stages. First, we use solely the evaluated loss, without visibility or occlusion mask. In the second stage, we further finetune the resulting model, extending the evaluated loss with the visibility mask, estimated as in~\cite{Meister2017}. For the objectives including our warp consistency loss (WarpC) or the warp-supervision loss, we use the same synthetic warp $W$ distribution than introduced in Sec.~\ref{sec-sup:glunet}. In the following, we give details about each training using an alternative loss. 

\begin{description}
\item[Warp-supervision + forward-backward:] Selecting a hyper-parameter is necessary to weight the forward-backward loss with respect to the warp-supervision objective. After manual tuning, we weight the forward-backward term with a constant factor equal to $0.05$. It ensures that the forward-backward term accounts for about half of the magnitude of the warp-supervision loss. 
For further implementation details, refer to  Sec.~\ref{sec-sup:glunet}. 
\item[Census:] The implementation details are the same than explained in Sec.~\ref{sec-sup:glunet}. Particularly, we found that downsampling the images to the flow resolution at each level for loss computation gave better results than upsampling the estimated flows to image resolution. 
\item[SSIM:] To compute the loss, we upsample the estimated flow from each level to image resolution, \ie $h \times w = 520 \times 520$ for the HNet and  $h_L \times w_L = 256 \times 256$ for the LNet. This strategy led to significantly better results than downsampling the images instead. As a result, because GLU-Net is a multi-scale architecture and the loss is computed using the flow from each resolution, the weights of the final training loss~\eqref{eq:multiscale-loss} are set to $\gamma_{1}=0.08, \gamma_{2}=0.08, \gamma_{3}=0.01, \gamma_{4}=0.01$. This gives equal contribution to all levels, since estimated flows at levels of L-Net and H-Net are upsampled to respectively $h_L \times w_L$ and $h \times w$.  
SSIM is computed with a window size of 11 pixels, following RANSAC-Flow~\cite{RANSAC-flow}. 
\item[SSIM + forward-backward:] The model trained using the SSIM loss is further finetuned with the combination of photometric SSIM and forward-backward consistency losses (eq.~\ref{eq:forward-backward} of the main paper). Both loss terms are balanced with a constant factor equal to $0.1$, applied to the forward-backward consistency term. It ensures that the forward-backward term accounts for about half of the magnitude of the SSIM loss. Implementation details are the same than when training with the SSIM loss only.
\item[SSIM + WarpC: ] For the WarpC loss, we follow the training procedure and implementation details provided in Sec.~\ref{sec-sup:glunet}, \ie we compute the loss at estimated flow resolution. For the SSIM loss term, we instead follow the training strategy explained above, \ie we compute the loss at image resolution.  For the WarpC term, the different levels weights of the final training loss~\eqref{eq:multiscale-loss} are set to be $\gamma_{1}=0.32, \gamma_{2}=0.08, \gamma_{3}=0.02, \gamma_{4}=0.01$, while for the SSIM loss term they are set to $\gamma_{1}=0.08, \gamma_{2}=0.08, \gamma_{3}=0.01, \gamma_{4}=0.01$. Each loss term, \ie WarpC and SSIM, is computed independently and the final loss is the sum of both. 
\item[Sparse ground-truth data:] Since the ground-truth is sparse, it is inconvenient to down-sample the ground-truth to different resolutions. We thus instead up-sample the estimated flow fields to the ground-truth resolution and compute the loss at this resolution. As for SSIM, we therefore use  $\gamma_{1}=0.08, \gamma_{2}=0.08, \gamma_{3}=0.01, \gamma_{4}=0.01$ for the level weights of the final training loss~\eqref{eq:multiscale-loss}. 
\end{description}

\section{Analysis of transformations W}
\label{sec-sup:transfo-analysis}
 
In this section, we analyse the impact of the sampled transformations' strength on the performance of the corresponding trained WarpC networks.
As explained in Sec.~\ref{sec-sup:transfo}, the strength of the warps $W$ is mostly controlled by the standard-deviation or range $\sigma_H$, used to sample the base homography and TPS transformations.
We thus analyse the effect of the sampling range $\sigma_H$ on the evaluation results of the corresponding WarpC networks, particularly WarpC-GLU-Net and WarpC-SemanticGLU-Net. 
We do not provide such analysis for WarpC-RANSAC-Flow because as mentioned in Sec.~\ref{sec-sup:RANSAC-flow}, RANSAC-Flow architecture is limited to a small range of displacements that it can estimate, which also limits the range $\sigma_H$ over which we can sample the warps $W$.

\begin{figure}[t]
\centering
\newcommand{\wid}{0.45\textwidth}
\includegraphics[width=0.98\columnwidth]{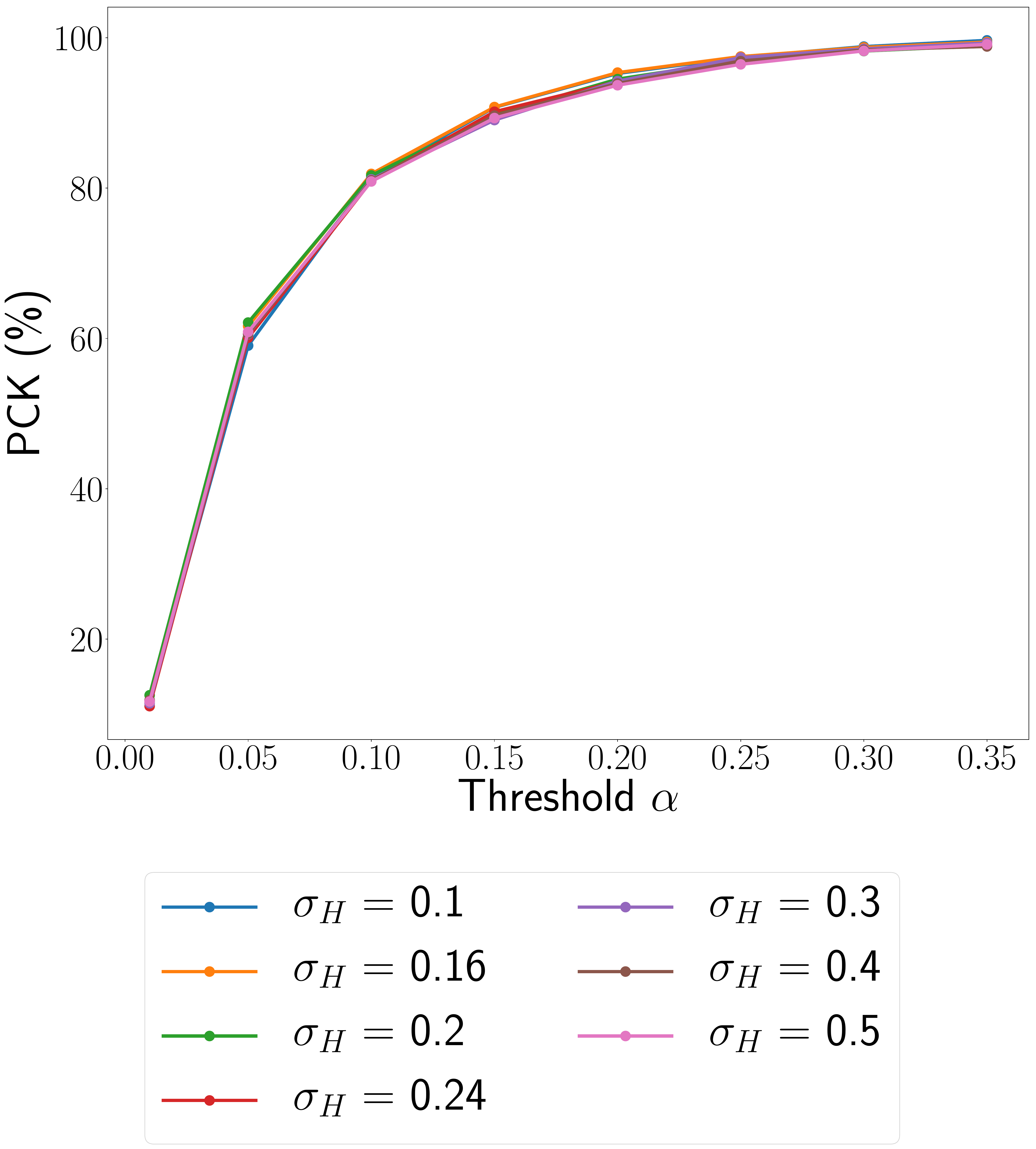}
\caption{PCK curves obtained on the PF-Pascal~\cite{PFPascal} images by WarpC-SemanticGLU-Net, for different sampling ranges $\sigma_H$ used to create the synthetic transformations $W$ during training.}
\label{Fig.:pck-pf-pascal}
\end{figure}

\begin{figure}[t]
\centering
\newcommand{\wid}{0.45\textwidth}
\includegraphics[width=0.98\columnwidth]{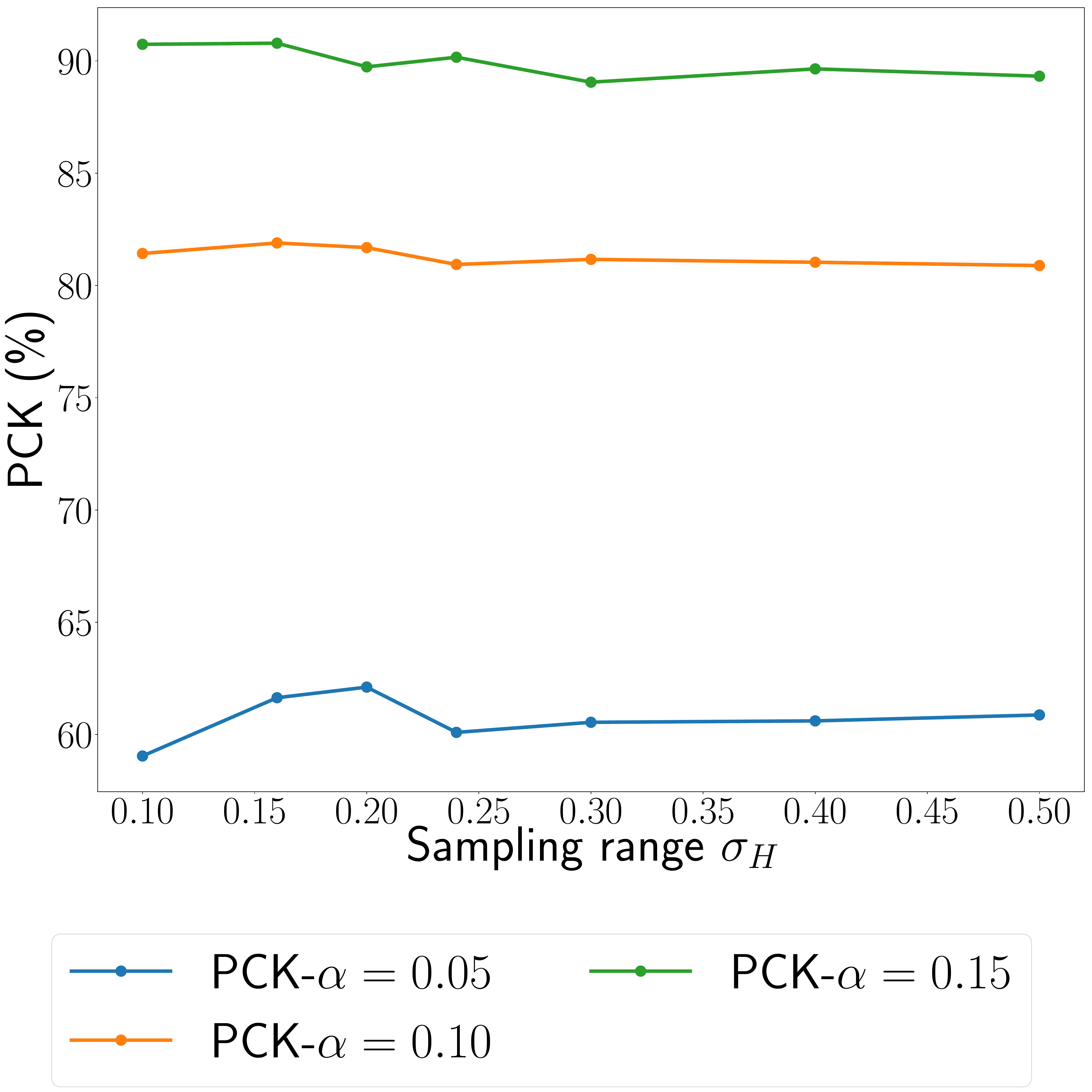}
\caption{PCK for $\alpha$ thresholds in $\left \{0.05, 0.1, 0.15 \right\}$ obtained on the PF-Pascal~\cite{PFPascal} images by WarpC-SemanticGLU-Net, for different sampling ranges $\sigma_H$ used to create the synthetic transformations $W$ during training.}
\vspace{-3mm}
\label{Fig.:pck-pf-pascal-2}
\end{figure}

\begin{figure*}[t]
\centering
\newcommand{\wid}{0.45\textwidth}
\subfloat[MegaDepth]{\includegraphics[width=\wid]{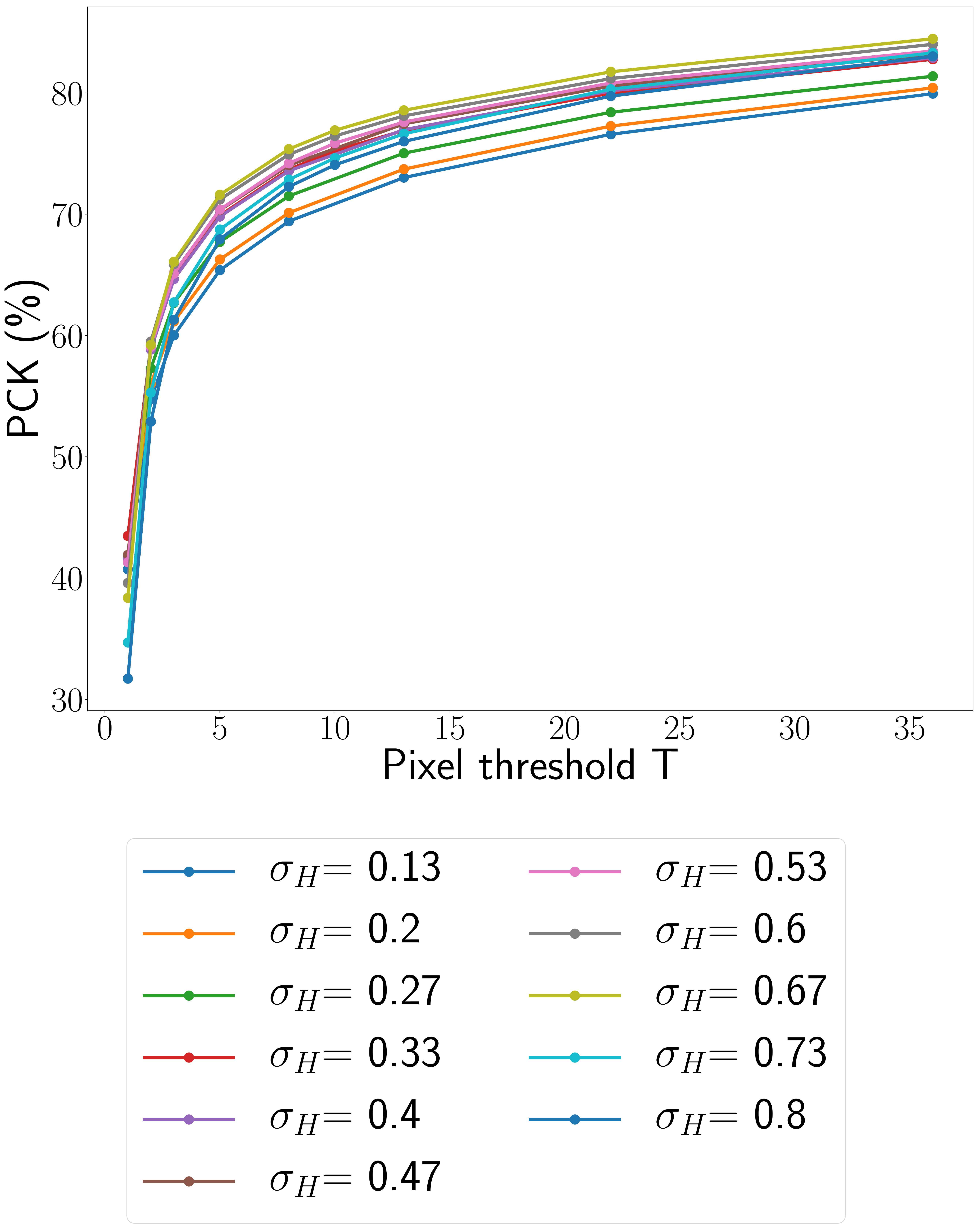}}
\subfloat[RobotCar]{\includegraphics[width=\wid]{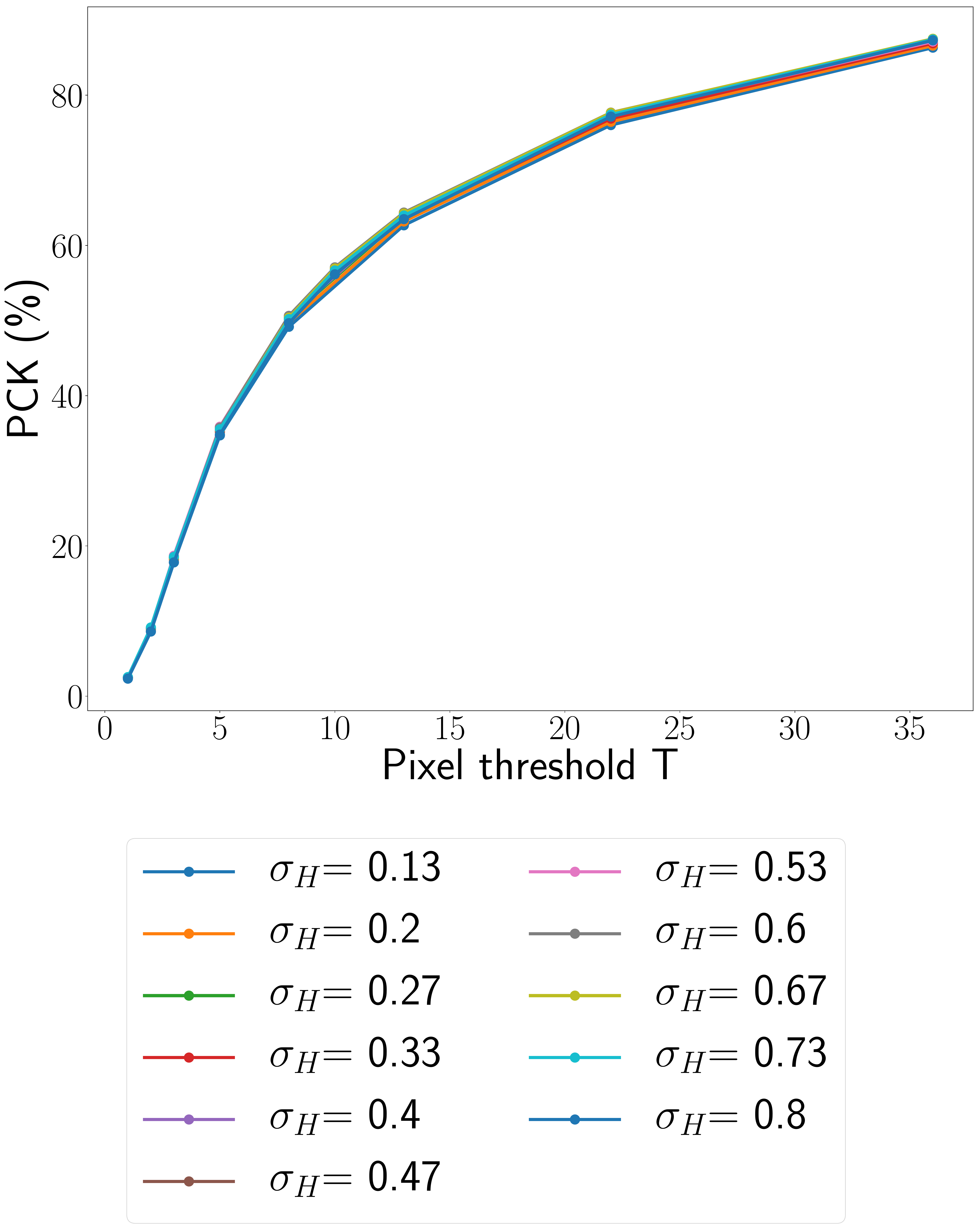}} 
\vspace{2mm}
\caption{PCK curves obtained by GLU-Net based networks trained using our warp consistency loss, for different sampling ranges $\sigma_H$. Transformations $W$ are sampled according to the first training stage.}
\label{Fig.:pck-megadepth-stage1-2}
\end{figure*}

\begin{figure*}[t]
\centering
\newcommand{\wid}{0.45\textwidth}
\subfloat[MegaDepth]{\includegraphics[width=\wid]{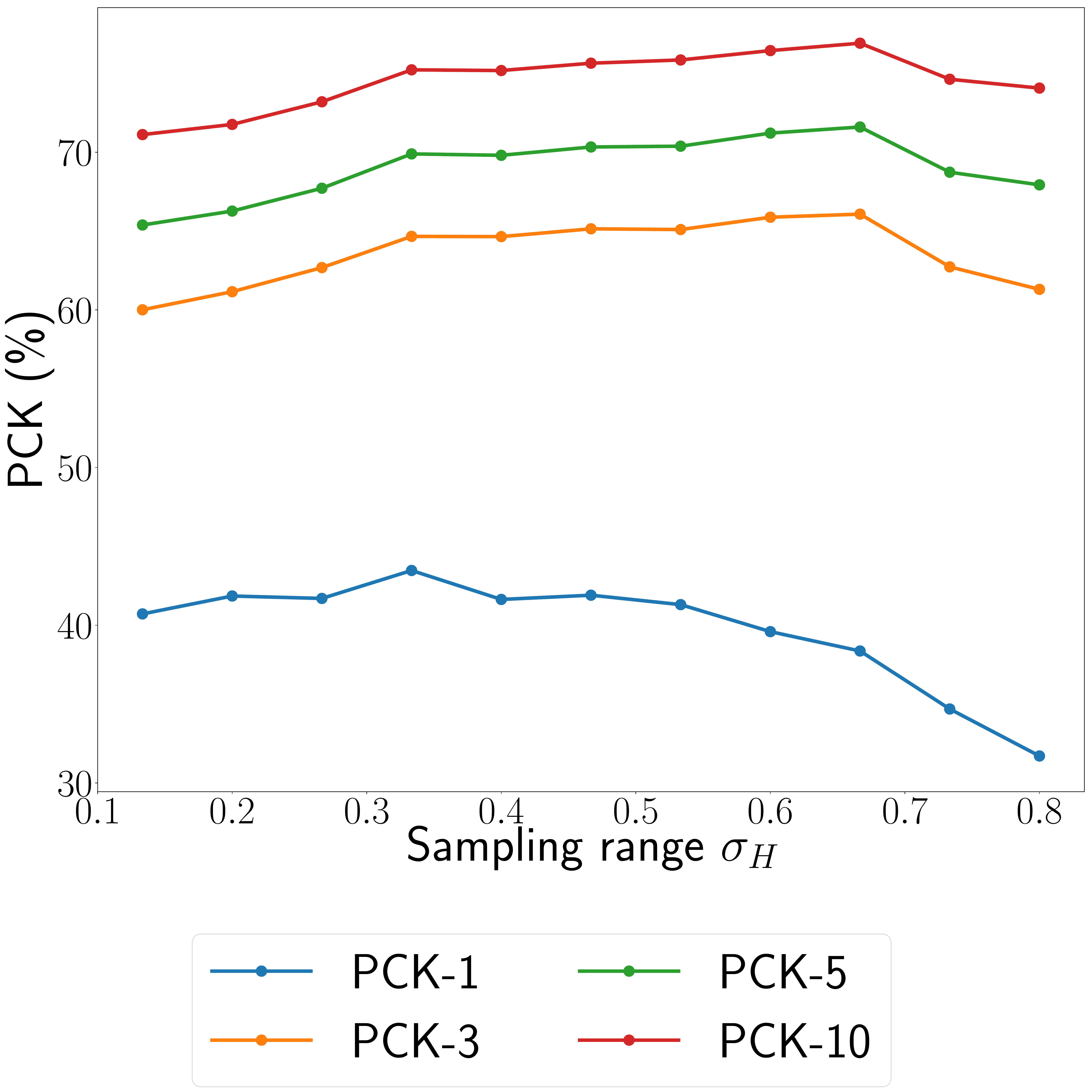}}
\subfloat[RobotCar]{\includegraphics[width=\wid]{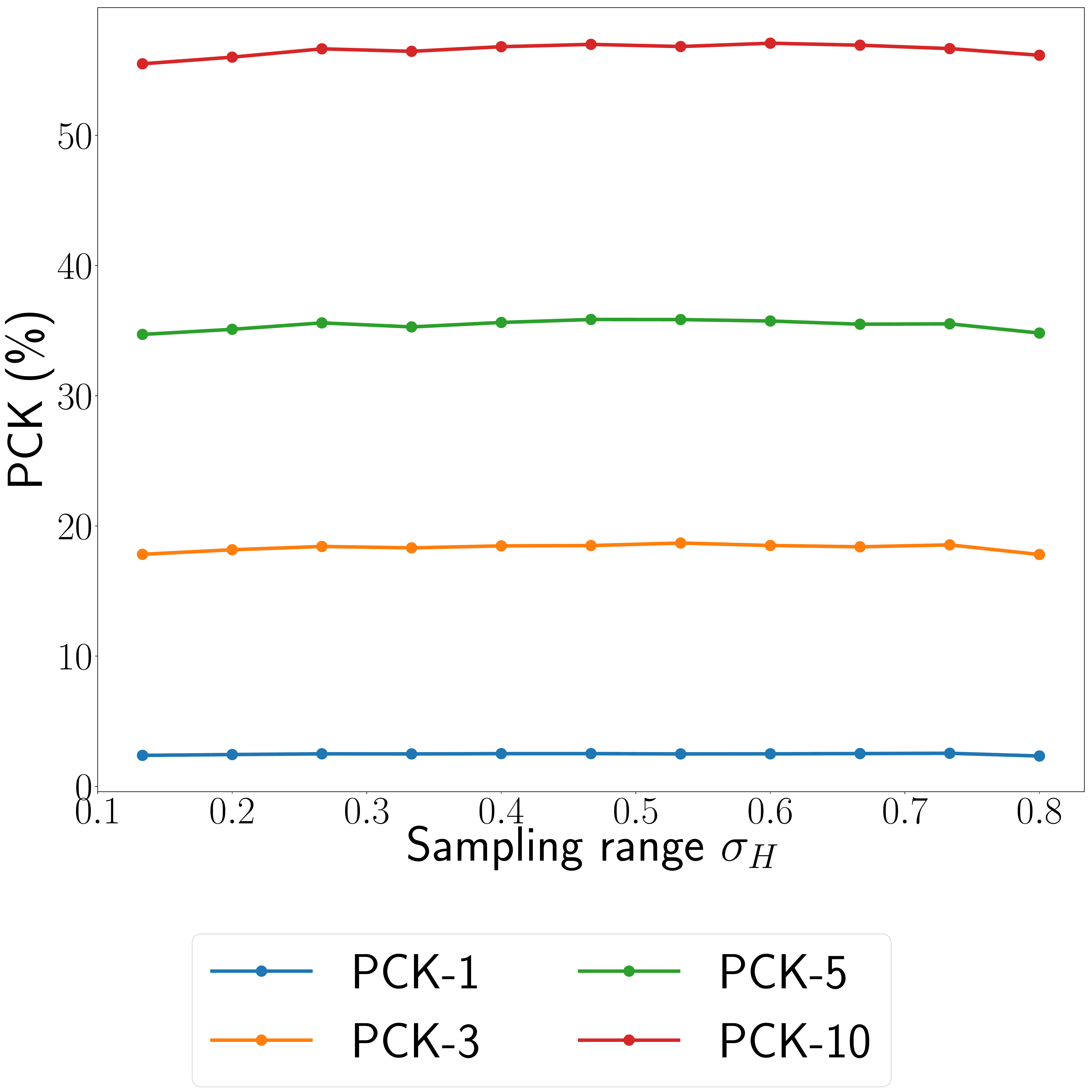}}
\vspace{2mm}
\caption{PCK for pixel thresholds in $\left \{1, 3, 5, 10 \right\}$ obtained by GLU-Net based networks trained using our warp consistency loss, for different sampling ranges $\sigma_H$. Transformations $W$ are sampled according to the first training stage.}
\label{Fig.:pck-robotcar-stage1-2}
\end{figure*}

While we choose a specific distribution $p_{W}$ to sample the transformations parameters used to construct the flow $W$, our experiments show that the performance of the trained networks according to our proposed warp consistency loss (Sec.~\ref{sec:our-loss} of the main paper) is relatively insensitive to the strength of the transformations $W$, if they remain in a reasonable bound. We present these experiments below.

\parsection{WarpC-GLU-Net} Specifically, we analyze the PCK curves obtained by GLU-Net based models, trained following our first training stage (Sec.~\ref{sec-sup:glunet}), for varying ranges $\sigma_H$ used to sample the TPS and homography transformations. Note that for all networks, the sampling distributions of the affine-tps transformations are the same. We plot in Fig.~\ref{Fig.:pck-megadepth-stage1-2} the resulting curves, computed on the MegaDepth and RobotCar datasets. 
For completeness, we additionally plot the PCK values for fixed pixel thresholds in $\left \{1, 3, 5, 10 \right \}$ versus the sampling range $\sigma_H$ in Fig.~\ref{Fig.:pck-robotcar-stage1-2}. 
On MegaDepth, increasing the sampling range $\sigma_H$ from $0.13$ to $0.67$ leads to an improvement of the resulting network's robustness to large geometric transformations, \ie an increase in PCK-3, 5 and 10. Further increasing $\sigma_H$ up to $0.8$ leads to a decrease in these PCK values. 
For PCK-1 however, networks trained with sampling ranges within $\left [ 0.13, 0.53 \right]$ obtain similar accuracy. The accuracy starts dropping for larger sampling ranges. 
We select $\sigma_H = 0.33$ because it obtains the best PCK-1 and good PCK-3, 5 and 10. Nevertheless, note that networks trained using sampling ranges within $\left [ 0.2, 0.53 \right]$ lead to relatively similar PCK metrics, within 2-3 \%.
Moreover, on RobotCar, all networks obtain similar PCK metrics, independently of the sampling range $\sigma_H$.

\parsection{WarpC-SemanticGLU-Net} As for WarpC-GLU-Net, we show that the performance of WarpC-SemanticGLU-Net is relatively insensitive to the strength of the transformations $W$, if they remain in a reasonable bound. 
Specifically, we analyze the PCK curves obtained by WarpC-SemanticGLU-Net based models, for varying ranges $\sigma_H$ used to sample the TPS and homography transformations of $W$ during training. Note that for all networks, the sampling distributions of the affine-tps transformations are the same. We plot in Fig.~\ref{Fig.:pck-pf-pascal} the resulting curves evaluated on the test set of PF-Pascal and in Fig.~\ref{Fig.:pck-pf-pascal-2} the results for specific PCK values. For sampling ranges $\sigma_{H}$ within $\left [ 0.1, 0.5 \right]$, the results of the corresponding trained WarpC-SemanticGLU-Net are all very similar overall. 
Particularly, the gap between all networks for $\alpha > 0.05$ is very small, within 1 \%. For $\alpha < 0.05$, differences amount to $4 \%$. We selected $\sigma_{H}=0.2$ because it led to a slightly better PCK for the low threshold $\alpha=0.05$.

\section{Time and memory efficiency} 
\label{sup-sec:time_memory}

\begin{figure}[t]
\centering%
\vspace{-1mm}
\newcommand{\wid}{0.22\textwidth}%
\includegraphics[width=0.38\textwidth]{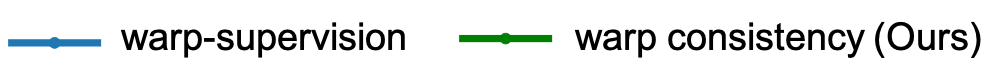} \\
\vspace{-5mm}
\subfloat{\includegraphics[width=\wid]{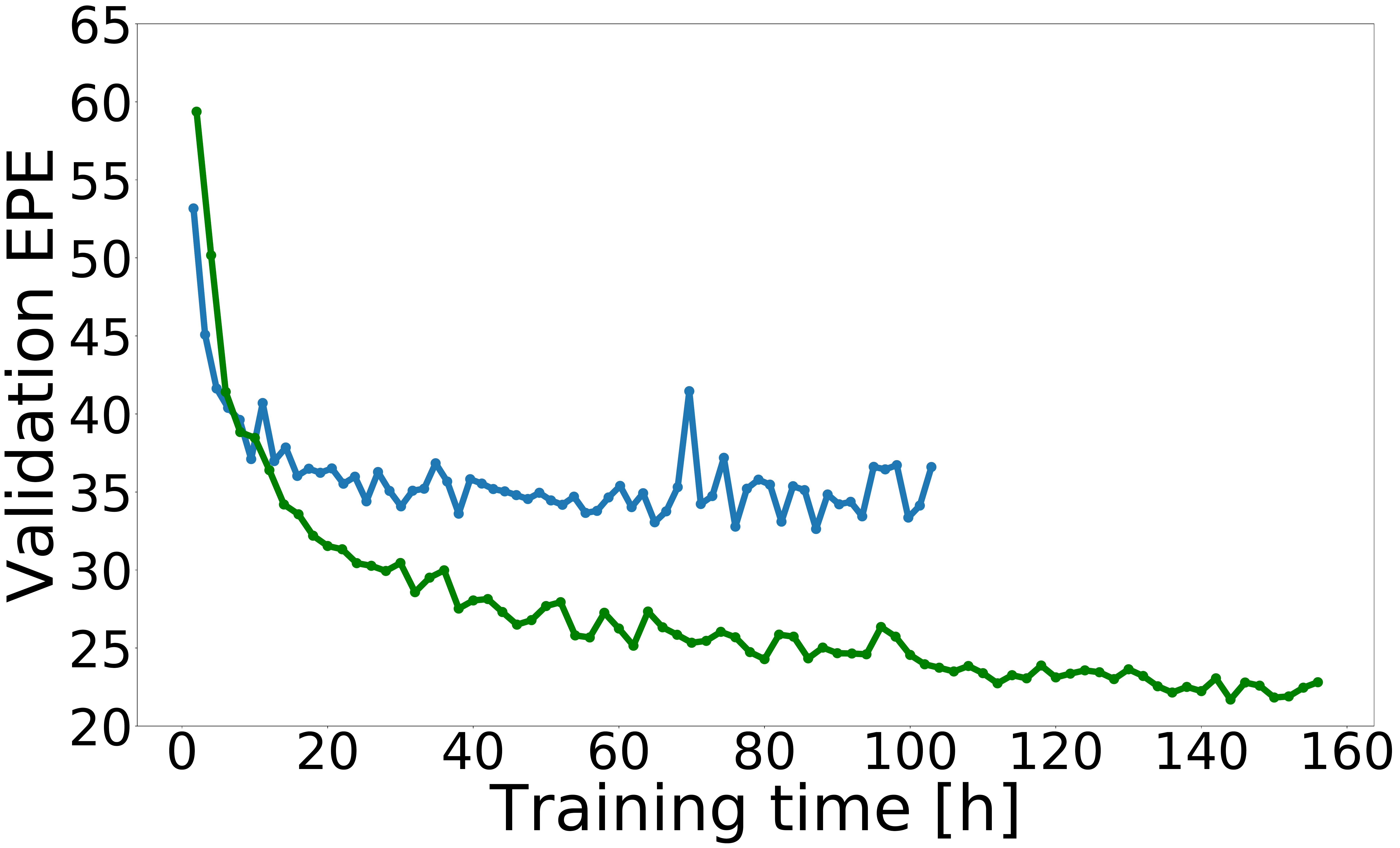}}~%
\subfloat{\includegraphics[width=\wid]{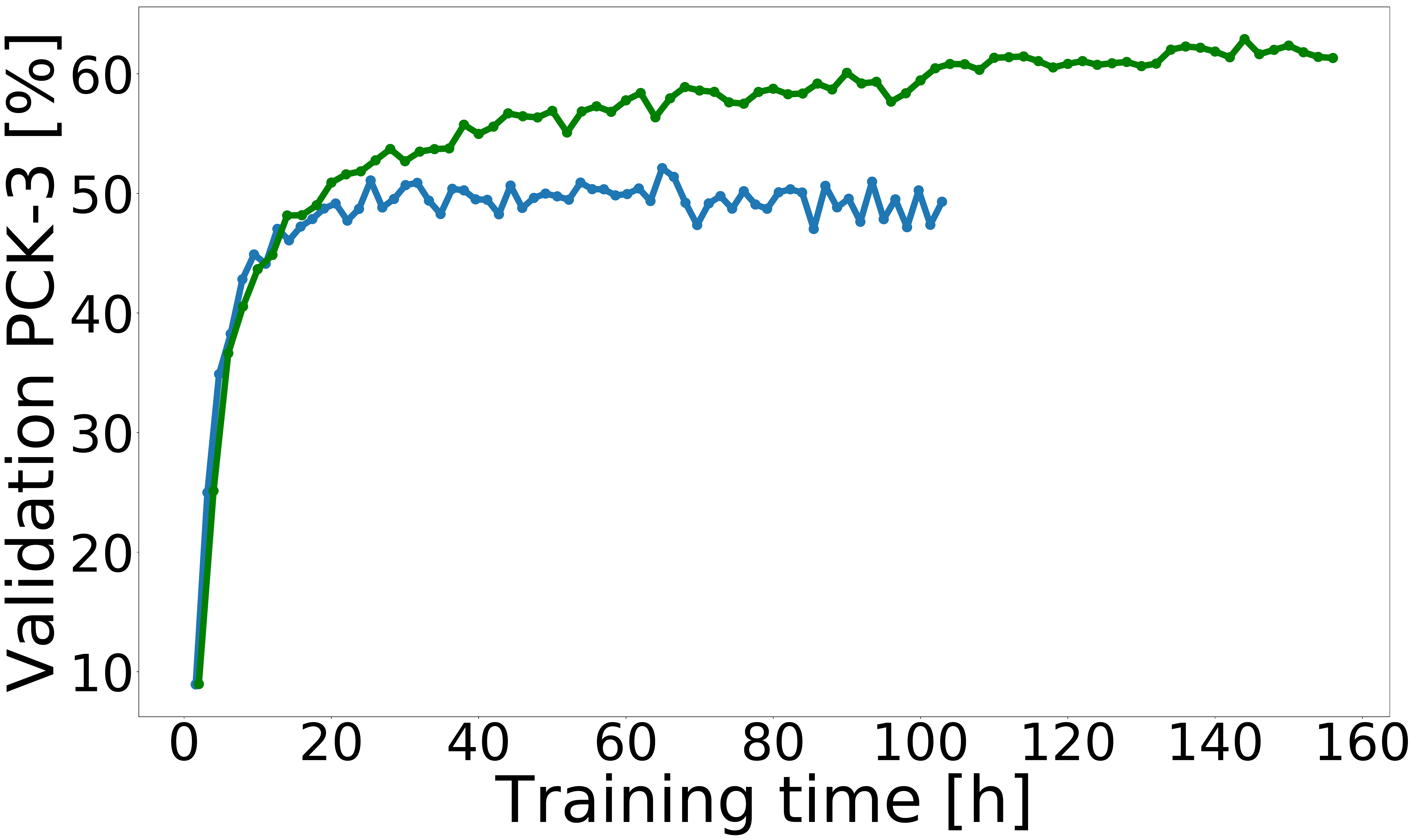}}~%
\vspace{-1mm}
\caption{Validation metrics \wrt training time for GLU-Net.}
\vspace{-3mm}
\label{fig:training-time}
\end{figure}

We here discuss the time and memory efficiency of our unsupervised approach during training and testing. 
Our loss formulation does not impact inference time and memory requirements, which remain at the baseline.
For WarpC-GLU-Net, on an RTX 2080Ti GPU with 11GB, we fit 10 image pairs for  warp-supervision during training and 6 image triplets for our objective. We use the same image size $520 \times 520$. The training times per pair/triplet are 190 ms for the former and 240 ms ($\times 1.26$) for ours. 
We visualize the convergence speed (validation performance) as a function of training time for both approaches in Fig.~\ref{fig:training-time}. While each iteration is slower, our warp consistency strategy leads to faster convergence and vastly better final performance. 
The overall training times of our models WarpC-GLU-Net, WarpC-RANSAC-Flow and WarpC-SemanticGLU-Net are 7, 4 and 1 days respectively. 

\section{Experimental setup and datasets}
\label{sec-sup:details-evaluation}

In this section, we first provide details about the evaluation datasets and metrics. We then explain the experimental set-up in more depth. 

\subsection{Evaluation metrics}

\parsection{AEPE} AEPE is defined as the Euclidean distance between estimated and ground truth flow fields, averaged over all valid pixels of the reference image.

\parsection{PCK} The Percentage of Correct Keypoints (PCK) is computed as the percentage of correspondences $\mathbf{\tilde{x}}_{j}$ with an Euclidean distance error $\left \| \mathbf{\tilde{x}}_{j} - \mathbf{x}_{j}\right \| \leq  T$, w.r.t.\ to the ground truth $\mathbf{x}_{j}$, that is smaller than a threshold $T$.

\subsection{Evaluation datasets and set-up}
\label{details-eval-data}

\parsection{HPatches} The HPatches dataset~\cite{Lenc} is a benchmark for geometric matching correspondence estimation. It depicts planar scenes, with transformations restricted to homographies.  As in DGC-Net~\cite{Melekhov2019}, we only employ the 59 sequences labelled with \verb|v_X|, which have viewpoint changes, thus excluding the ones labelled \verb|i_X|, which only have illumination changes. Each image sequence contains a query image and 5 reference images taken under increasingly larger viewpoints changes, with sizes ranging from $450 \times 600$ to $1613 \times 1210$. 

\parsection{MegaDepth} The MegaDepth dataset~\cite{megadepth} depicts real scenes with extreme viewpoint changes. No real ground-truth correspondences are available, so we use the result of SfM reconstructions to obtain sparse ground-truth correspondences. We follow the same procedure and test images than~\cite{RANSAC-flow}, spanning 19 scenes. More precisely, 1600 pairs of images were randomly sampled, that shared more than 30 points. The test pairs are from different scenes than the ones we used for training and validation. Correspondences were obtained by using 3D points from SfM reconstructions and projecting them onto the pairs of matching images. It results in approximately 367K correspondences. During evaluation, following~\cite{RANSAC-flow}, all the images and ground-truths are resized to have minimum dimension 480 pixels. 

\parsection{RobotCar} Images in RobotCar depict outdoor road scenes, taken under different weather and lighting conditions. While the image pairs show similar view-points, they are particularly challenging due to their many textureless regions. 
For evaluation, we use the correspondences originally introduced by~\cite{RobotCarDatasetIJRR}. Following~\cite{RANSAC-flow}, all the images and ground-truths are resized to have minimum dimension 480 pixels.

\parsection{TSS} The TSS dataset~\cite{Taniai2016} contains 400 image pairs, divided into three groups: FG3DCAR, JODS, and PASCAL, according to the origins of the images. The dense flow fields annotations for the foreground object in each pair is provided along with a segmentation mask. Evaluation is done on 800 pairs, by also exchanging query and reference images. Evaluation is done by computing PCK for a pixel threshold computed with respect to query image size.

\parsection{PF-Pascal} The PF-PASCAL~\cite{PFPascal} benchmark is built from the PASCAL 2011 keypoint annotation dataset~\cite{BourdevM09}. It consists of 20 diverse object categories, ranging from chairs to sheep. Sparse manual annotations are provided for 300 image pairs. Evaluation is done by computing PCK for a pixel threshold computed with respect to query image size.

\parsection{PF-Willow} The PF-WILLOW dataset consists of 900 image pairs selected from a total of 100 images~\cite{PFWillow}. It spans four object categories. Sparse annotations are provided for all pairs.  For evaluation, we report the PCK scores with multiple thresholds ($\alpha$ = 0.05, 0.10, 0.15) with respect to bounding box size in order to compare with prior methods.

\section{Additional method analysis experiments}
\label{sec-sup:ablation}

In this section, we extend the method analysis corresponding to Sec.~\ref{sec:method-analysis} of the main paper. 

\begin{table}[t]
\centering
\resizebox{\columnwidth}{!}{%
\begin{tabular}{lccc|ccc|cc} \toprule
& \multicolumn{3}{c}{\textbf{MegaDepth}} & \multicolumn{3}{c}{\textbf{RobotCar}} & \multicolumn{2}{c}{\textbf{HPatches}} \\
& PCK-1  & PCK-5 & PCK-10  & PCK-1 & PCK-5 & PCK-10  & AEPE & PCK-5 \\ \midrule
Stage1, pretrained VGG16  & 43.47 & \textbf{69.90} & \textbf{75.23} & \textbf{2.49} & \textbf{35.28} & \textbf{56.45} & 22.83 & 78.60 \\ 
Stage1, from scratch & \textbf{43.74} & 68.21 & 73.07 & 2.36 & 34.14 & 54.76 & \textbf{22.75} & \textbf{81.73} \\
\midrule
Stage2, pretrained VGG16  &  50.61 & \textbf{78.61} & \textbf{82.94} & \textbf{2.51} & \textbf{35.92} & \textbf{57.44} & \textbf{21.00} & 83.24 \\
Stage2, from scratch & \textbf{51.16} & 77.64 & 81.86 & 2.43 & 35.12 & 56.53 & 21.22 & \textbf{83.83}\\
\bottomrule
\end{tabular}%
}\vspace{1mm}\caption{Feature backbone training for both training stages of WarpC-GLU-Net.}\label{Tab.:comparison-feature-training}\vspace{-4mm}
\end{table}

\parsection{Feature backbone training} We train WarpC-GLU-Net from scratch, including the VGG-16 feature backbone, without initializing it with the pre-trained VGG-16 weights on ImageNet. We compare this version to the one using the pre-trained VGG-16 weights, which are fixed during both training stages. 
In general, both networks achieve similar results, allowing us to use ImageNet initialization to reduce overall training time.
In the first training stage, training the feature backbone from scratch leads to a slightly better accuracy on MegaDepth (PCK-1) as compared to the pre-trained VGG-16 version. However, the resulting model is somewhat less robust to large displacements, evidenced by the lower PCK-10 results. On RobotCar, the network with feature training also obtains slightly worst results, which could be due to the fact that the version using pre-trained weights on ImageNet saw more image diversity. However, training from scratch leads to better performance on HPatches, with a significant $+3 \%$ in PCK-5. In the second training stage, the trend is the same but the gap between the two network trainings further reduces.

\begin{table}[b]
\centering
\vspace{-3mm}
\resizebox{\columnwidth}{!}{%
\begin{tabular}{llccc|ccc|cc} \toprule
& & \multicolumn{3}{c}{\textbf{MegaDepth}} & \multicolumn{3}{c}{\textbf{RobotCar}} & \multicolumn{2}{c}{\textbf{HPatches}} \\
& & PCK-1  & PCK-5 & PCK-10  & PCK-1 & PCK-5 & PCK-10  & AEPE & PCK-5 \\ \midrule
I & SSIM \eqref{eq:photo-loss} & 51.93 & 69.58 & 71.58 & 2.18 & 31.48 & 51.65 & 38.62 & 62.61 \\

II & SSIM + f-b \eqref{eq:photo-loss}+\eqref{eq:forward-backward} & 52.59 & 70.78 & 72.78 & 2.12 & 31.86 & 52.13 & 35.79 & 64.48  \\
III & SSIM + smoothness + f-b & 55.00 & 71.24 & 73.10 & 2.42 & 34.76 & 55.75 & 38.13 & 66.46 \\
IV & Warp-superv. \eqref{eq:warp-supervision} & 38.51 & 60.33 & 66.57 & 2.30 &  33.21 & 54.19 & 26.88 & 78.07 \\ 
V & Warp-superv. + SSIM \eqref{eq:warp-supervision}+\eqref{eq:photo-loss} & \textbf{56.58} & 73.81 & 75.69 & 2.27 & 33.05 & 53.97 & 29.50 & 71.34  \\ 
VI & \textbf{WarpC} \eqref{eq:warp-supervision}+\eqref{eq:vis-w-bipath} &  50.61 & \textbf{78.61} & \textbf{82.94} & \textbf{2.51} & \textbf{35.92} & \textbf{57.44} & \textbf{21.00} & \textbf{83.24} \\
VII & \textbf{WarpC} + SSIM & 55.82 & 74.89 & 77.08 &   2.38 & 34.56 & 55.50 & 26.04 & 72.44\\
\bottomrule
\end{tabular}%
}\vspace{1mm}\caption{Additional comparison and combination of alternative losses.
}\label{Tab.:comparison-loss-extention}
\end{table}

\begin{figure*}[t]
\centering%
\vspace{-4mm}
\subfloat[]{\includegraphics[width=0.48\textwidth]{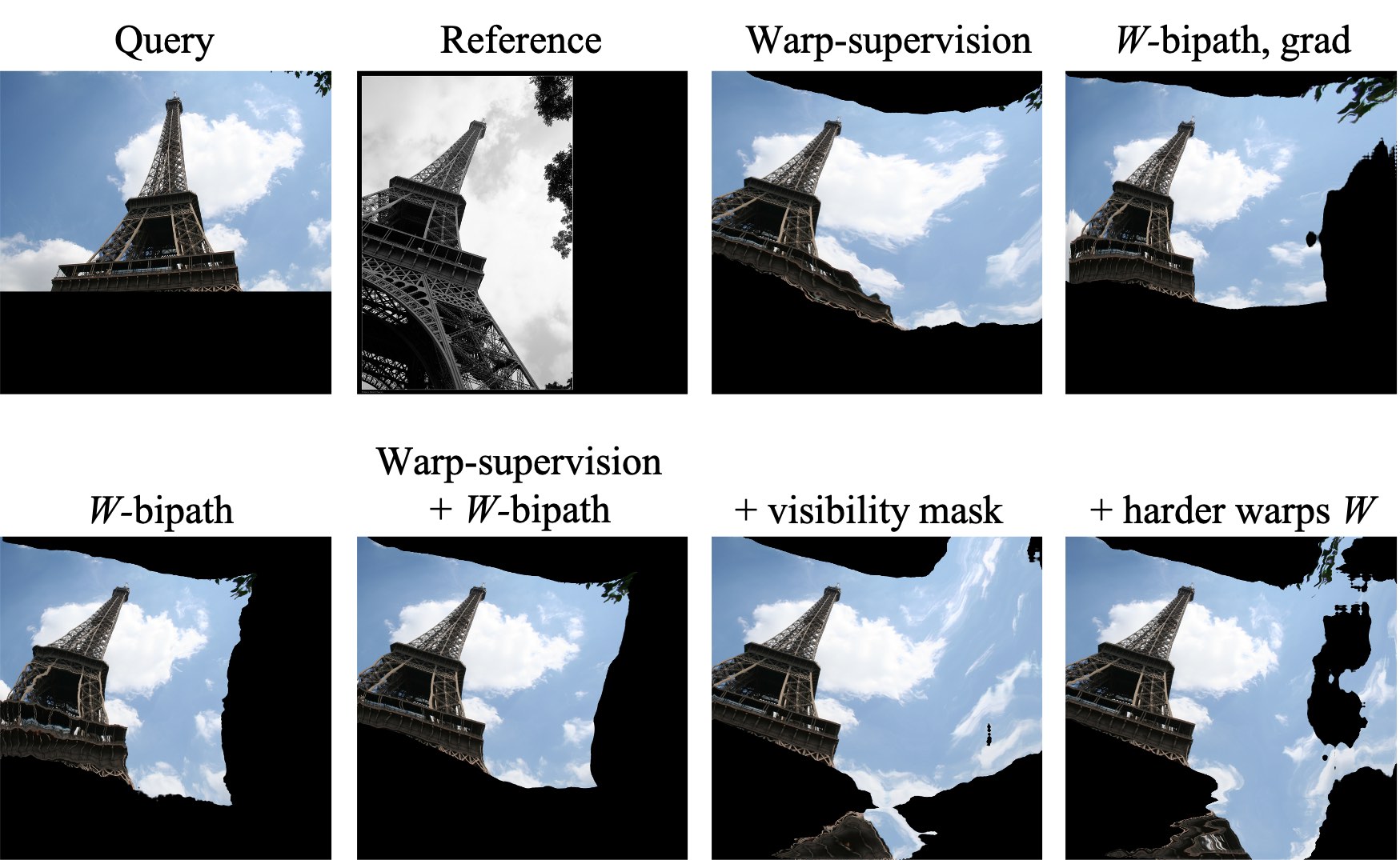}}  \hspace{3mm}
\subfloat[]{\includegraphics[width=0.48\textwidth]{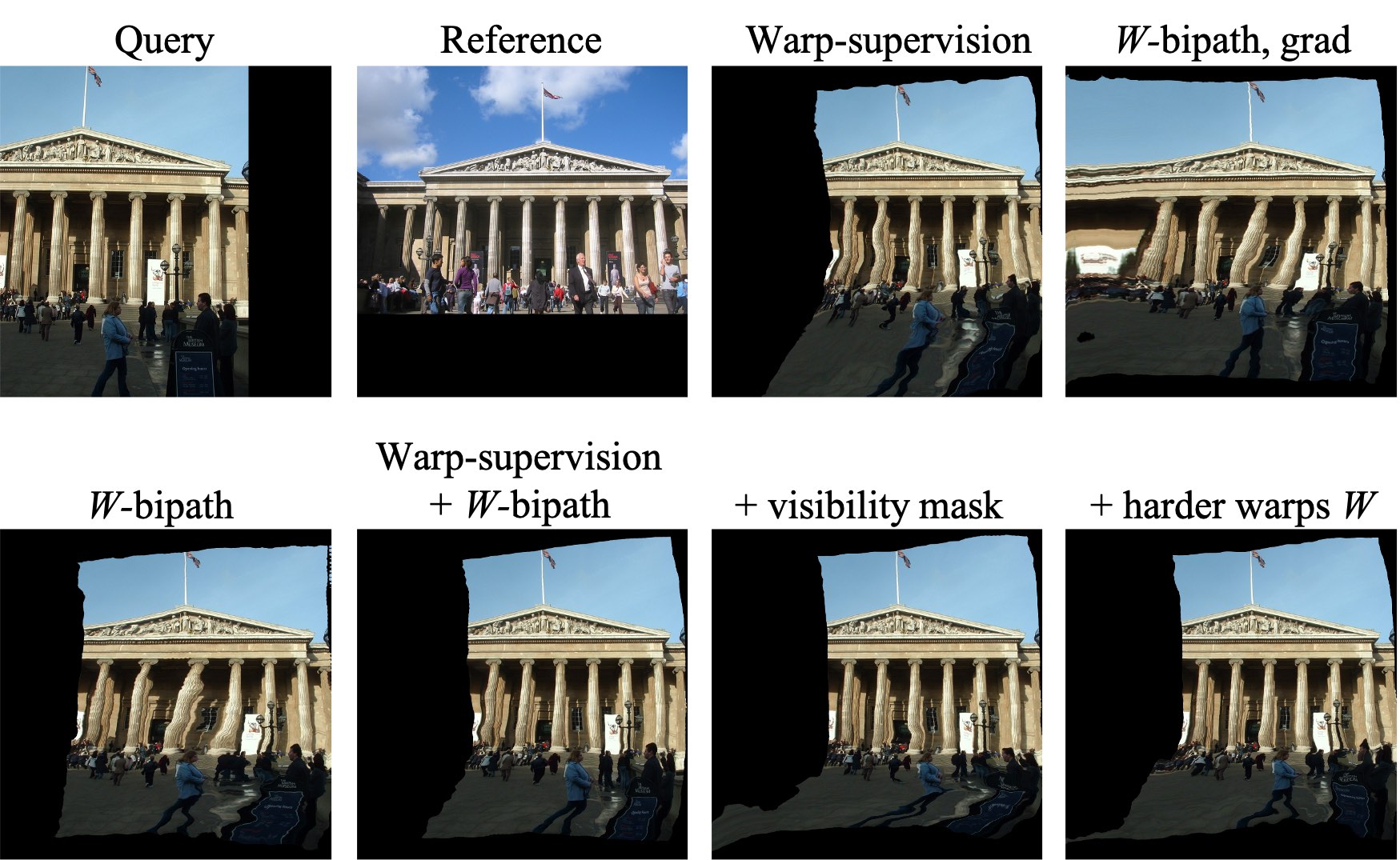}}  \\

\subfloat[]{\includegraphics[width=0.5\textwidth]{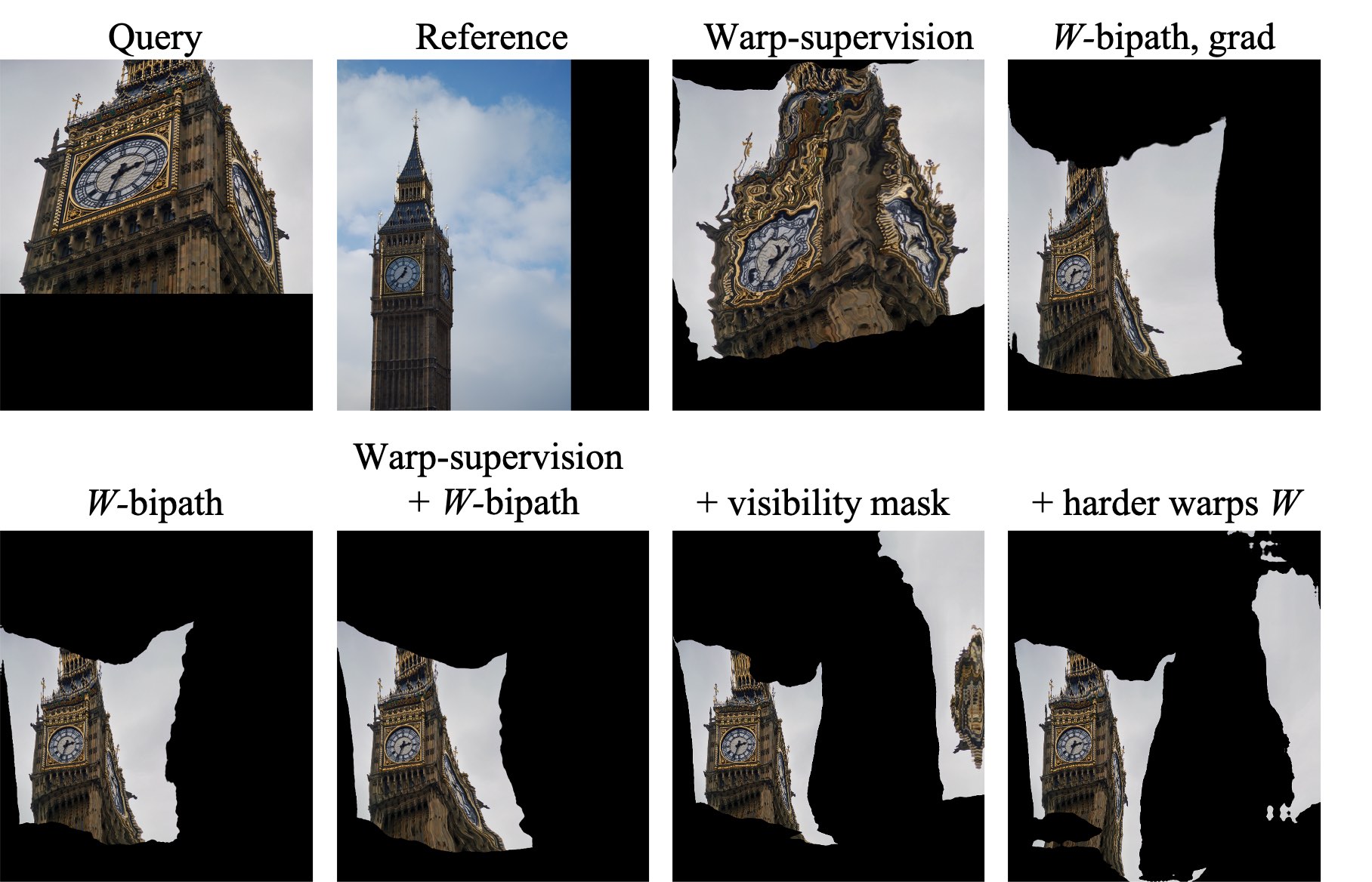}}  
\subfloat[]{\includegraphics[width=0.5\textwidth]{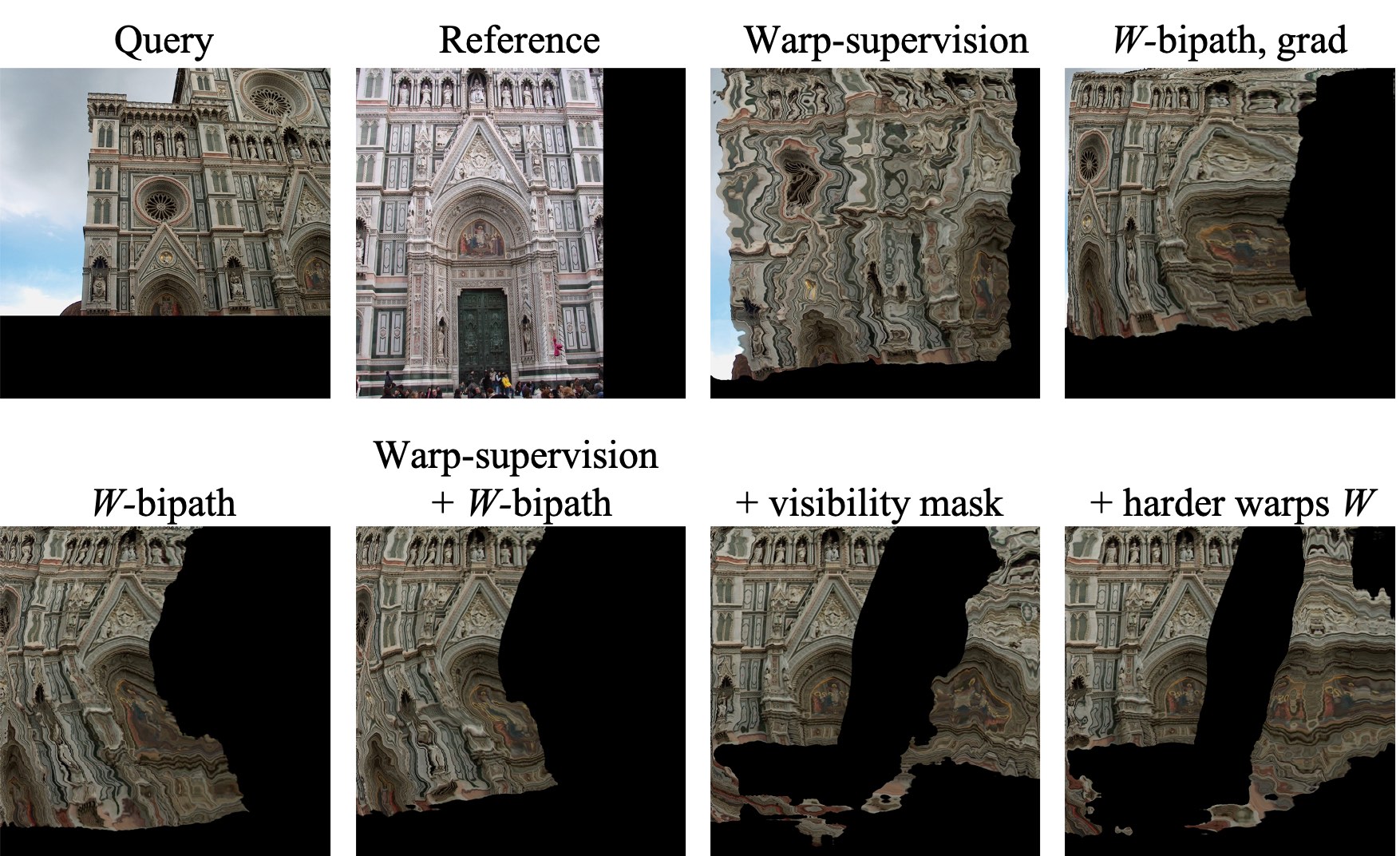}}  \\ 
\vspace{1mm}
\caption{Qualitative ablation study of our unsupervised training approach, using GLU-Net as base network. We employ images of the MegaDepth dataset. Note that in the dense estimation settings, the network must predict a match for every pixels in the reference, even in obviously occluded regions. Only correspondences found in overlapping regions are relevant nevertheless.}
\vspace{-4mm}
\label{Fig.:ablatio-quali}
\end{figure*}

\parsection{Additional comparison to alternative unsupervised losses} We here provide additional comparison to other unsupervised losses. We also evaluate more combinations of losses. Results are reported in Tab.~\ref{Tab.:comparison-loss-extention}, which extends Tab.~\ref{tab:comparison-loss} of the main paper. 
For completeness, we train a version of GLU-Net using standard unsupervised losses in the literature, \ie photometric (SSIM~\cite{WangBSS04} here), forward-backward (eq.~\ref{eq:forward-backward} of the main paper) and smoothness loss in (III). For the smoothness loss, we use a first order smoothness constraint, following~\cite{BackToBasics}. Adding the smoothness loss leads to an improvement compared to only SSIM (I) or the combination of SSIM and forward-backward (II), particularly on the RobotCar dataset. Nevertheless, it is still significantly lower than our proposed unsupervised approach (VI) for all metrics and datasets, except for PCK-1 on MegaDepth. 
Moreover, it also obtains lower metrics than the combination of our warp consistency loss with the photometric SSIM loss (VII) on the MegaDepth and HPatches dataset. The RobotCar dataset depicts scenes with little geometric transformations but large appearance variations in the form of seasonal or day-time changes for example (see Fig.~\ref{Fig.:robotcat}). As a result, the photometric consistency is strongly violated on those images, while a smoothness loss is beneficial, which explains that the combination of the three classical unsupervised losses (III) leads to slightly better results than the combination of our proposed approach and the photometric SSIM loss (VII). Nevertheless, our proposed warp consistency approach (VI) alone outperforms all other methods on RobotCar. 

We also train the combination of warp-supervision loss and SSIM (V). It leads to an improvement compared to SSIM (I) for all dataset and metrics. Nevertheless, on RobotCar and HPatches, the improvement brought by the warp-supervision loss in (V) is significantly lower than the increase brought by combining our proposed warp consistency loss (WarpC) with SSIM in (VII). On MegaDepth, WarpC combined with SSIM (VII) achieves better performance than warp-supervision and SSIM (V) for PCK-5 and PCK-10, for a slightly lower performance on sub-pixel accuracy (PCK-1). This confirms that the warp consistency loss enables to handle the large appearance and geometric changes present between real image pairs, while the warp-supervision loss mostly focuses on getting accuracy to small displacements. 
Moreover, note that both combinations of SSIM with warp-supervision (V) or WarpC (VII) obtain worse results than our warp consistency loss (WarpC) in (VI) for all datasets and thresholds, except for PCK-1 on MegaDepth.

\parsection{Qualitative ablation study} Next, we show qualitative results of our ablation study, corresponding to Tab.~\ref{tab:ablation-study} of the main paper. We warp the queries according to the flows estimated by GLU-Net networks trained with different losses, and present the corresponding qualitative results in Fig.~\ref{Fig.:ablatio-quali}.  Note that in the dense estimation settings, the network is obligated to predict a match for every pixels in the reference, even in obviously occluded regions. Occluded regions can be filtered out using \eg a forward-backward consistency mask~\cite{Meister2017}, or by letting the network predict a visibility mask as in~\cite{RANSAC-flow, Melekhov2019}.

On image pairs (c) and (d), the superiority of the \w loss as opposed to the warp-supervision loss is obvious. The warp-supervision loss, solely relying on synthetic training image pairs, is not equipped to handle the large and complex 3D motions present in these example pairs. On the contrary, the \w constraint benefits from direct supervision to improve the predictions between real image pairs during training. 

The benefit of not back-propagating the gradients through the estimated flow used in the warping operation ('\w, grad' refers to version with back-propagation) is best illustrated in examples (b) and (c). It leads to a generally more accurate and stable flow predictions. The \w loss version without back-propagation is used as default for the rest of the section, unless otherwise stated. 

Combining the warp-supervision with the \w loss drives the network to be more accurate. It is particularly visible on image pairs (a) and (b). On both these examples, combining the warp-supervision with the \w loss results in a more stable and accurate estimated flow. 

\begin{figure}[t]
\centering%

\includegraphics[width=0.48\textwidth]{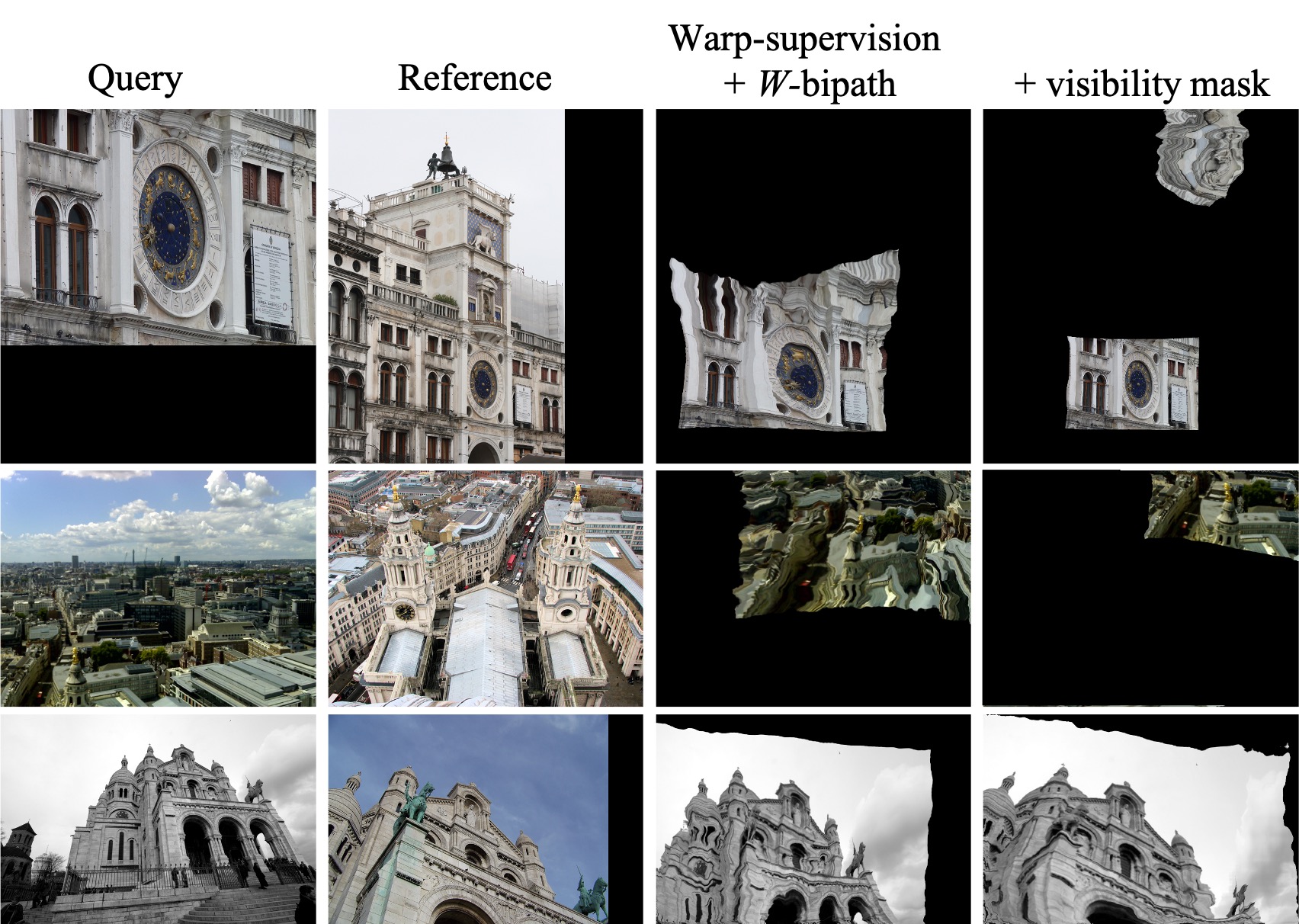}
\caption{Impact of including the visibility mask (eq.~\ref{eq:vis-mask} of the main paper) in the \w training loss (eq.~\ref{eq:vis-w-bipath} of the main paper).}
\label{Fig.:ablation-vis-mask}
\vspace{-3mm}
\end{figure}

The impact of extending the \w objective with our visibility mask (eq.~\ref{eq:vis-w-bipath} of the main paper) is well illustrated on examples (c) and (d). In the former, training with the visibility mask permits to 'clean' the estimated flow and produces a much more accurate prediction. In (d), it allows to get the correct overall geometric transformations and removes most of the shakiness present for previous networks. In general, introducing the visibility mask is a crucial step, which enables the network to better deal with very large geometric variations, such as drastic scale or view-point changes, as visualised in Fig.~\ref{Fig.:ablation-vis-mask}. 

Finally, training using harder warps $W$ leads to improved accuracy to small details, as evidenced in example (b), where the columns in the warped query appear straighter.

\parsection{Method analysis on semantic data} For completeness, we also empirically analyze and decompose our proposed warp consistency loss, when trained and evaluated on semantic data. We follow the training procedure detailed in Sec.~\ref{sec-sup:semanticglunet}.
In Tab.~\ref{Tab.:method-analysis-semantic}, we show results on the TSS~\cite{Taniai2016}, PF-Pascal~\cite{PFPascal} and PF-Willow~\cite{PFWillow} datasets. 
From the pre-trained SemanticGLU-Net, further finetuning on PF-Pascal using solely the warp-supervision objective improves upon SemanticGLU-Net on all datasets and metrics. 
Using the \w loss instead leads to slightly better performance on TSS images, but drastic improvement on PF-Willow and PF-Pascal. This is because image pairs of those two datasets are generally much harder and ambiguous than for TSS images, and therefore benefit more from the supervision on real image pairs provided by our warp consistency objective during training. 
Further combining both objectives (WarpC-SemanticGLU-Net) leads to a substantial improvement on TSS for similar performances than solely the \w loss on PF-Pascal. The combination with the warp-supervision objective also results in an additional boost in accuracy for the lowest threshold $\alpha = 0.05$ on PF-Willow. For reference, we visualize image examples and the performance of WarpC-SemanticGLU-Net on TSS in Fig.~\ref{Fig.:TSS_CARS},~\ref{Fig.:TSS_JODS},~\ref{Fig.:TSS_PASCAL}, and on PF-Pascal in Fig.~\ref{Fig.:intro_PF_1},~\ref{Fig.:intro_PF_2} ,~\ref{Fig.:intro_PF_3}. 
Also note that training with the photometric SSIM loss diverges, because it cannot handle the radical appearance and shape variations between multiple instances of the same object class.

\begin{table}[t]
\centering
\resizebox{\columnwidth}{!}{%
\begin{tabular}{lc|cc|cc} \toprule
& \multicolumn{1}{c}{\textbf{TSS}} & \multicolumn{2}{c}{\textbf{PF-Pascal}} & \multicolumn{2}{c}{\textbf{PF-Willow}}  \\
&  Avg. & $\alpha\!=\!0.05 $ & $\alpha\!=\!0.1$ & $\alpha\!=\!0.05 $ & $\alpha\!=\!0.1$  \\ \midrule
SemanticGLU-Net & 82.8 & 46.0 & 70.6 & 36.4 & 63.8 \\
Warp-superv. (eq. 3 of m.p) & 85.2  & 48.8 & 72.4 & 39.7 & 67.5\\ 
\w (eq. 8 of m.p) & 85.5 &   \textbf{62.9} & \textbf{82.0} & 47.1 & \textbf{75.4} \\
\w + Warp-sup. (\textbf{WarpC}) & \textbf{87.2} & 62.1 & 81.7 & \textbf{49.0} & 75.1 \\ 
\bottomrule
\end{tabular}%
}\vspace{1mm}\caption{Warp consistency graph analysis on semantic data.
}\label{Tab.:method-analysis-semantic}

\end{table}

\section{Detailed results}
\label{sec-sup:results}

In this section, we provide additional results for pose estimation using RANSAC-Flow as base network. We also further evaluate WarpC-GLU-Net on the HPatches dataset~\cite{Lenc} and compare it to state-of-the-art methods. 
We then provide more detailed results on the PF-Pascal semantic dataset~\cite{PFPascal}. We additionally give evaluation results on the PF-Willow dataset~\cite{PFWillow} and the SPair-71K dataset~\cite{spair}. 
We also show the possible extension of our unsupervised training approach to the optical flow task. 
Finally, we present extensive qualitative results on multiple geometric and semantic matching datasets. 

\subsection{Results on pose estimation}

Since RANSAC-Flow predicts a matchability mask along with the dense correspondences, we additionally evaluate both jointly for the task of pose estimation. Specifically, we follow the standard set-up of~\cite{OANet} and evaluate on 4 scenes of the YFCC100M dataset~\cite{YFCC}, each comprising 1000 image pairs. 

\parsection{YFCC100M} The YFCC100M dataset represents touristic landmark images. The ground-truth poses were created by generating 3D reconstructions from a subset of the collections~\cite{heinly2015_reconstructing_the_world}. We use the evaluation procedure introduced in RANSAC-Flow~\cite{RANSAC-flow}. In particular, the original images and ground-truths are resized to have a minimum dimension of 480.

\begin{table}[b]
\centering
\resizebox{0.95\columnwidth}{!}{%
\begin{tabular}{lccc}
\toprule
             &  mAP @5\textdegree &  mAP @10\textdegree & mAP @20\textdegree  \\ \midrule
Superpoint~\cite{superpoint} & 30.50 & 50.83 & 67.85  \\ 
SIFT~\cite{SIFT} & 46.83 & 68.03 & 80.58 \\        
D2D~\cite{D2D} & 55.58 & 66.79 & -  \\ \midrule  
RANSAC-Flow~\cite{RANSAC-flow} (SegNet) & \textbf{63.48} & \textbf{72.93} & \textbf{81.59} \\
\textbf{WarpC-RANSAC-Flow} (SegNet)& 62.90 & 72.48 & 81.56 \\ \midrule  
RANSAC-Flow~\cite{RANSAC-flow} &  30.93 & 38.20 & 46.88  \\
\textbf{WarpC-RANSAC-Flow} & \textbf{61.85} & \textbf{71.24} & \textbf{79.86} \\
\bottomrule
\end{tabular}%
}\vspace{1mm}
\caption{Two-view geometry estimation on YFCC100M~\cite{YFCC}. Including an additional segmentation network (SegNet) makes the overall training supervised. }
\label{Tab.:YCCM}
\end{table}

\newcommand{\best}[1]{\textcolor{BrickRed}{\textbf{#1}}}
\newcommand{\second}[1]{\textcolor{NavyBlue}{\textit{#1}}}
\begin{table*}[t]
\centering
\vspace{-4mm}\resizebox{0.98\textwidth}{!}{%
\begin{tabular}{lll|cccc|ccc|ccc}
\toprule
 & &  & \multicolumn{4}{c}{\textbf{TSS}} & \multicolumn{3}{c}{\textbf{PF-Pascal}} & \multicolumn{3}{c}{\textbf{PF-Willow}} \\
 & &  & \multicolumn{4}{c}{PCK @ $\alpha_{img}$} & \multicolumn{3}{c}{PCK @ $\alpha_{img}$} & \multicolumn{3}{c}{PCK @ $\alpha_{bbox}$} \\
 
Supervision & Methods  & Features & FG3DCar & JODS & Pascal & Avg. & $\alpha=0.05$ &  $\alpha=0.10$  &  $\alpha=0.15$ & $\alpha=0.05$ &  $\alpha=0.10$  &  $\alpha=0.15$ \\ \midrule
segmentation mask & SF-Net~\cite{SFNet} & ResNet-101 & - & - & - & - & 53.6 & \second{81.9} & \second{90.6} & 46.3 & 74.0 & 84.2 \\ \midrule
warp-supervision & CNNGeo(S)~\cite{Rocco2017a} & ResNet-101 & 90.1 & 76.4 & 56.3 & 74.4  & 41.0 & 69.5 & 80.4 & 36.9 & 69.2 & 77.8\\
(synthetic pairs) & GLU-Net~\cite{GLUNet} & VGG-16   &    93.2    &   73.3  &   71.1    &    79.2   & 42.2 & 69.1 & 83.1 &  30.4 & 57.7 & 72.9 \\ 
& GLU-Net-GOCor~\cite{GOCor} & VGG-16  & 94.6 &  77.9 &  77.7 & 83.4 & 36.6 & 56.8 & - & - & - & - \\ 
& A2Net~\cite{SeoLJHC18} & ResNet-101 &  - & - & - & -  & 42.8 &  70.8 & 83.3 & 36.3 & 68.8 & 84.4\\ \midrule

image-level & WeakAlign~\cite{Rocco2018a}& ResNet-101 & 90.3  & 76.4 & 56.5  & 74.4 & 49.0 & 74.8 & 84.0 & 38.2 & 71.2 & 85.8 \\ 
labels & RTNs~\cite{Kim2018} & ResNet-101 & 90.3  & 76.4 & 56.5  & 74.4 & 55.2 & 75.9 & 85.2 & 41.3 & 71.9 & 86.2 \\
& PARN~\cite{Jeon} & ResNet-101 & 89.5  & 75.9 & \second{71.2}  & 78.8  & - & - & - & - & - & - \\
& NC-Net~\cite{Rocco2018b} &  ResNet-101 & 94.5 & 81.4 & 57.1 & 77.7  & 54.3 & 78.9 & 86.0 & 33.8 & 67.0 & 83.7 \\ 
& DCCNet~\cite{DCCNet}  & ResNet-101 & 93.5  & \second{82.6} & 57.6  & 77.9  & 55.6 & \best{82.3} & 90.5 &43.6 & 73.8 & 86.5  \\
& DHPF~\cite{MinLPC20} & ResNet-101 &   - & - & - & - & 56.1 & 82.1 & \best{91.1} & \best{50.2} & \best{80.2} & \best{91.1}\\ 
& SAM-Net~\cite{Kim2019} & VGG-19  & \second{96.1} & 82.2 & 67.2 & 81.8 & \second{60.1} & 80.2 & 86.9 & - & - &  - \\ \midrule

warp-supervision & Semantic-GLU-Net~\cite{GLUNet} & VGG-16 & 94.4 & 75.5  & 78.3 & \second{82.8}  & 46.0 & 70.6 & 83.3 & 36.4 & 63.8 & 78.4\\
image-level labels & \textbf{WarpC-SemanticGLU-Net} (Ours) & VGG-16 & \best{97.1} & \best{84.7} & \best{79.7} & \best{87.2} & \best{62.1} & 81.7 & 89.7 & \second{49.0} & \second{75.1} & \second{86.9} \\ 
\bottomrule
\end{tabular}%
}\vspace{1mm}
\caption{PCK [\%] obtained by different state-of-the-art unsupervised methods on the TSS~\cite{Taniai2016}, PF-Pascal~\cite{PFPascal} and PF-Willow~\cite{PFWillow} datasets for the task of semantic matching. 
Results from~\cite{Rocco2017a, SeoLJHC18, Rocco2018a, Kim2018, Rocco2018b, DCCNet} are from~\cite{MinLPC20}. 
Best results are highlighted in red, while second best are in blue. We compare our approach WarpC-SemanticGLU-Net to methods specifically and exclusively designed for semantic data, trained unsupervised. On the contrary, our proposed warp consistency loss (Sec.~\ref{sec:our-loss} of the main paper) offers a general formulation, applicable to multiple tasks, including geometric and semantic matching. In the last section of the table, we highlight the improvement brought by our unsupervised finetuning. }
\label{Tab.:suppl-TSS}
\end{table*}

\begin{table}[b]
\centering
\resizebox{0.95\columnwidth}{!}{%
\begin{tabular}{lccc}
\toprule
             &  AEPE $\downarrow$   & PCK-1 (\%) $\uparrow$ & PCK-5 (\%) $\uparrow$  \\ \midrule
DGC-Net~\cite{Melekhov2019} &  33.26 & 12.00   & 58.06  \\
GLU-Net~\cite{GLUNet}  &   25.05 &  39.55   &  78.54  \\ 
GLU-Net-GOCor~\cite{GOCor} &  \textbf{20.16} & \textbf{41.55} & 81.43 \\
GLU-Net* &  25.04 & 39.37 & 78.60   \\
WarpC-GLU-Net &  21.00 & 41.00 & \textbf{83.24} \\
\bottomrule
\end{tabular}%
}\vspace{1mm}
\caption{HPatches homography dataset~\cite{Lenc}.}
\label{Tab.:geo-match-HP}
\end{table}

\parsection{mAP} For the task of pose estimation, we use mAP as the evaluation metric, following~\cite{OANet}. The absolute rotation error $\left | R_{err}  \right |$ is computed as the absolute value of the rotation angle needed to align ground-truth rotation matrix $R$ with estimated rotation matrix $\hat{R}$, such as
\begin{equation}
    R_{err} = cos^{-1}\frac{Tr(R^{-1}\hat{R}) -1}{2} \;,
\end{equation} 
where operator $Tr$ denotes the trace of a matrix. The translation error $T_{err}$ is computed similarly, as the angle to align the ground-truth translation vector $T$ with the estimated translation vector $\hat{T}$. 
\begin{equation}
    T_{err} = cos^{-1}\frac{T \cdot \hat{T}}{\left \| T \right \|\left \| \hat{T} \right \|} \;,
\end{equation}
where $\cdot$ denotes the dot-product. The accuracy Acc-$\kappa$ for a threshold $\kappa$ is computed as the percentage of image pairs for which the maximum of $T_{err}$ and $\left | R_{err}  \right |$ is below this threshold. mAP is defined according to original implementation~\cite{OANet}, \ie mAP @5\textdegree\, is equal to Acc-5, mAP @10\textdegree\,  is the average of Acc-5 and Acc-10, while mAP @20\textdegree\,  is the average over Acc-5, Acc-10, Acc-15 and Acc-20. 

\parsection{Results} RANSAC-Flow infers the dense flow field relating an image pair, coupled with a predicted matchability mask, both trained unsupervised. Pose estimation on YFCC100M evaluates the performance of the predicted flow and mask jointly. Indeed, for pose estimation, confidence or mask prediction is \emph{crucial} in order to select the accurate matches from the dense flow output and further use them to compute the pose.
Results on YFCC100M are presented in Tab.~\ref{Tab.:YCCM}. 
In the original RANSAC-Flow work, results are only reported when using an additional semantic segmentation network (SegNet) to better filter unreliable correspondences, in \eg sky. However, using a segmentation networks makes the overall method supervised. We therefore present results without any segmentation network, purely relying on RANSAC-Flow outputs. 
Without this additional segmentation, the performance of RANSAC-Flow is drastically reduced. In contrast, WarpC-RANSAC-Flow, trained with our unsupervised approach (Sec.~\ref{sec-sup:RANSAC-flow}), can directly estimate highly robust and generalizable matchability masks. The predicted masks of RANSAC-Flow and our approach WarpC-RANSAC-Flow are visually compared in Fig.~\ref{Fig.:YFCC}, in yellow and red respectively.
Crucially, training with a photometric objective does not permit to identify unreliable matching regions such as the sky, that fit the brightness constancy assumption of the photometric objective. These regions, when included for pose estimation computation, will drastically reduce the performance of the network. On the other hand, our proposed unsupervised objective enables to identify accurate matching regions and to filter out outliers or unreliable regions, leading to drastically better results.

\subsection{Results on HPatches}

We here present results of WarpC-GLU-Net against baseline GLU-Net* and state-of-the-art methods on the geometric matching homography dataset HPatches~\cite{Lenc} in Tab.~\ref{Tab.:geo-match-HP}. Our approach WarpC-GLU-Net scores second in AEPE and PCK-1, after the recent GLU-Net-GOCor, which uses the GOCor module~\cite{GOCor} in replacement to the feature correlation layer. We could also use our unsupervised learning approach to train GLU-Net-GOCor and further benefit from the improvement brought by GOCor. 
WarpC-GLU-Net is nevertheless significantly better than GLU-Net and baseline GLU-Net*, which both use a warp-supervision training loss.

\subsection{Additional results on semantic matching} 
\label{subsec-sup:semantic-results}

In this section, we present additional results on semantic data. While our proposed warp supervision objective offers a general training approach, applicable to multiple tasks such as semantic as well as geometric matching, we here compare it to methods specifically and exclusively designed for semantic data. 

\parsection{PF-Pascal and PF-Willow} In Tab.~\ref{Tab.:suppl-TSS}, we extend Tab.~\ref{tab:TSS} of the main paper, by showing PCK results for the threshold $\alpha=0.15$ on the PF-Pascal dataset. We additionally report evaluation results on the PF-Willow dataset~\cite{PFWillow}. 

On the PF-Pascal data, our proposed approach WarpC-SemanticGLU-Net ranks first for the lowest threshold $\alpha = 0.05$. For the second and third thresholds, it is marginally behind the current state-of-the-art DCC-Net~\cite{DCCNet} and DHPF~\cite{MinLPC20} (0.6 \% for $\alpha=0.1$ and 1.4 \% for $\alpha=0.15$). Note however, that these networks use a much stronger and deeper pre-trained feature backbone, namely ResNet-101, while we employ a VGG-16 backbone. 
On PF-Willow, WarpC-SemanticGLU-Net ranks second for all thresholds, shortly after the recent DHPF~\cite{MinLPC20}. 
Nevertheless, it obtains substantially better results than all other methods excluding DHPF, especially for the lowest threshold $\alpha=0.05$ with a notable improvement of $+2.7 \%$ compared to next best approach. We also note that DHPF predicts a cost volume as final output whereas we infer the dense flow field relating an image pair. On the TSS dataset, where dense ground-truth flows are available, our approach outperforms all previous approaches by a large margin.

Importantly, as highlighted in the last section of the table, our unsupervised finetuning leads to an impressive improvement compared to original SemanticGLU-Net: $+16.1\%$, $+11.1\%$ and $+6.4\%$ for thresholds $\alpha=\left\{0.05, 0.1, 0.15\right\}$ on the PF-Pascal dataset, and $+12.6\%$, $+11.3\%$ and $+8.5\%$ on the PF-Willow dataset for the same thresholds. As a result, while we chose a relatively weak baseline on these datasets, our unsupervised finetuning makes the resulting model very competitive, achieving first or second best metrics overall on four PCK thresholds out of six. 
Moreover, any other baseline could be used instead, benefiting from our unsupervised warp consistency finetuning. 

\parsection{SPair-71k} We also evaluated our unsupervised approach on the SPair-71k benchmark~\cite{spair}. It includes 70,958 image pairs of 18 object categories from PASCAL 3D+~\cite{6836101} and PASCAL VOC 2012~\cite{EveringhamGWWZ10}, providing 12,234 pairs for testing. This benchmark is more challenging than other datasets~\cite{PFPascal, PFWillow, Taniai2016} for semantic correspondence evaluation, as it covers significantly large variations of viewpoint, scale, truncation and occlusion. For the evaluation metric, we used the PCK with respect to the object bounding box and $\alpha = 0.1$. 
We finetuned our WarpC-Semantic-GLU-Net with our unsupervised warp consistency strategy on the training set of SPair-71K. We compare to other unsupervised approaches trained or finetuned on the same data in Tab.~\ref{tab:spair}. For comparison, we also further finetuned the original Semantic-GLU-Net~\cite{GLUNet} on SPair-71k with the warp-supervision objective. 
Our WarpCSemantic-GLU-Net is competitive with other unsupervised approaches. As before, while the Semantic-GLU-Net baseline architecture appears quite weak on the SPair-71K images, note the significant improvement (+9.2 \%) brought by our unsupervised warp consistency finetuning, as opposed to simple warp-supervision. 
We believe that training another stronger baseline would further improve results.

\begin{table}[t]
\centering
\caption{PCK for $\alpha=0.1$ with respect to object bounding box on SPair-71k~\cite{spair}.
We compare to unsupervised approaches trained or finetuned on the training set of Spair-71k~\cite{spair}. }
\vspace{1mm}
\resizebox{0.45\textwidth}{!}{%
\begin{tabular}{ll|c}
\toprule
Methods & Feature backbone & PCK @ $\alpha_{bbox}$ [\%]  \\
\midrule
CNNGeo~\cite{Rocco2017a} & ResNet-101 & 20.6 \\
WeakAlign~\cite{Rocco2018a}  & ResNet-101 &  20.9 \\
A2Net~\cite{SeoLJHC18} & ResNet-101 & 22.3 \\ 
NC-Net~\cite{Rocco2018b}  & ResNet-101 & 20.1 \\ 
DHPF~\cite{MinLPC20}  & ResNet-101 & \textbf{27.7} \\
SF-Net~\cite{SFNet}  & ResNet-101 & 26.5 \\
SemanticGLU-Net~\cite{GLUNet} (warp-sup.) & VGG-16 & 14.3   \\
\textbf{WarpCSemanticGLU-Net}  & VGG-16 & 23.5 \\
\bottomrule
\end{tabular}%
}
\label{tab:spair}
\end{table}

\subsection{Extension to optical flow data}

Here, we show the generalization capabilities of our unsupervised approach for the optical flow task. 
We report results on KITTI and MPI Sintel in Tab.~\ref{tab:optical-flow} by comparing our approach (WarpC-GLU-Net) with the baseline (GLU-Net*) trained using only warp-supervision. We evaluate according to the standard metrics, namely AEPE and Fl for KITTI and AEPE and PCKs for Sintel. Note that we use the \emph{same weights} as in the paper, which are trained for the dense geometric matching task on the MegaDepth training set, and therefore not well suited for the optical flow setting. Still, WarpC-GLU-Net obtains significantly better results than GLU-Net*, showing the benefit of our unsupervised training, even when trained for a different domain. We believe that unsupervised training on road-scene videos, such as KITTI raw, would further improve the results. 

\begin{table}[b]
\centering
\caption{Results on the training splits of KITTI and Sintel.}
\vspace{1mm}
\resizebox{0.49\textwidth}{!}{%
\begin{tabular}{l@{~~}c@{~~}c|@{~~}c@{~~}c@{~~}|@{~~}c@{~~}c@{~~}|@{~~}c@{~~}c}
\toprule
             & \multicolumn{2}{c}{\textbf{KITTI-2012}} & \multicolumn{2}{c}{\textbf{KITTI-2015}} & \multicolumn{2}{c}{\textbf{Sintel Clean}} & \multicolumn{2}{c}{\textbf{Sintel Final}}\\ 
  & AEPE   & F1   (\%)& AEPE              & Fl  (\%)   & AEPE   & PCK-1  (\%) & AEPE    & PCK-1  (\%)  \\ \midrule
GLU-Net*  &  3.37 & 17.38 & 10.90 & 36.06 & 5.74 & 69.47 & 6.60 & 61.33 \\
WarpC-GLU-Net  &  \textbf{3.09} & \textbf{16.32} & \textbf{9.35} & \textbf{33.65} &  \textbf{5.23} & \textbf{70.86} & \textbf{6.30} & \textbf{62.83}  \\
\bottomrule
\end{tabular}%
}
\label{tab:optical-flow}
\end{table} 

\subsection{Qualitative results}

\begin{figure*}[t]
\centering%
\vspace{-4mm}
\includegraphics[width=0.90\textwidth]{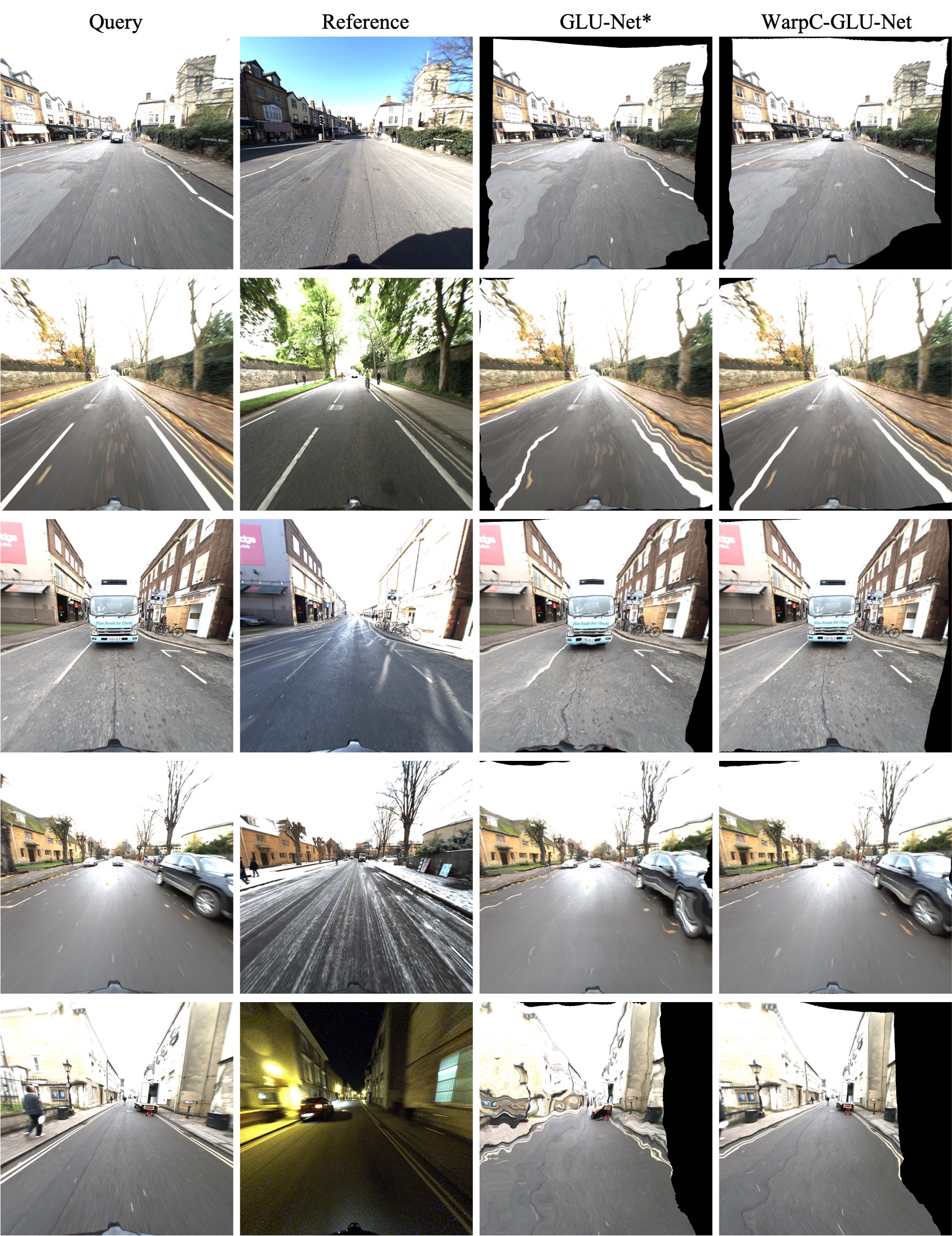}
\caption{Visual comparison on image pairs of the RobotCar dataset~\cite{RobotCar}, of GLU-Net* and WarpC-GLU-Net. 
We visualize the query images warped according to the flow fields estimated by both networks. The warped queries should align with the reference images.}
\vspace{-4mm}
\label{Fig.:robotcat}
\end{figure*}

\begin{figure*}[t]
\centering%
\vspace{-15mm}
\includegraphics[width=0.78\textwidth]{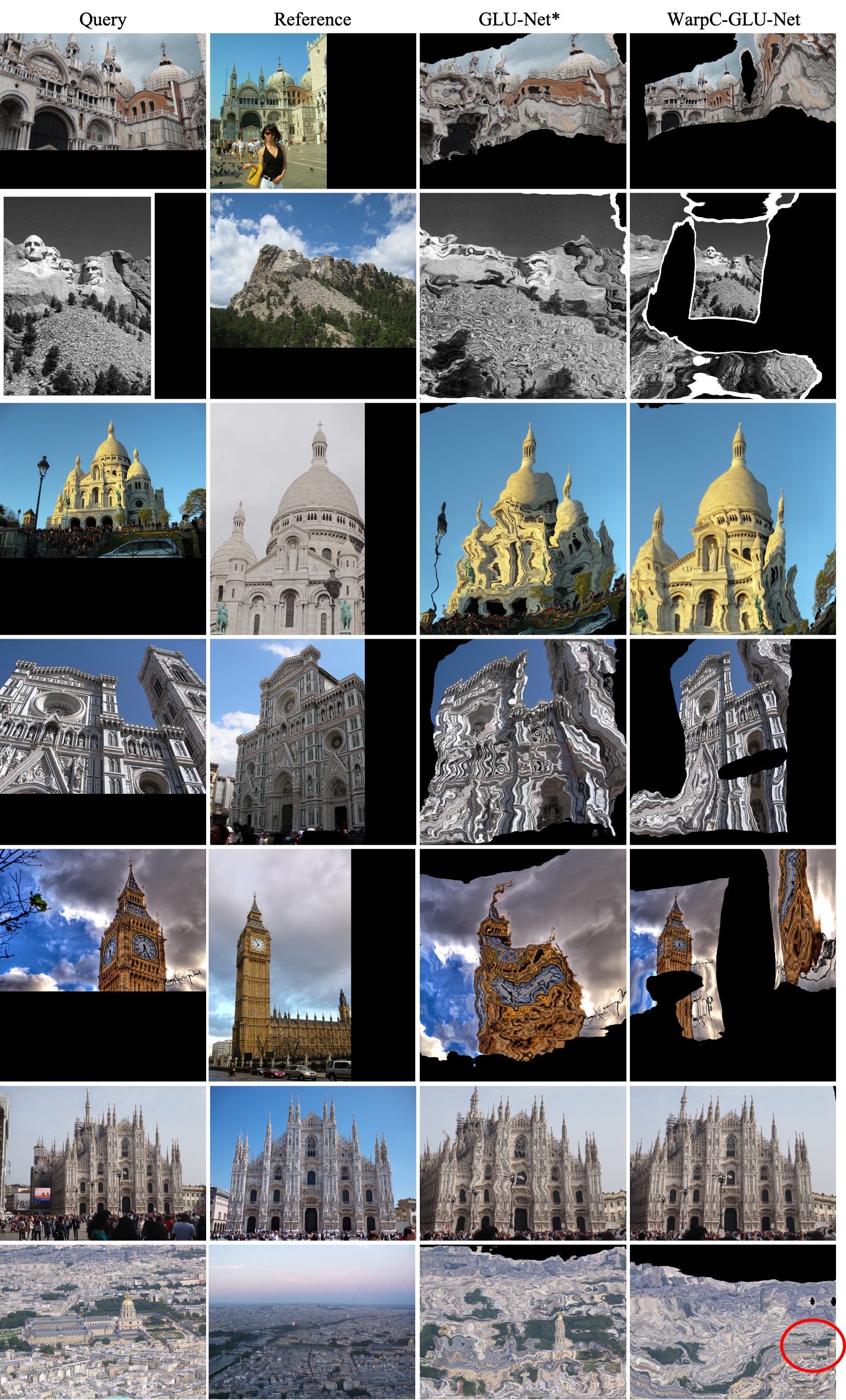}
\caption{Visual comparison on image pairs of the MegaDepth dataset~\cite{megadepth}, of GLU-Net* and WarpC-GLU-Net. 
We visualize the query images warped according to the flow fields estimated by both networks. The warped queries should align with the reference images in overlapping regions. Note that in the dense estimation settings, the network is obligated to predict a match for every pixels in the reference, even in obviously occluded regions. Only correspondences found in overlapping regions are relevant nevertheless. In the last row, we highlight the overlapping region in red, because it is particularly difficult to see. }
\vspace{-4mm}
\label{Fig.:mega-glu-1}
\end{figure*}

\begin{figure*}[t]
\centering%
\vspace{-15mm}
\includegraphics[width=0.77\textwidth]{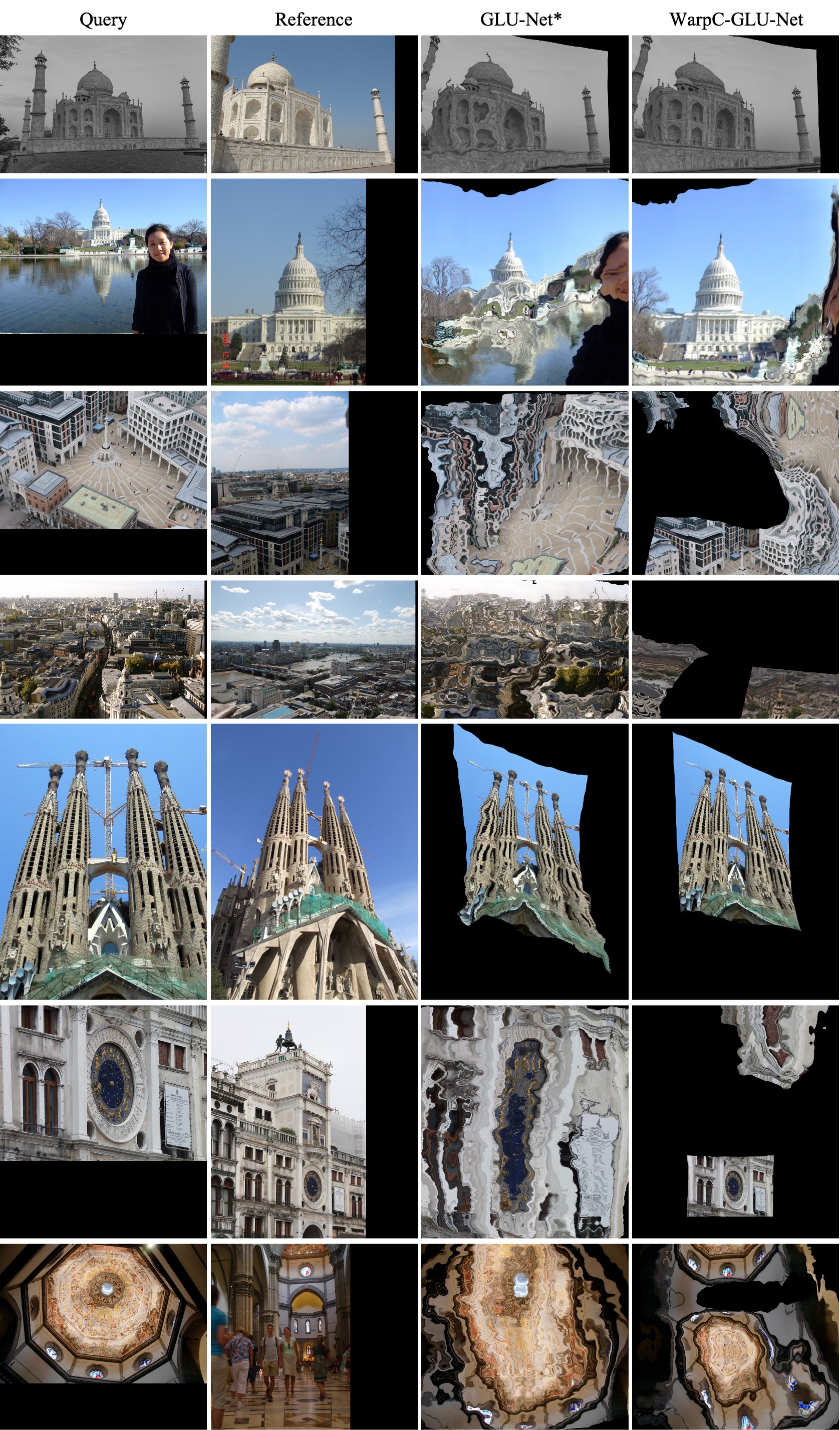}
\caption{Visual comparison on image pairs of the MegaDepth dataset~\cite{megadepth}, of GLU-Net* and WarpC-GLU-Net. We visualize the query images warped according to the flow fields estimated by both networks. The warped queries should align with the reference images in overlapping regions. Note that in the dense estimation settings, the network is obligated to predict a match for every pixels in the reference, even in obviously occluded regions. Only correspondences found in overlapping regions are relevant nevertheless. }
\vspace{-4mm}
\label{Fig.:mega-glu-2}
\end{figure*}

Finally, we provide extensive qualitative visual examples of the performance of our WarpC models. 
We first qualitatively compare baseline GLU-Net* and our approach WarpC-GLU-Net on images of MegaDepth and RobotCar in Fig.~\ref{Fig.:mega-glu-1},~\ref{Fig.:mega-glu-2} and~\ref{Fig.:robotcat} respectively. WarpC-GLU-Net is significantly more accurate than GLU-Net*. It can also handle very drastic scale and view-point changes, where GLU-Net* often completely fails. This is thanks to our \w objective, which provides supervision on the network predictions between the real image pairs, as opposed to the warp-supervision objective. Also note that in the dense estimation settings, the network must predict a match for every pixels in the reference, even in obviously occluded regions. Only correspondences found in overlapping regions are relevant nevertheless. Moreover, occluded regions can be filtered out using \eg a forward-backward consistency mask~\cite{Meister2017}, or by letting the network predict a visibility mask as in~\cite{RANSAC-flow, Melekhov2019}. This is particularly important for MegaDepth images, in which some image pairs have overlapping ratios below $10\%$. 
On RobotCar images in Fig.~\ref{Fig.:robotcat}, our approach WarpC-GLU-Net better handles large appearance variations, such as seasonal or day-night changes.

\begin{figure*}[t]
\centering%
\vspace{-4mm}
\includegraphics[width=0.90\textwidth]{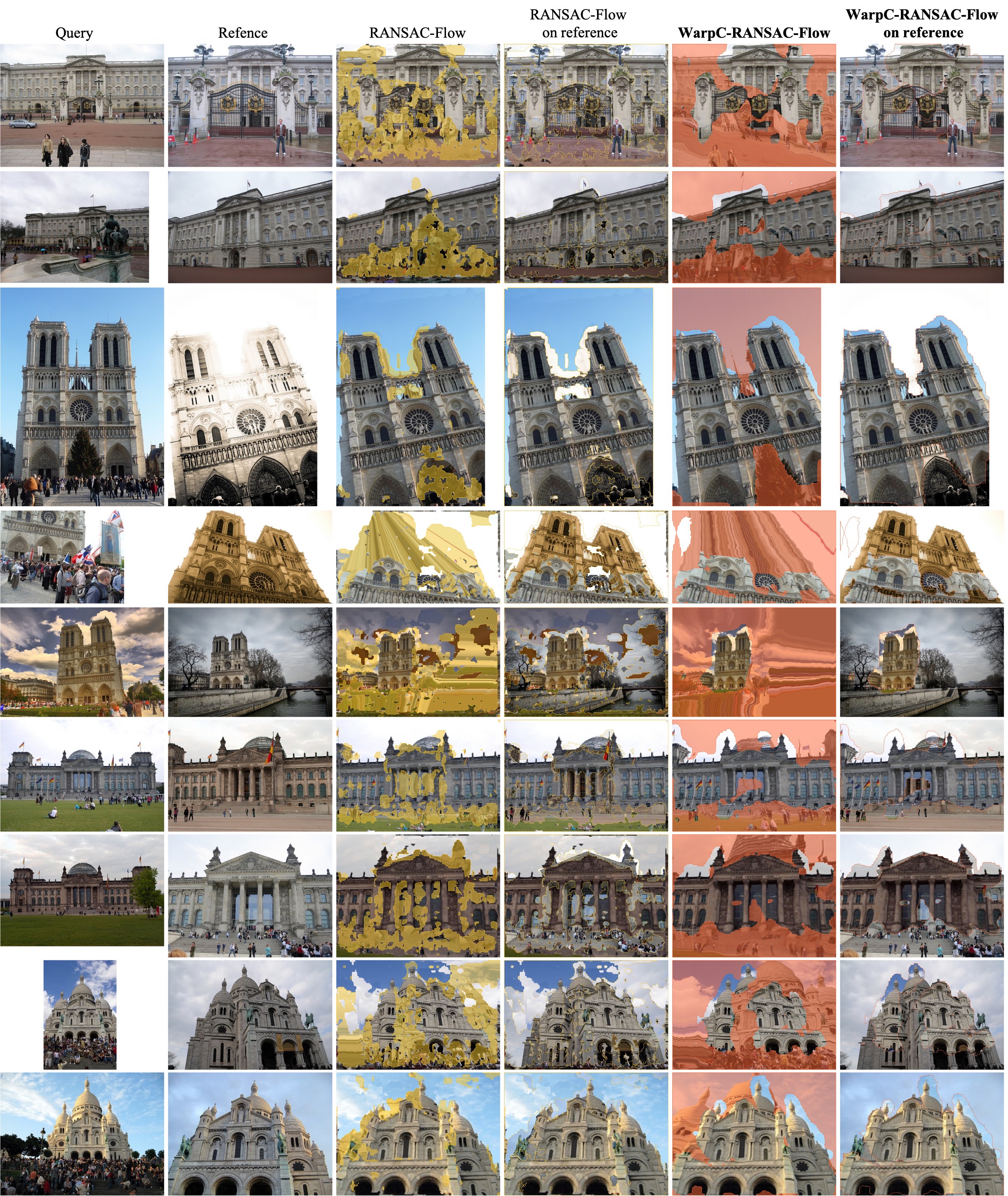}
\caption{Visual comparison of RANSAC-Flow and our approach WarpC-RANSAC-Flow on image pairs of the YFCC100M dataset~\cite{YFCC}. In the 3$^{rd}$ and 5$^{th}$ columns, we visualize the query images warped according to the flow fields estimated by the RANSAC-Flow and WarpC-RANSAC-Flow respectively. Both networks also predict a confidence map, according to which the regions represented in respectively yellow and red, are unreliable or inaccurate matching regions. In the 4$^{th}$ and last columns, we overlay the reference image with the warped query, in the identified accurate matching regions. }
\vspace{-4mm}
\label{Fig.:YFCC}
\end{figure*}

We further show qualitative results of RANSAC-Flow and WarpC-RANSAC-Flow on YFCC100M images in Fig.~\ref{Fig.:YFCC}. Contrary to RANSAC-Flow, the masks predicted by WarpC-RANSAC-Flow correctly filter out unreliable, homogeneous or ambiguous regions, such as the sky or field. 

\begin{figure*}[t]
\centering%
\vspace{-4mm}
\includegraphics[width=0.850\textwidth]{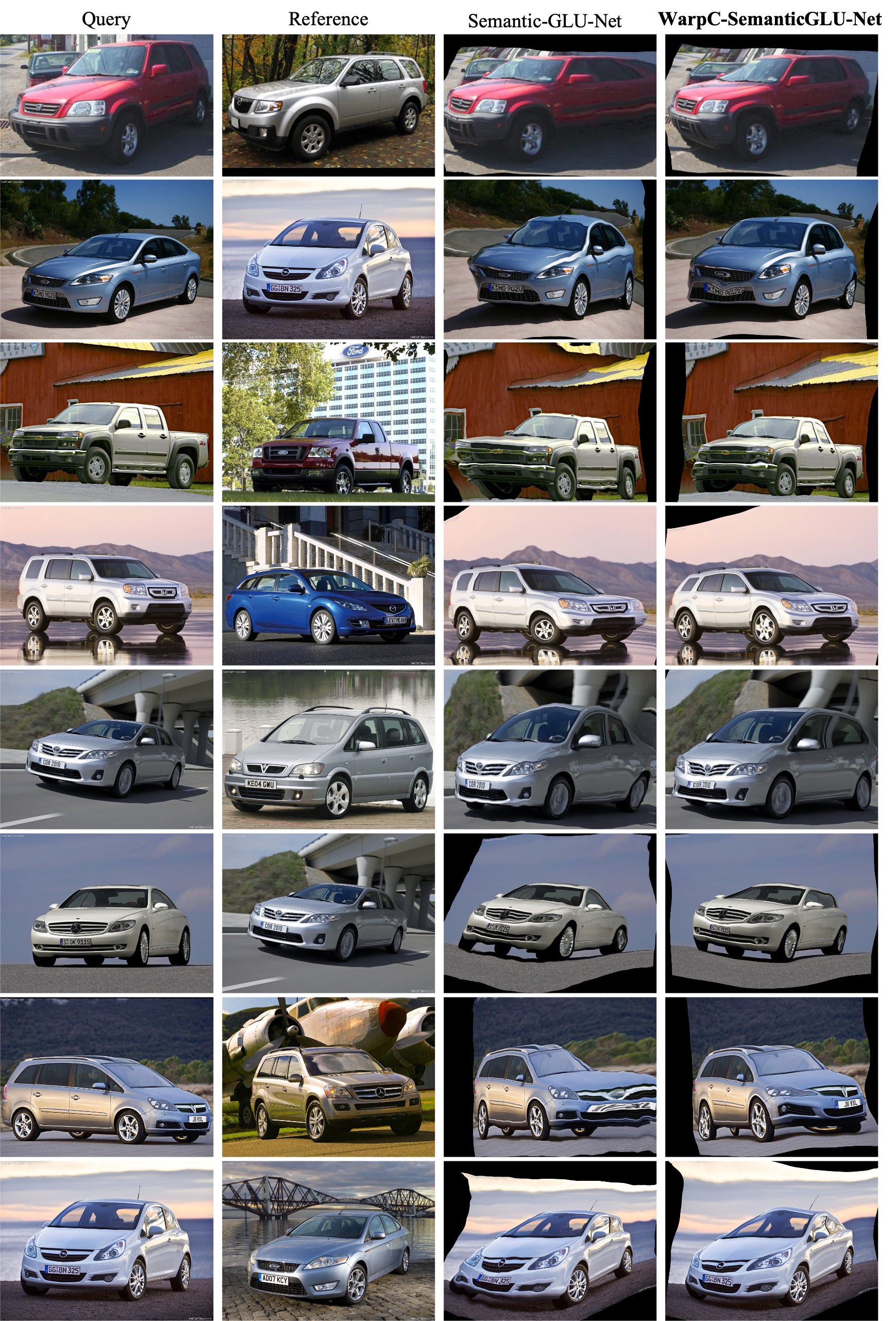}
\caption{Visual comparison on image pairs of the TSS dataset~\cite{YFCC} FD3Car, of original Semantic-GLU-Net~\cite{GLUNet}, trained with the warp-supervision loss on a collection of street-view images, and our approach WarpC-Semantic-GLU-Net, where the network is further finetuned on semantic data using our proposed unsupervised loss. We visualize the query images warped according to the flow fields estimated by Semantic-GLU-Net and WarpC-Semantic-GLU-Net respectively. The warped queries should align with the reference images.}
\label{Fig.:TSS_CARS}
\end{figure*}

\begin{figure*}[t]
\centering%
\vspace{-4mm}
\includegraphics[width=0.90\textwidth]{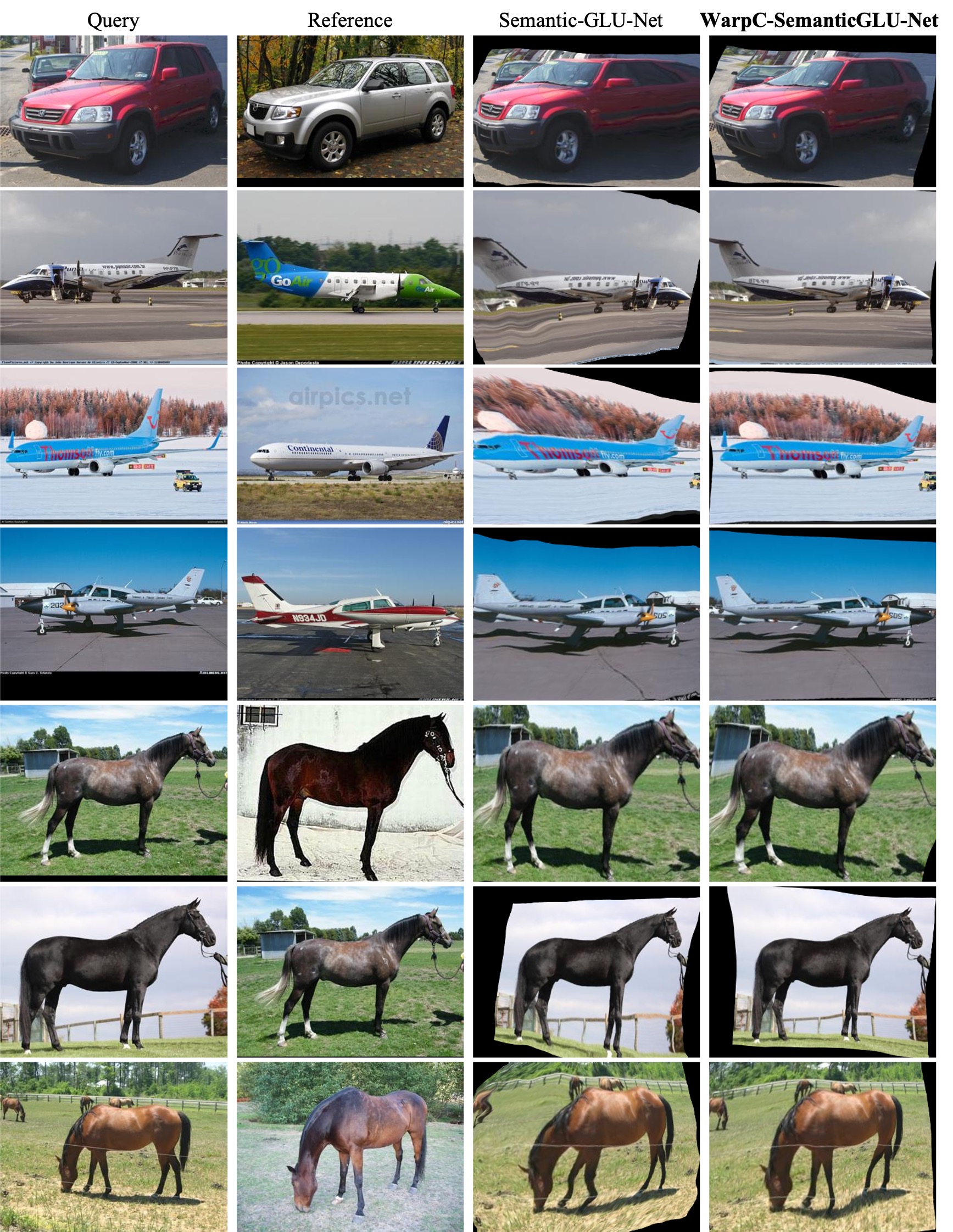}
\caption{Visual comparison on image pairs of the TSS dataset~\cite{Taniai2016} JODS, of original Semantic-GLU-Net~\cite{GLUNet}, trained with the warp-supervision loss on a collection of street-view images, and our approach WarpC-Semantic-GLU-Net, where the network is further finetuned on semantic data using our proposed unsupervised loss. We visualize the query images warped according to the flow fields estimated by Semantic-GLU-Net and WarpC-Semantic-GLU-Net respectively. The warped queries should align with the reference images.}
\vspace{-4mm}
\label{Fig.:TSS_JODS}
\end{figure*}

\begin{figure*}[t]
\centering%
\vspace{-4mm}
\includegraphics[width=0.90\textwidth]{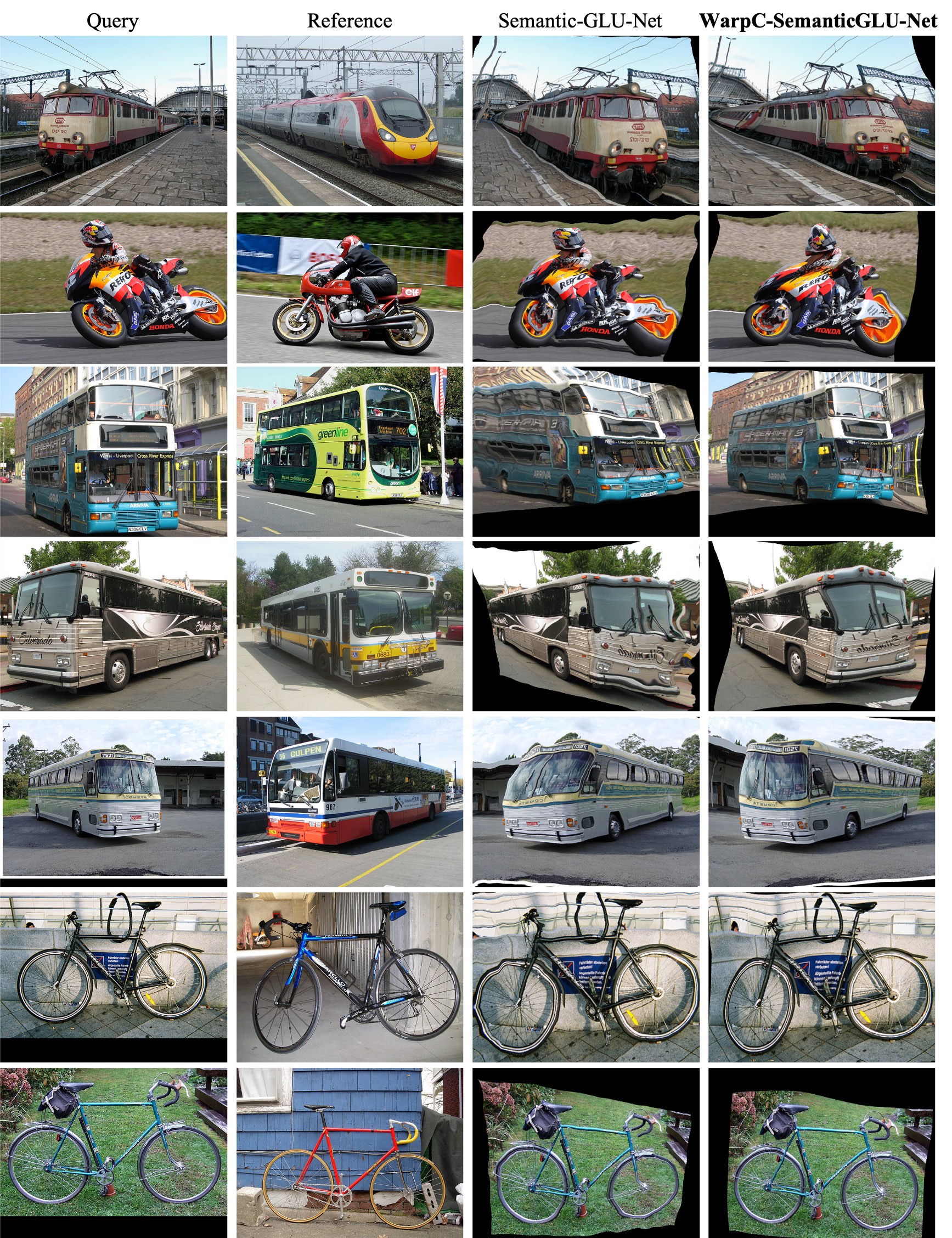}
\caption{Visual comparison on image pairs of the TSS dataset~\cite{Taniai2016} PASCAL, of original Semantic-GLU-Net~\cite{GLUNet}, trained with the warp-supervision loss on a collection of street-view images, and our approach WarpC-Semantic-GLU-Net, where the network is further finetuned on semantic data using our proposed unsupervised loss. We visualize the query images warped according to the flow fields estimated by Semantic-GLU-Net and WarpC-Semantic-GLU-Net respectively. The warped queries should align with the reference images.}
\vspace{-4mm}
\label{Fig.:TSS_PASCAL}
\end{figure*}

\begin{figure*}[t]
\centering%
\vspace{-8mm}\includegraphics[width=0.75\textwidth]{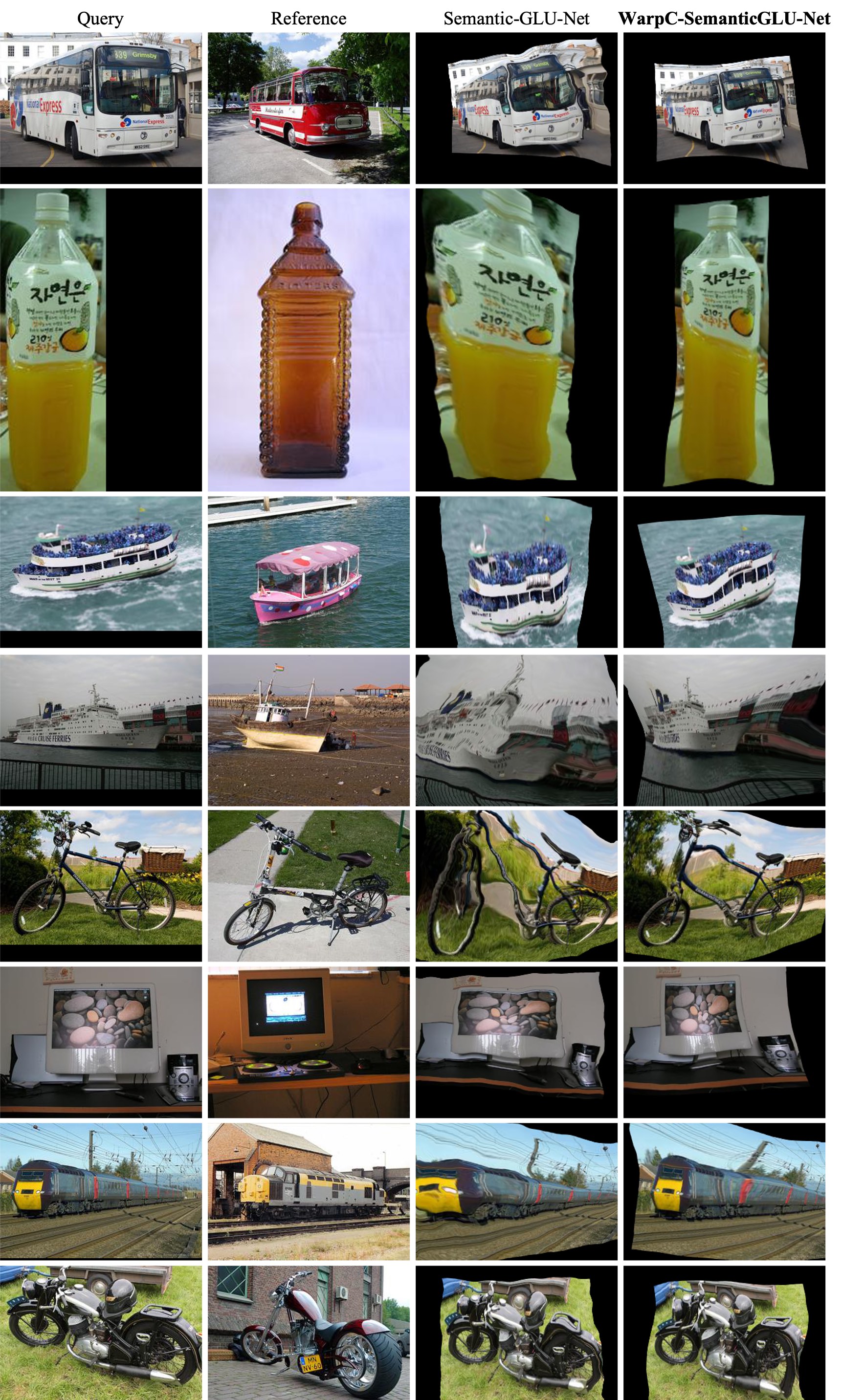}
\caption{Visual comparison on image pairs of the PF-Pascal dataset~\cite{PFPascal}, of original Semantic-GLU-Net~\cite{GLUNet}, trained with the warp-supervision loss on a collection of street-view images, and our approach WarpC-Semantic-GLU-Net, where the network is further finetuned on semantic data using our proposed unsupervised loss. We visualize the query images warped according to the flow fields estimated by Semantic-GLU-Net and WarpC-Semantic-GLU-Net respectively. The warped queries should align with the reference images.}
\vspace{-4mm}
\label{Fig.:intro_PF_1}
\end{figure*}

\begin{figure*}[t]
\centering%
\vspace{-8mm}\includegraphics[width=0.75\textwidth]{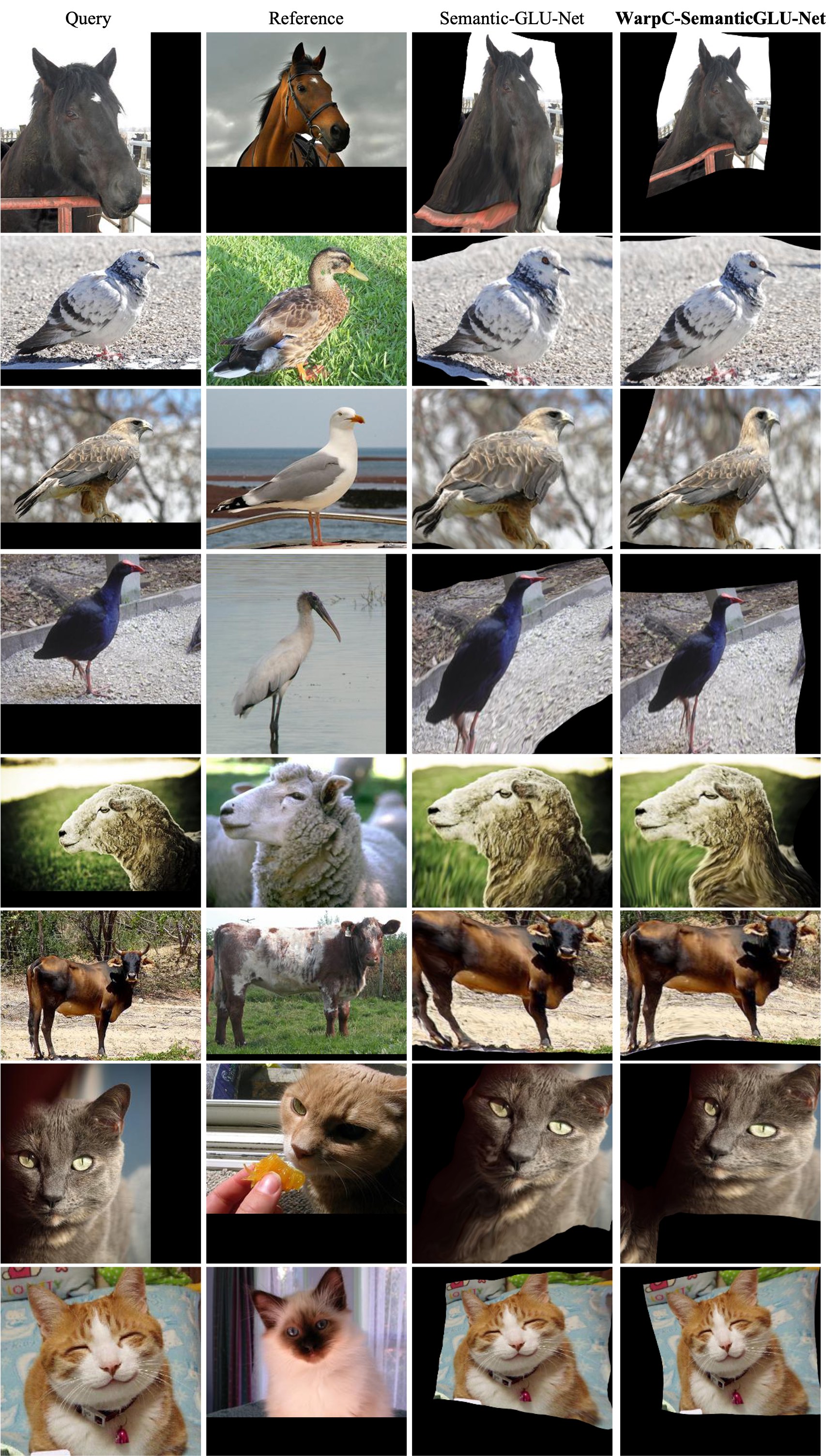}
\caption{Visual comparison on image pairs of the PF-Pascal dataset~\cite{PFPascal}, of original Semantic-GLU-Net~\cite{GLUNet} and our approach WarpC-Semantic-GLU-Net.}
\vspace{-4mm}
\label{Fig.:intro_PF_2}
\end{figure*}

\begin{figure*}[t]
\centering%
\vspace{-15mm}\includegraphics[width=0.75\textwidth]{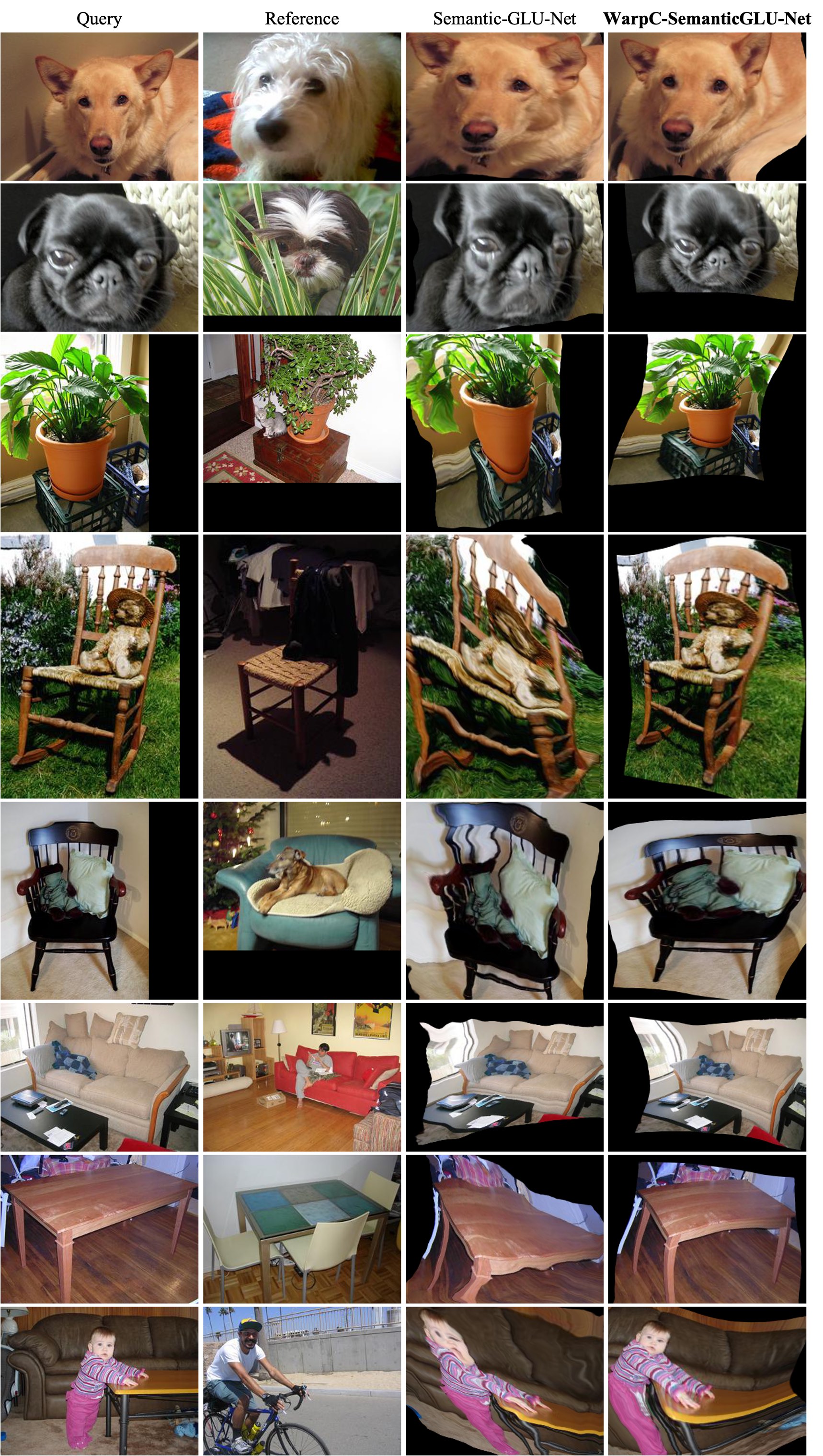}
\caption{Visual comparison on image pairs of the PF-Pascal dataset~\cite{PFPascal}, of original Semantic-GLU-Net~\cite{GLUNet} and our approach WarpC-Semantic-GLU-Net.}
\vspace{-4mm}
\label{Fig.:intro_PF_3}
\end{figure*}

We then show the performance of WarpC-SemanticGLU-Net compared to SemanticGLU-Net on images of TSS in Fig.~\ref{Fig.:TSS_CARS},~\ref{Fig.:TSS_JODS} and~\ref{Fig.:TSS_PASCAL}. Our unsupervised finetuning brings visible robustness to the large appearance changes and shape variations inherent to the semantic matching task. 
Finally, we also qualitatively compare both networks on images of the PF-Pascal dataset in Fig.~\ref{Fig.:intro_PF_1},~\ref{Fig.:intro_PF_2} and~\ref{Fig.:intro_PF_3}. The PF-Pascal dataset shows more diverse object categories than TSS images. WarpC-SemanticGLU-Net manages to accurately align challenging image pairs, such as the chair examples which are particularly cluttered.

\end{document}